\documentclass[acmlarge]{acmart}

\AtBeginDocument{%
  \providecommand\BibTeX{{%
    \normalfont B\kern-0.5em{\scshape i\kern-0.25em b}\kern-0.8em\TeX}}}
    
\setcopyright{acmcopyright}
\acmJournal{IMWUT}
\acmYear{2022}
\acmVolume{6}
\acmNumber{4}
\acmArticle{172}
\acmMonth{12}
\acmPrice{\$15.00}
\acmDOI{10.1145/3569464}

\usepackage[ruled]{algorithm2e}
\usepackage{enumerate}
\usepackage{enumitem}
\usepackage{graphicx}
\usepackage{hyperref}
\usepackage{multirow, makecell}
\usepackage{nccmath}
\usepackage{ragged2e}
\usepackage[caption=false,font=normalsize]{subfig}
\usepackage{threeparttable} 
\usepackage{url}
\usepackage{xspace}

\newcommand{\ie}{\emph{i.e.},\xspace}

\newcommand{\eg}{\emph{e.g.},\xspace}
\newcommand{\etc}{\emph{etc.}\xspace}
\newcommand{\etal}{\emph{et al.}\xspace}

\newcommand\figref[1]{Fig.~\ref{#1}}

\newcommand\tabref[1]{Tab.~\ref{#1}}

\newcommand\secref[1]{Sec.~\ref{#1}}

\newcommand\equref[1]{Equ.(\ref{#1})}

\newcommand{\sysname}{{\sf AdaEnlight}\xspace}
\newcommand{\sysnameposs}{{\sf AdaEnlight's}\xspace}

\ifodd 1

\newcommand\rev[1]{\textcolor{black}{#1}}
\newcommand\newrev[1]{\textcolor{black}{#1}}

\else

\newcommand\rev[1]{#1}

\fi

\begin{document}

\title{AdaEnlight: Energy-aware Low-light Video Stream Enhancement on Mobile Devices}

% \author{Anonymous}\affiliation{%
% \institution{Anonymous}
% \country{Anonymous}
% }

\author{Sicong Liu}
\orcid{0000-0003-4402-1260}
\affiliation{%
  \institution{Northwestern Polytechnical University}
  \department{School of Computer Science}
  \city{Xi'an}
  \country{China}
}

\author{Xiaochen Li}
\orcid{0000-0003-2653-5786}
\affiliation{%
  \institution{Northwestern Polytechnical University}
  \department{School of Computer Science}
  \city{Xi'an}
  \country{China}
}

\author{Zimu Zhou}
\orcid{0000-0002-5457-6967}
\affiliation{%
  \institution{City University of Hong Kong}
  \department{School of Data Science}
  \city{Hong Kong}
  \country{China}
}

\author{Bin Guo}
\orcid{0000-0001-6097-2467}
\affiliation{%
  \institution{Northwestern Polytechnical University}
  \department{School of Computer Science}
  \city{Xi'an}
  \country{China}
  \thanks{Corresponding author: guob@nwpu.edu.cn}
}

\author{Meng Zhang}
\orcid{0000-0003-0637-249X}
\affiliation{%
  \institution{Northwestern Polytechnical University}
  \department{School of Computer Science}
  \city{Xi'an}
  \country{China}
}

\author{Haocheng Shen}
\orcid{0000-0001-8472-6244}
\affiliation{%
  \institution{Northwestern Polytechnical University}
  \department{School of Computer Science}
  \city{Xi'an}
  \country{China}
}

\author{Zhiwen Yu}
\orcid{0000-0002-9905-3238}
\affiliation{%
  \institution{Northwestern Polytechnical University}
  \department{School of Computer Science}
  \city{Xi'an}
  \country{China}
}

% \author{xx xx}
% \affiliation{%
%   \institution{Northwestern Polytechnical University}
%   \department{School of Computer Science}
%   \city{Xi'an}
%   \country{China}
% }

\renewcommand{\shortauthors}{Liu et al.}
\renewcommand{\shorttitle}{\sysname: Energy-aware Low-light Video Stream Enhancement on Mobile Devices}

\begin{abstract}
The ubiquity of camera-embedded devices and the advances in deep learning have stimulated various intelligent mobile video applications.
These applications often demand on-device processing of video streams to deliver real-time, high-quality services for privacy and robustness concerns.
However, the performance of these applications is constrained by the raw video streams, which tend to be taken with small-aperture cameras of ubiquitous mobile platforms in dim light.
Despite extensive low-light video enhancement solutions, they are unfit for deployment to mobile devices due to their complex models and and ignorance of system dynamics like energy budgets.
In this paper, we propose \sysname, an energy-aware low-light video stream enhancement system on mobile devices.
It achieves real-time video enhancement with competitive visual quality while allowing runtime behavior adaptation to the platform-imposed dynamic energy budgets.
We report extensive experiments on diverse datasets, scenarios, and platforms and demonstrate the superiority of \sysname compared with state-of-the-art low-light image and video enhancement solutions.
\end{abstract}




\maketitle

\section{Introduction}
\label{sec:intro}

%1. 移动摄像头嵌入设备逐渐流行，捕获的泛在视频流越来越多。然而这类视频经常受到低照度影响。需要做增强。在本地增强的需求很重要，因为本地视频处理的应用越来越多。--》引出视频增强的必要性。
The pervasive deployment of camera-embedded mobile devices \eg smartphones, wearables, tablets, and robots has stimulated a wide spectrum of novel mobile video based applications.
Examples include face detection on smartphones for user authentication \cite{bib:percom2020:fan}, outdoor surveillance with mobile robots \cite{bib:IDST2021:al, bib:INDIN12:Juang}, and object detection with drones~\cite{bib:ICCV2019:zhu}, \etc
These applications are features with two characteristics.
\textit{(i)}
There is a growing interest in \textit{on-device processing} of these video streams rather than cloud offloading for privacy concerns or real-time interactions \cite{bib:Computer17:Ananthanarayanan}.
\textit{(ii)}
The video streams captured by mobile devices often suffer from \textit{low-light effect} since they are typically taken by non-professional users with narrow-aperture cameras \cite{bib:ConTEL19:Teli, bib:TIP19:Ren}.
Low-light videos can notably impair user experience \eg for video display, and downstream application accuracy, \eg face detection. 

Despite extensive research on low-light image and video enhancement \cite{bib:ICCV19:Chen, bib:CVPR20:Guo, bib:TIP16:Guo, bib:ICCV19:Jiang, bib:BMVC18:Lv, bib:TPAMI21:Li, bib:TPAMI21:Li2, bib:CVPR21:Zhang}, they are unfit for deployment to mobile devices because they fail to meet the following requirements.
\begin{itemize}
    \item \textit{Real-time Processing on Resource-constrained Mobile Devices.}
    The \textit{real-time} processing of video stream is necessary for application's responsiveness, and \textit{on-device} processing benefits the user privacy. 
    However, mainstream low-light video enhancement schemes are computation-intensive \cite{bib:BMVC18:Lv, bib:ICCV19:Jiang, bib:ICCV19:Chen, bib:CVPR21:Zhang}.
    Even the state-of-the-art lightweight image enhancement model Zero-DCE++ \cite{bib:TPAMI21:Li2} fails to achieve real-time processing on mobile devices such as a Raspberry Pi ($5$ $frames/s$ for $270 \times 480$ RGB images, see \secref{sec:experiment}).
    \item \textit{Adapting to Diverse Energy Supply of Mobile Devices.}
    \textit{Energy awareness} is crucial for the long-term operation of the above video processing since mobile devices are often battery-powered~\cite{bib:SOSP99:Flinn, bib:todaes2000:benini}.
    It is desired that the low-light video processing module can adapt its enhancement quality according to the energy budget at runtime.
    Despite prior studies on adaptive mobile computer vision \cite{bib:sigcomm2020:kim, bib:mobicom2019:baig, bib:hotcloud2019:wang, bib:SIGCOMM2016:sun, bib:mobicom2019:lee, bib:tmc2020:yi} and general energy profiling of neural networks \cite{bib:CVPR17:Yang, bib:TCS20:Lee, bib:ssc2017:yang, bib:ICCAD2019:wu,bib:islped2018:ke, bib:ISPASS2019:parashar}, there lacks a runtime energy profiler and adaptation loop dedicated to deep learning enabled low-light video enhancement.
\end{itemize}

\begin{figure}[t]
  \centering
  \includegraphics[width=.9\textwidth]{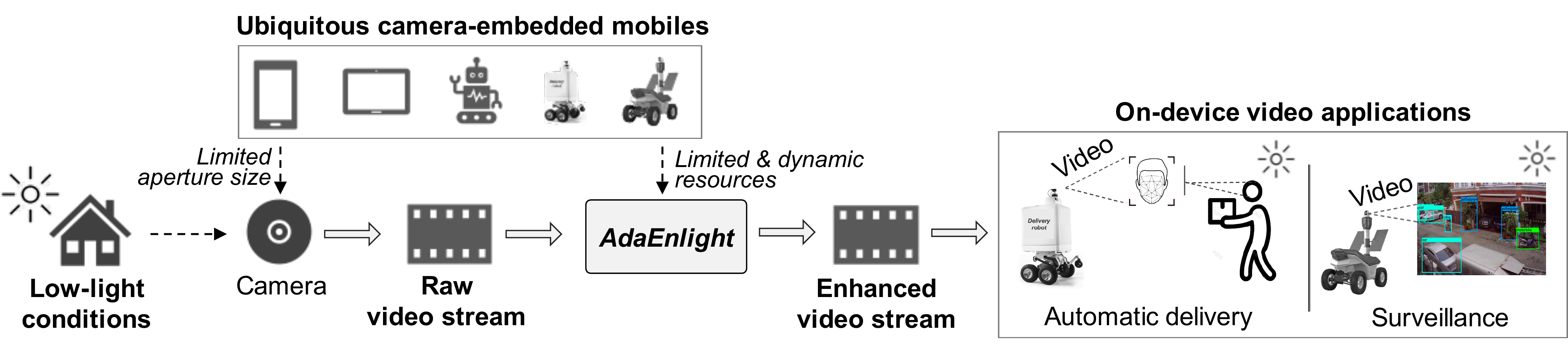}
  \caption{Illustration of \sysname as a video enhancement middleware for downstream mobile video applications.}
  \label{fig:middleware_overview}
  \vspace{-3mm}
\end{figure}

In this paper, we propose \sysname, an energy-aware on-device low-light video enhancement solution for mobile platforms. 
It is challenging to achieve high enhanced visual quality, low execution latency, and adaptation to energy budget on mobile platforms.
\sysname tackles these challenges via a modular system design and a set of novel algorithms.
%Specifically, \sysname decouples the requirements on visual quality, latency and energy into two functional modules, fast low-light video enhancement and energy-aware adaptation controller.
Observing that latency bottleneck of the state-of-the-art image enhancement schemes \cite{bib:CVPR20:Guo, bib:TPAMI21:Li2} lies in the iterative enhancement curve function, \sysname devises a novel non-iterative curve function for acceleration without compromising enhancement quality.
\sysname also incorporates temporal consistency among frames to mitigate the flicking problem \cite{bib:TPAMI21:Li} when enhancing video streams.  
To adapt to the dynamic energy budget imposed by mobile platforms, \sysname exploits a runtime energy profiler to estimate the energy cot of the video enhancement process at runtime and configures its hyper-parameters (\eg computation reuse and frame resolution) accordingly.
We implement \sysname as a compact video pre-processing middleware to provide enhanced video streams for various downstream mobile video applications such as facial recognition in automatic package delivery and object detection in video surveillance (see \figref{fig:middleware_overview}).

Our main contributions are summarized as follows.
\begin{itemize}
    \item 
    To the best of our knowledge, \sysname is the first \textit{on-device} low-light \textit{video} enhancement system for \textit{mobile} scenarios. It achieves near real-time and high-quality processing on commodity mobile platforms and stays adaptive to the dynamic energy budget.
    \item
    The key technical novelties include a non-iterative low-light enhancement model that breaks %removes 
    the latency bottleneck of the state-of-the-art enhancement schemes and enforces temporal consistency between video frames. %incorporates temporal consistency.
    Also, it adopts a runtime energy-aware controller that integrate a runtime energy profiler to adjust the video enhancement behaviors adaptively.
    \item 
    We evaluate \sysname on four public benchmarks and real-world mobile scenarios on three mobile platforms (Honor 9 smartphone, raspberry Pi 4B, and Jetson AGX Xavier).
    Experiments show that \sysname achieves near real-time ($30\sim50ms$ per frame) video enhancement with competitive visual quality to existing image and video enhancement schemes. 
    \sysname also enables agile self-adaptation to satisfy the dynamic energy budgets at runtime.
\end{itemize}

% The main contributions of this work are summarized as follows.
% %6. 三个贡献点
% \begin{itemize}
%     \item We present a novel low-light video enhancement model by removing the latency bottleneck of the state-of-the-art low-light visual enhancement model and enforcing temporal consistency between video frames.
%     \item To the best of our knowledge, this is the first work that integrates the energy-aware adaptation controller into the video enhancement system. Also, we propose an accurate runtime energy profiler for it.
%     \item Experiments show that \sysname achieves the real-time (\eg $30\sim50ms$ per frame) video enhancement with competitive visual quality to existing image/video enhancement baselines. \sysname also enables the agile self-adaptation for satisfying dynamic energy budgets.
% \end{itemize}

%7. 文章布局
In the rest of this paper, we present an overview of \sysname as well as its functional modules in \secref{sec:overview}, \secref{sec:algorithm}, and \secref{sec:controller}.
We show the evaluations in \secref{sec:experiment}, review related work in \secref{sec:related}, and conclude in \secref{sec:conclude}.

\section{\sysname Overview}
\label{sec:overview}
This section presents our problem scope and the design overview of \sysname. 

\subsection{Problem Scope}
\label{subsec:scope}
In short, we target at on-device low-light enhancement of ubiquitous video streams.
We elaborate on the concrete motivations and requirements below.

\begin{figure}[t]
  \centering
  \includegraphics[width=.97\textwidth]{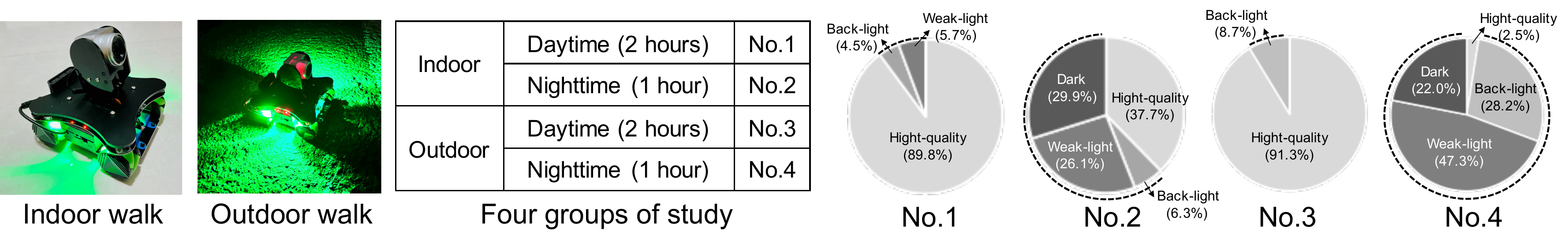}
  \caption{An example study of the ubiquity of low-light video streams in the mobile video surveillance scenario.}
  \label{fig:study}
  \vspace{-3mm}
\end{figure}

\subsubsection{Motivations}
\label{subsubsec:motivations}
Our work is motivated by the ubiquity of camera-embedded devices (\eg smartphones, wearables, tablets, and robots) in everyday life for various video applications ranging from video surveillance \cite{bib:INDIN12:Juang} to video conferencing \cite{app:zoom}.
These video streams, which we call as \textit{ubiquitous video streams}, are often captured by \textit{non-professional users} and are likely to experience \textit{low-light} conditions.
The reasons are two-fold.
\textit{(i)} 
The limited camera aperture size on mobile devices restrict the overall amount of light that reaches the camera sensor and thereby decrease the signal-to-noise ratio of the captured videos~\cite{bib:ConTEL19:Teli}.
\textit{(ii)} 
Non-professional users may take videos under uncontrolled lighting conditions, \eg back-lit, weak light, and extremely dark environments \cite{bib:TIP19:Ren}.
Also, we conduct an example study using a mobile robot (\ie Yahboom) to show the ubiquity of low-light video streams in the daily video surveillance scenario. 
The robot continuously collects video streams by random walking indoor and outdoor for 6 hours, at daytime and nighttime. 
\figref{fig:study} shows the details of four groups of study. 
The proportion of low-light video streams is up to $87.5\%$ (\ie group No.4) and $10.5\%$ (\ie group No.1) at nighttime and daytime, respectively. 

\begin{figure*}[t]
  \centering
    \subfloat[]{\label{fig:encode_unencode}
    \includegraphics[height=0.22\textwidth]{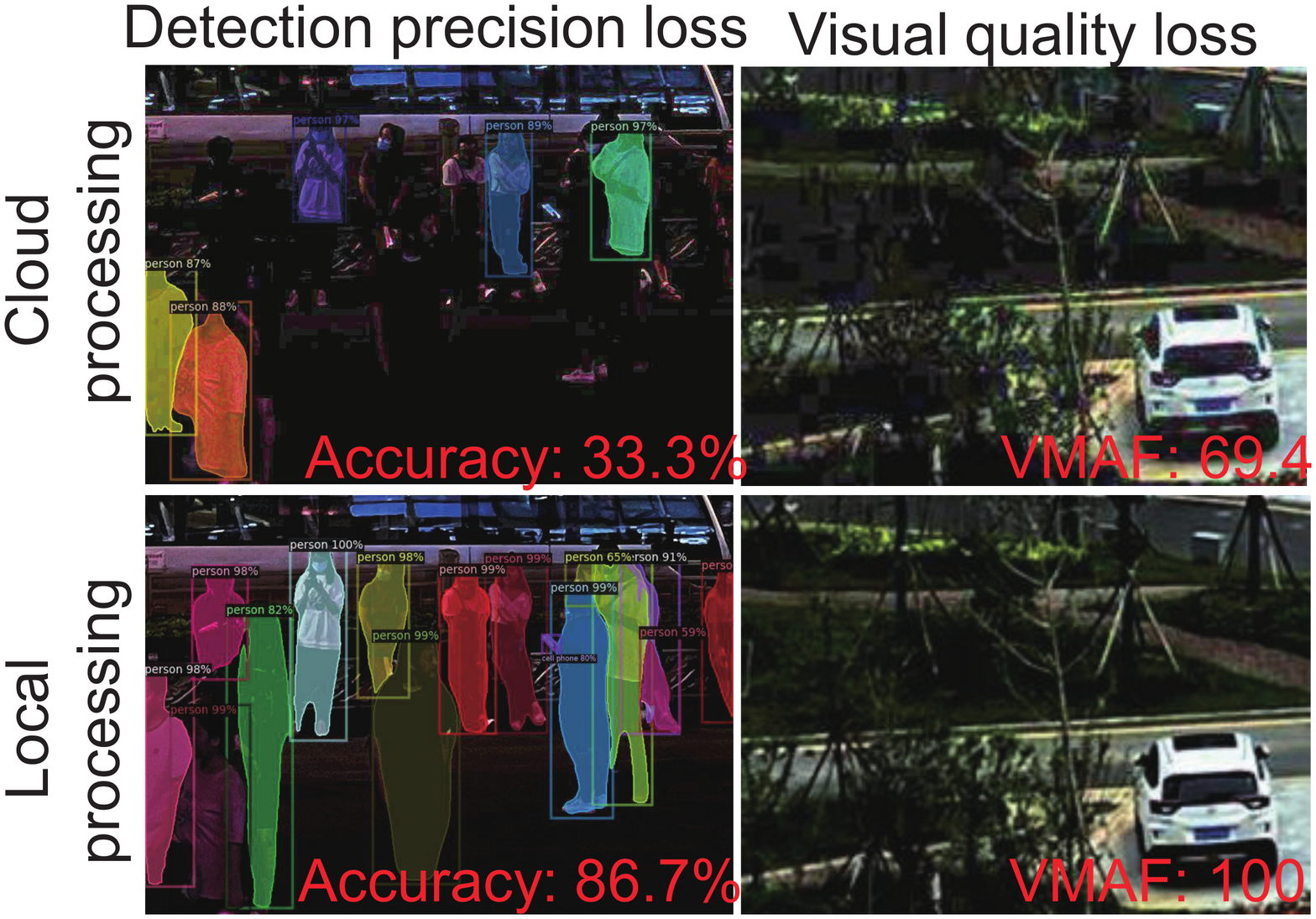}}
    \hspace{15pt}
    \subfloat[]{\label{fig:bitrate_vmaf}
    \includegraphics[height=0.22\textwidth]{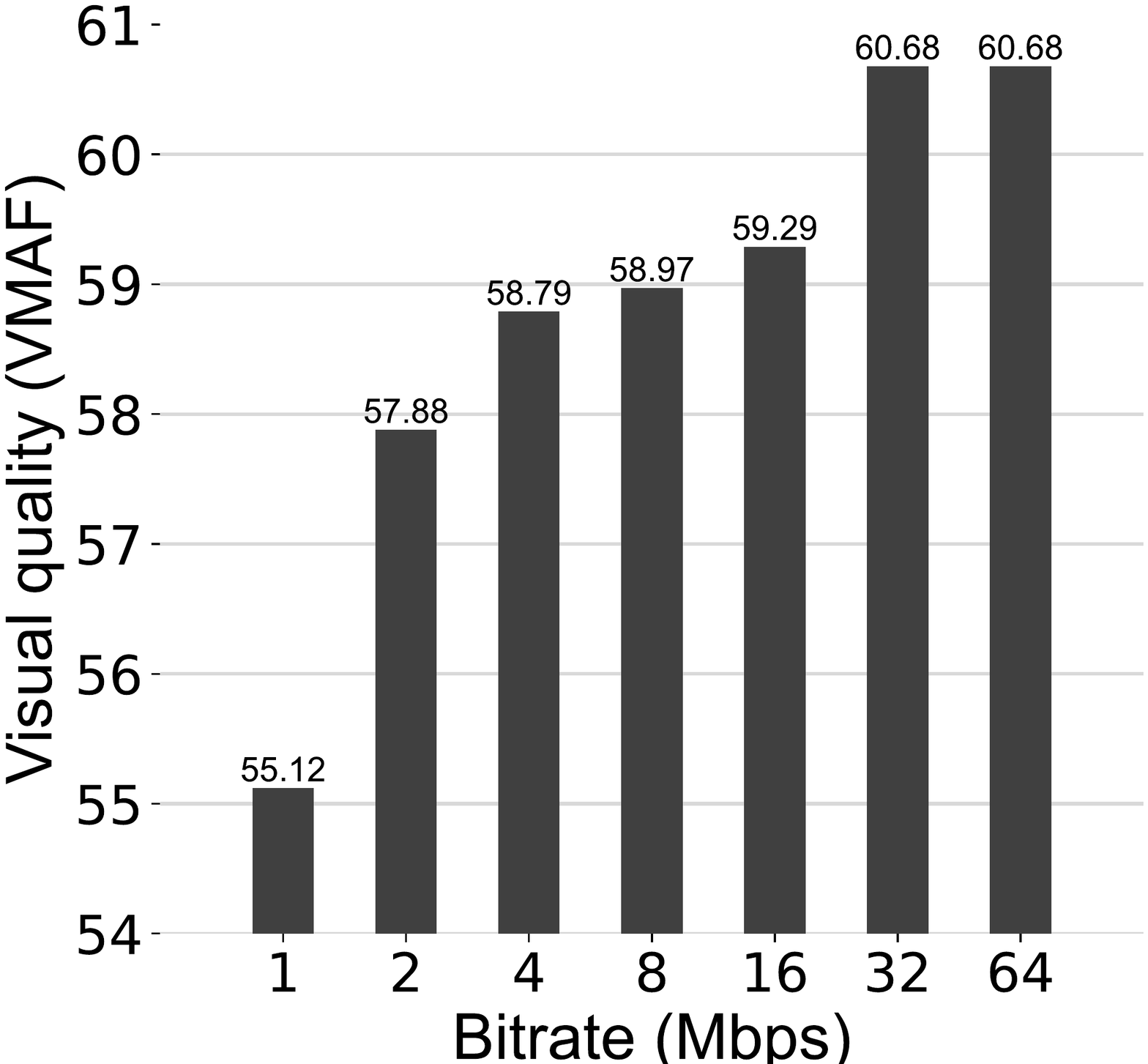}}
    \vspace{-3mm}
\caption{\rev{Comparison of video enhancement locally on-device and remotely on-cloud. (a) Example frames enhanced by local processing and remote processing. The upper row follows the encode-decode-enhance pipeline of remote processing, while the lower row are directly enhanced as in local processing. (b) Enhanced video quality of the encode-decode-enhance pipeline under different frame bitrates into the video codec.}}
\label{fig:local_compare}
\vspace{-3mm}
\end{figure*}

\rev{We consider video enhancement \textit{locally on-device} rather than \textit{remotely on cloud} for the following reasons.}
\begin{itemize}
    \item 
    The ubiquitous video streams may be immediately consumed on the mobile devices \cite{bib:Computer17:Ananthanarayanan}.
    For example, a surveillance drone may run object recognition algorithms on video streams taken in dim light and only uploads suspicious frames to cloud servers to save bandwidth.
    An automatic delivery robot may authenticate the target recipient before delivery by running face recognition algorithms locally for privacy concerns. 
    \item 
    \rev{On-device processing avoids the extra loss in visual quality due to video encoding during data transmission.
    Specifically, offloading video enhancement to cloud follows an ``encode-decode-enhance'' pipeline.
    Frames are encoded at the mobile device, transmitted via wireless networks, decoded at the cloud, and finally fed into the video enhancement module.
    Video encoding is necessary due to the limited wireless bandwidth.
    Yet it can decrease the enhanced visual quality because many mainstream video encoding schemes \eg H.264 \cite{bib:H2642003overview} are lossy.
    \figref{fig:encode_unencode} shows an example, where the upper row shows the frames enhanced following cloud processing \ie encode-decode-enhance, while the lower row shows the results of local processing \ie directly enhance.
    There is notable loss in detection precision (53.4\% accuracy loss in the upper left figures) and visual quality (30.6 VMAF loss in the upper right figures).
    \figref{fig:bitrate_vmaf} provides a more quantitative study.
    We enhance one example video clip from our self-collected dataset (MobileScene, see \secref{sec_exp_setup} for details) following the ``encode-decode-enhance'' pipeline using the standard H.264 video codec for encoding/decoding and \sysname for enhancement. 
    We vary the input bitrate into the codec and measure the visual quality of the enhanced frames by the VMAF index \cite{bib:li2016toward}.
    As is shown, the bitrate (and thus the network bandwidth) limits the visual quality of video enhancement.
    The VMAF index is only $55$ at the bitrate of $1$ Mbps.
    Even at a bitrate of $64$ Mbps (higher than the uplink bandwidth of commercial 5G networks \cite{t-mobile-5G-report}), the VMAF index is only around $60$, which is lower than that of local processing (see \tabref{tb_energy} and \tabref{tb_energy_nano} in our evaluations).}
\end{itemize}

% \begin{figure*}[t]
%   \centering
%     \subfloat{\includegraphics[height=0.12\textwidth]{image/compressed_1.pdf}}
%     \subfloat{\includegraphics[height=0.12\textwidth]{image/uncompressed_1.pdf}}
%     \hspace{6pt}\vspace{6pt}
%     \subfloat{\includegraphics[height=0.12\textwidth]{image/compressed_2.pdf}}
%     \subfloat{\includegraphics[height=0.12\textwidth]{image/uncompressed_2.pdf}}
%     \hspace{6pt}
%     \subfloat{\includegraphics[height=0.12\textwidth]{image/compressed_3.pdf}}
%     \subfloat{\includegraphics[height=0.12\textwidth]{image/uncompressed_3.pdf}}
% \caption{Visual results comparison between cloud and local process. In each pair of images, the left one is processed with encode-decode-enhance-encode(pipeline of cloud process) while the right one is processed with enhance-encode(pipeline of local process). }
% \label{fig:encode_unencode}
% \vspace{-3mm}
% \end{figure*}

% \begin{figure*}[t]
%     \centering
%     \includegraphics[width=.5\textwidth]{image/Bitrate_VMAF.pdf}
%     \caption{Visual quality with different bitrate}
%     \label{fig:bitrate_vmaf}
% \end{figure*}

%load the facial recognition algorithm to locally detect the human face on video streams, for privacy preservation.
%And only the correct result that matches the target recipient can trigger the delivery of goods.
%
%In addition, the facial recognition for identity authentication on the mobile phone also desires to process the user's video stream locally for protecting privacy and reducing response delay.

\subsubsection{Requirements}
\label{subsubsec:requirements}
To support the low-light enhancement of ubiquitous video streams on mobile devices, a solution should meet the following criteria.
\begin{itemize}
    \item \textit{Near Real-time Video Enhancement}.
    As with other mobile computer vision tasks ~\cite{bib:icce2014:song, bib:nca2020:wijnands, bib:cvpr2018:liu},
    %\TODO{refs}, 
    the low-light enhancement model should be lightweight to fit in the limited execution resources on mobile platforms and to achieve near real-time processing of the video streams.
    \item \textit{Energy-aware Adaptation}.
    Since many mobile devices are battery-powered, it is crucial that the video enhancement is energy-aware. 
    Particularly, it is desirable that the video enhancement process can trade off between the output visual quality and the platform-imposed energy budget at runtime.
\end{itemize}
Despite prior research on the low-light image/video enhancement \cite{bib:ICCV19:Chen, bib:CVPR20:Guo, bib:TIP16:Guo, bib:ICCV19:Jiang, bib:BMVC18:Lv, bib:TPAMI21:Li, bib:TPAMI21:Li2, bib:CVPR21:Zhang} and adaptive mobile computer vision \cite{bib:sigcomm2020:kim, bib:mobicom2019:baig, bib:hotcloud2019:wang, bib:SIGCOMM2016:sun, bib:mobicom2019:lee, bib:tmc2020:yi}, no solution fulfills the above requirements simultaneously, which motives the design of \sysname. 
\rev{Note that we focus on enhancement of RGB videos taken in low light. 
We leave extreme low light enhancement to future work because such cases often require videos stored in raw formats \cite{bib:ICCV19:Chen, bib:ICCV19:Jiang, bib:TPAMI21:Li}, which are not pervasively supported in mobile and embedded platforms.}

\subsection{Solution Overview}
\label{subsec:solution}
Our solution, \sysname, is an energy-aware low-light video enhancement scheme that achieves near real-time video processing with competitive enhanced visual quality against the state-of-the-arts \cite{bib:CVPR20:Guo, bib:TPAMI21:Li2, bib:BMVC18:Lv}, while allowing runtime enhancement behavior adaptation to platform-imposed energy budgets.
This is achieved by decoupling the requirements in \secref{subsubsec:requirements} with two functional modules: \textit{fast low-light video enhancement} and \textit{energy-aware adaptation controller}.
\begin{itemize}
    \item \textit{Fast Low-Light Video Enhancement (\secref{sec:algorithm})}.
    This module aims at near real-time low-light video enhancement on mobile platforms without compromising enhancement quality.
    Prior low-light enhancement schemes are unfit for our purposes because they either induce unacceptable latency \cite{bib:BMVC18:Lv, bib:ICCV19:Jiang} or are not optimized for videos \cite{bib:CVPR20:Guo, bib:TPAMI21:Li2} \eg suffering from the flicking problem \cite{bib:TPAMI21:Li}.
    In \sysname, we design a new low-light video enhancement model by removing the latency bottleneck of a state-of-the-art low-light image enhancement model \cite{bib:CVPR20:Guo} and enforcing temporal consistency between frames.
    \item \textit{Energy-aware Adaption Controller (\secref{sec:controller})}.
    This module dynamically adjusts the frame resolution and computation reuse of the video enhancement model to maximize the enhanced video quality while minimizing the energy consumption.
    Such adaptation is non-trivial due to the multi-objective nature of the optimization problem and the challenge to estimate energy consumption at runtime.
    In \sysname, we develop an optimizer to heuristically solve the multi-objective optimization problem and propose a runtime energy profiler to estimate the energy cost of the video enhancement model at runtime.
    This module also contains a resource monitor and analyzer to automate the adaptation process.
\end{itemize}

\begin{figure}[t]
  \centering
  \includegraphics[width=.85\textwidth]{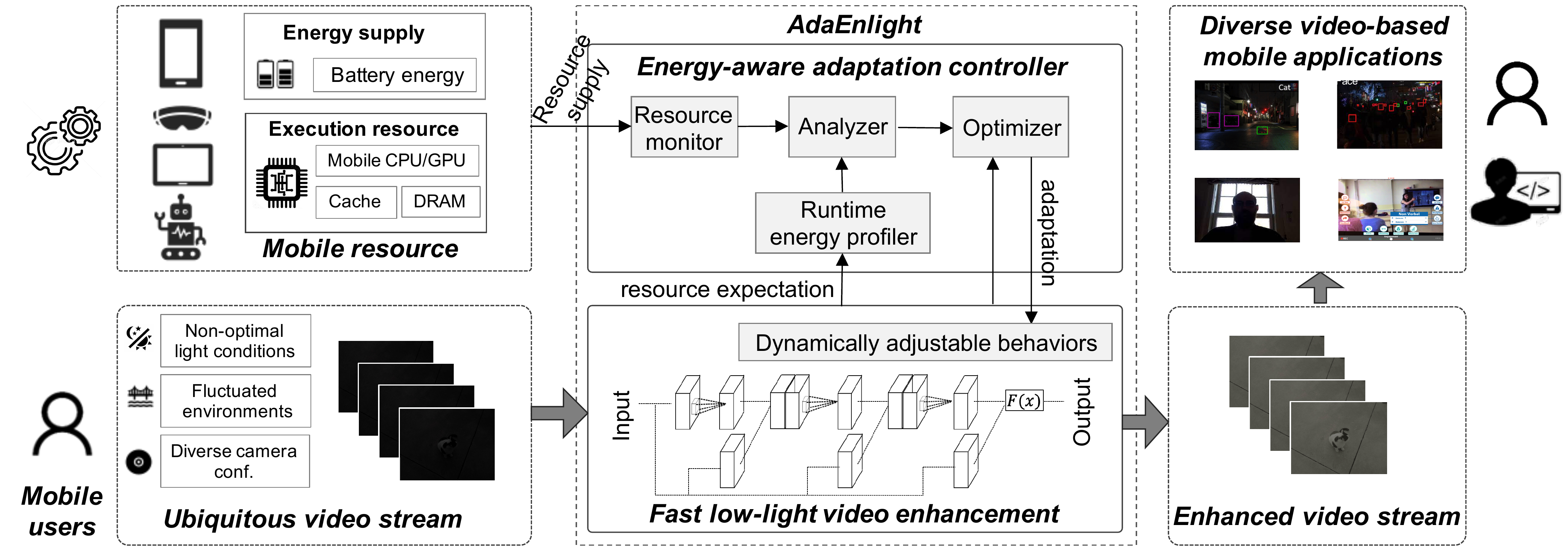}
  \caption{Workflow of \sysname.}
  \label{fig:overview}
  \vspace{-8mm}
\end{figure}

\figref{fig:overview} shows the workflow of \sysname.
The ubiquitous video streams captured by a camera-embedded mobile device are fed into \sysnameposs fast low-light video enhancement module for visual quality enhancement.
Meanwhile, \sysnameposs energy-aware adaptation module adds an extra control flow to configure the video enhancement module's parameters according to the dynamic energy budgets of the mobile device.

% \lsc{
% The mobile user captures the raw video streams using the camera-embedded mobile devices, then \sysname pre-processes them before delivering them to the video applications.
% %
% The captured video is fed into \sysnameposs fast low-light video enhancement module for visual quality enhancement. 
% %
% And the \sysnameposs energy-aware adaptation module adds an extra control flow to tune the video enhancement module's parameters for satisfying the dynamic energy budgets.
% %
% The input of the control flow is the platform's resource availability, and the output is the suitable parameter configurations of the video enhancement module.
% }

We implement \sysname as a compact video pre-processing middleware.
It transparently augments the low-light video streams into high-quality ones for downstream on-device video applications such as facial recognition and object detection.
As a middleware, \sysname can resolve the application diversity and the system complexity and effectively deal with multiple co-existing demands and constraints from different aspects (\eg visual quality, latency, energy).
Application developers do not have to initiate the video enhancement or be involved in the details of system monitoring and analysis.

% It can transform the raw (\eg low-light) video streams into high-quality ones for benefiting subsequent on-device video applications, such as facial recognition and object detection.
% %
% As a system external to the video application, \sysname can resolve the application diversity and the system complexity and effectively deal with multiple co-existing demands and constraints from different aspects (\eg visual quality, latency, energy).
% %
% Application developers do not have to initiate the video enhancement or be involved in the details of system monitoring and analysis.
%lsc:中间件思想这里先写上吧，不影响。
% \TODO{... or remove description about middleware if not much to say here} %lsc:开源与否，咱们投稿时再看代码完整情况。

%Next, we explain the fast low-light video enhancement and the energy-aware adaptation controller in sequel.

\section{Fast Low-light Video Enhancement}
\label{sec:algorithm}
This section presents \sysname's fast low-light video enhancement model.
It is built upon prior curve Based image enhancement solutions (\secref{subsec:algo:primer}).
To satisfy our special requirements, \ie near \textit{real-time} processing and \textit{video} enhancement, we exploit Gamma correction based curve to accelerate the enhancement speed (\secref{subsec:algo:gamma}) and extend the image-specific solution to videos by incorporating temporal consistency between frames (\secref{subsec:algo:temporal}). 
The detailed implementations are in \secref{subsec:algo:together}.

\subsection{Primer on Curve Based Image Enhancement}
\label{subsec:algo:primer}
Low-light image enhancement takes a low-light image as input, and outputs a normal-light, high-quality image \cite{bib:TPAMI21:Li}. 
We base our design upon Zero-DCE \cite{bib:CVPR20:Guo}, a recent zero-shot learning low-light image enhancement scheme via deep curvature estimation.
The reasons are as follows.
% \begin{itemize}
%   \item 
\textit{(i)}. We choose the \textit{zero-shot learning} based solution to eliminate the need for paired training data, \ie videos of the same scene taken in dim and normal illumination. This is desired because access to such paired videos is limited in real-world mobile applications with high variability.
    % \item 
\textit{(ii)}. We select the \textit{curve-based model} for it yields competitive image enhancement performance despite its lightweight architecture \cite{bib:TPAMI21:Li2}, which holds potential for execution on resource-constrained mobile devices.
% \end{itemize}
As next, we provide a brief review on Zero-DCE \cite{bib:CVPR20:Guo} and discuss its limitations for fast video enhancement.

\subsubsection{Principles of Zero-DCE}
The design of Zero-DCE \cite{bib:CVPR20:Guo} follows the curve adjustment in photo editing software.
It derives an \textit{image-specific curve} that maps a low-light image to its enhanced version.
The curve is defined as a high-order pixel-wise function learned by a deep neural network, where its loss function is designed as a set of differentiable non-reference losses to enable zero-shot learning. 
%Specifically, 
The curve function is defined as:
\begin{equation}\label{equ:dce:curve}
    E_{n}(\mathbf{x}) = E_{n-1}(\mathbf{x}) + A_{n}(\mathbf{x})\cdot E_{n-1}(\mathbf{x}) \cdot \left(1 - E_{n-1}(\mathbf{x})\right)
\end{equation}
where $\mathbf{x}$ is the pixel coordinates of the input image $I(\mathbf{x})$, $E_{n}(\mathbf{x})$ is the enhanced output image (note that $E_{0}(\mathbf{x})=I(\mathbf{x})$), $n$ is the number of iteration to control the curvature, $A_{n}(\mathbf{x})$ is a parameter map of the same size as $I(\mathbf{x})$, where each element in $A_{n}(\mathbf{x})$ is a trainable curve parameter in $[-1,1]$.
Each pixel in $I(\mathbf{x})$ is normalized to $[0,1]$ and the enhancement is performed separately to three RGB channels.
The curve function $E_{n}(\mathbf{x})$ is learned via a neural network with a loss function that consists of spatial consistency, exposure control, color constancy, and illumination smoothness. 
\rev{The spatial consistency loss $L_{spa}$ encourages spatial coherence of the enhanced image. 
The exposure loss $L_{exp}$ controls the exposure level of enhanced image. 
The color loss $L_{col}$ corrects the potential color deviations. 
The illumination smoothness loss $L_{tvA}$ preserves monotonicity relations between neighboring pixels.
More details can be found in \cite{bib:CVPR20:Guo}.}
%shows after training with such a loss function, the enhance model can enhance the image or single video frame well enough. }

\subsubsection{Limitations of Zero-DCE}
\label{subsubsec:limitation}
\begin{figure*}[t]
  \centering
   \subfloat[Zero-DCE]{
   \includegraphics[height=0.1\textwidth]{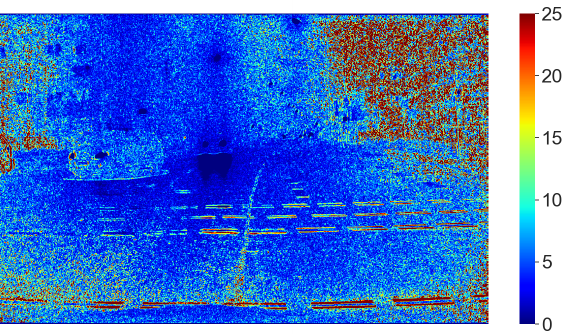}}
   \hspace{15pt}
   %\hfill
   \subfloat[Zero-DCE++]{
   \includegraphics[height=0.1\textwidth]{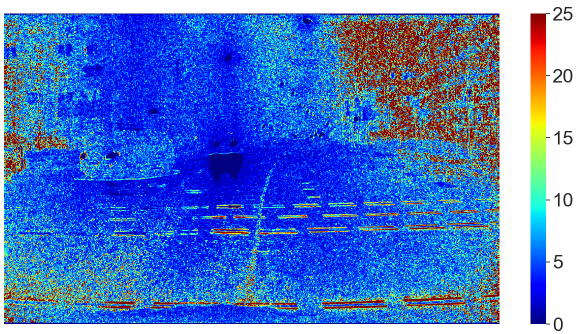}}
   %\hfill
  %\hfill
  \hspace{15pt}
  \subfloat[\sysname]{
  \includegraphics[height=0.1\textwidth]{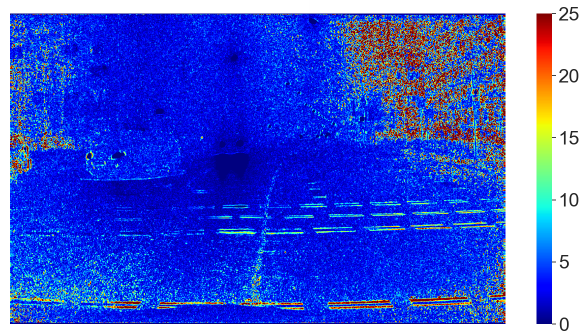}}
\caption{Illustration of the flicking problems when applying low-light image enhancement schemes, \eg (a) Zero-DCE \cite{bib:CVPR20:Guo} and (b) Zero-DCE++ \cite{bib:TPAMI21:Li2} to videos. The figures show the difference between two consecutive frames. The greater the difference, the more serious the flicking problem is. In comparison, (c) shows the low frame difference of \sysname.}
%\TODO{add color bar to the figure}}
\label{fig:flicking}
\vspace{-5mm}
\end{figure*}

Despite being \rev{one of the most \textit{lightweight image enhancement}} scheme, Zero-DCE fails to support \textit{fast video enhancement} on resource-constrained mobile devices for the following reasons.
% \begin{itemize}
%     \item 
    %%%%%%%%%%
    % 实时处理需求: 24帧，30帧(这两个标准使用较多,倾向24帧)
    % 测试帧尺寸: 270*480*3
    % 测试平台: Raspberry 4Pi 4GB(硬件), NCNN(软件)
    % Zero-DCE 11.743S
    % Zero-DCE++ 0.193S
    % Ours 0.05S
    %%%%%%%%%%
\textit{(i)}, the iterative procedure in \ie \equref{equ:dce:curve}, for image enhancement,  incurs \textit{notable delay} that prohibits real-time video enhancement.
For example, Zero-DCE takes an average of $11.74s$ to enhance a $270 \times 480$ RGB image on a Raspberry Pi 4B platform (detailed setups in \secref{sec_exp_setup}).
The follow-up Zero-DCE++ \cite{bib:TPAMI21:Li2} accelerates Zero-DCE by adopting depth-wise separable convolutions and reusing the curve parameter maps $A_{n}(\mathbf{x})$.
However, Zero-DCE++ still induces $0.19s$  delay, \ie only $5$ frames per second.
The main bottleneck lies in the iterative procedure, where $n$ is empirically optimized to $8$ in both Zero-DCE and Zero-DCE++.
    % \item
\textit{(ii)}, directly applying image enhancement schemes to videos leads to \textit{flicking} between frames \cite{bib:ICCV19:Chen, bib:CVPR21:Zhang}. 
\figref{fig:flicking} shows the difference between two consecutive images enhanced by Zero-DCE and Zero-DCE++.
There is notable differences between adjacent frames (up to 9.192) and thereby results in flicking in videos.
% \end{itemize}
%%%%%%%%%%
% 采用相邻帧的差分热力图
% 如See Motion in the Dark中的Figure 9
%%%%%%%%%%
These limitations motivate us to rethink the design for the fast video enhancement in two aspects, as will be explained in \secref{subsec:algo:gamma} and \secref{subsec:algo:temporal}.

\subsection{Non-iterative Frame Enhancement with Gamma Correction-based Curve}
\label{subsec:algo:gamma}
%慕神：我把这节改了一些。因为后面note 2提到了迭代过程，和标题non-iterative有冲突，怕读者疑惑。
As previously mentioned (\secref{subsubsec:limitation}), the latency bottleneck of curve-based image enhancement~\cite{bib:CVPR20:Guo, bib:TPAMI21:Li2} 
%such as Zero-DCE \cite{bib:CVPR20:Guo} and Zero-DCE++ \cite{bib:TPAMI21:Li2} 
lies in the iterative enhancement process.
We observe that the iterative procedure may be eliminated if a non-iterative curve function \rev{with higher non-linearity} is adopted.
Therefore, we propose to replace the traditional \rev{quadratic} curve in \equref{equ:dce:curve} by an \rev{exponential} curve to induce higher-order non-linearity and smaller latency.
Specifically, we propose the following non-iterative image enhancement scheme. 
\begin{equation}\label{equ:ada:curve}
    E(\mathbf{x}) = I(\mathbf{x})^{\Gamma(\mathbf{x})}
\end{equation}
where $I(\mathbf{x})$ and $E(\mathbf{x})$ denote the input and enhanced image, respectively.
$\Gamma(\mathbf{x})$ are pixel-wise trainable parameters learned from a neural network.
The detailed training methodology of $\Gamma(\mathbf{x})$ are deferred to \secref{subsec:algo:together}.
%
%\fakeparagraph{Discussions}
We make two detail notes on our design of \equref{equ:ada:curve}.
\begin{itemize}
    \item 
    The exponential curve function in \equref{equ:ada:curve} is inspired by the Gamma correction, which is typically used to flexibly adjust the luminance or tristimulus in images \cite{bib:JIVP16:Rahman}.
    Our novelty is to apply pixel-wise parameters $\Gamma(\mathbf{x})$ for fine-grained adjustment, which is learned from a neural network.
    In contrast, conventional Gamma correction adopts the same coefficient $\Gamma$ for the entire image, which tends to introduce artifacts and color deviations to the enhanced images \cite{bib:TPAMI21:Li}.
    %In contract, we apply pixel-wise parameters $\Gamma(\mathbf{x})$ for fine-grained adjustment learned from a neural network.
    \item
    One may want to design an iterative enhancement process by rewriting \equref{equ:ada:curve} as
    $E_n(\mathbf{x}) = E_{n-1}(\mathbf{x})^{\Gamma_n(\mathbf{x})}$, where $E_{0}(\mathbf{x})=I(\mathbf{x})$.
    We exclude this option because our experiments show that the enhanced visual quality does not notably increase with more iterations (see \secref{subsec:exp_ite}).
    %d in \figref{fig:iterations}.
\end{itemize}
To summarize, the learned pixel-wise Gamma correction-based curve in \equref{equ:ada:curve} tends to deliver high-quality visual enhancement while avoiding the long latency. 
We defer the quantitative comparisons with the traditional iterative enhancement scheme to \secref{subsec:exp:bench}.
%Therefore, \sysname can use the proposed  curve to boost the efficiency and speed of visual enhancement for each video frame.}

\subsection{Incorporating Temporal Consistency Loss for Video Enhancement}
\label{subsec:algo:temporal}
As mentioned in \secref{subsec:algo:primer}, naively adopting image enhancement schemes to videos \rev{collected by mobile devices} causes flicking.
A remedy is to account for temporal consistency between frames when designing the loss function to train the enhancement curve \ie \equref{equ:ada:curve}.
Given two enhanced frames $E_{t+1}(\mathbf{x})$ and $E_{t}(\mathbf{x})$ at timestamp $t+1$ and $t$, we consider their temporal consistency with the following losses. 
\begin{itemize}
    \item \textit{Exposure Consistency Loss}:
    We define the exposure consistency loss between two enhanced frames as
    \rev{
        \begin{equation}\label{equ:ada:exposure}
            \mathcal{L}_{exp, temporal} = \sum_{x}\left|\sum_{i=r,g,b} E_{t+1}^{i}(\mathbf{x}) - \sum_{i=r,g,b} E_{t}^{i}(\mathbf{x})\right|
        \end{equation}
    }
    \item \textit{Color Consistency Loss}:
    We define the color consistency loss between two enhanced frames as
    \rev{
        \begin{equation}\label{equ:ada:color}
            \mathcal{L}_{col, temporal} = \sum_{x}\sum_{k=r,g,b} \left|\frac{E_{t+1}^{k}(\mathbf{x})+c}{\sum_{i=r,g,b} E_{t+1}^{i}(\mathbf{x}_i)+c} - \frac{E_{t}^{k}(\mathbf{x})+c}{\sum_{i=r,g,b} E_{t}^{i}(\mathbf{x}_i)+c}\right|
        \end{equation}
    }
\end{itemize}
where $E^{i}(\mathbf{x})$ denotes the enhanced image of one channel, where $i=r,g,b$, which represents the RGB channels.
\rev{$c$ is a delta value empirically set to 0.0001 to avoid division by zero in case of dark pixels}.
The exposure consistency loss sums over the RGB channels to avoid the impact of colors when calculating the temporal consistency.
The color consistency loss, in contrast, is divided by the exposure to mitigate the impact of pixel-wise exposure when calculating the temporal consistency.

\rev{Note that to avoid the influence of motion between frames, we employ the Gunnar-Farneback optical flow \cite{bib:farneback2003two}, a dense optical flow alignment method as a pre-processing step.
Specifically, we calculate motions between the two frames (frame $t+1$ and frame $t$) and get the motion data of each pixel of frame $t$, denoted as $(dx,dy)$. 
Then we modify the location of each pixel in frame $t$ as $(x+dx,y+dy)$.
The resulting fame as named as the aligned frame $t$. 
Finally, we calculate the absolute difference between the aligned frame $t$ and frame $t+1$.}
% 光流对齐叙述

%%%%%%%%%
% 在exposure consistency loss中先在RGB维度上求和。再求相邻帧之间的差异。此时屏蔽了色彩的影响.
% 在color consistency loss中，先将RGB通道分别除以像素点亮度，屏蔽像素点亮度的影响。
% 如对于两个像素点：(50,50,50)和(75,75,75)有着相似的颜色(灰色和更明亮的灰色)，但其分布在相邻帧中会体现为明暗闪烁的视觉现象
% 如对于两个像素点：(50,100,150)和(150,100,50)有着相似的亮度，但其颜色存在差异。其分布在相邻帧中会体现为色彩闪烁的视觉现象。
%%%%%%%%%
% \figref{fig:temporal} shows the differences in exposure and color between two adjacent frames before enhancement, and enhanced without/with considering temporal consistency.
% As is shown, there is considerable noise in exposure and color if temporal consistency is ignored, whereas the noise is notably suppressed if the exposure and color consistency losses are considered.

% \begin{figure}[t]
%   \centering
%   \includegraphics[width=.50\textwidth]{image/todo.pdf}
%   \caption{Differences in exposure and color between two adjacent images without enhancement, enhanced by Zero-DCE \cite{bib:CVPR20:Guo}, Zero-DCE++ \cite{bib:TPAMI21:Li2}, and our method (considering exposure and color consistency losses).}
%   \label{fig:temporal}
% \end{figure}

\subsection{Putting It Together}
\label{subsec:algo:together}
In short, our fast low-light video enhancement module is a deep curve-based enhancement scheme with a learned pixel-wise Gamma correction based curve \ie \equref{equ:ada:curve}.
The curve parameters are trained by considering both the conventional image enhancement loss as well as the temporal consistency loss \ie \equref{equ:ada:exposure} and \equref{equ:ada:color}.
Below we explain the detailed model architecture, its training strategy and how it is designed to trade off between video enhancement quality and platform-imposed requirements at runtime.

\subsubsection{Model Architecture}
\label{subsubsec:algo:model}
\begin{figure}[t]
  \centering
  \includegraphics[width=.85\textwidth]{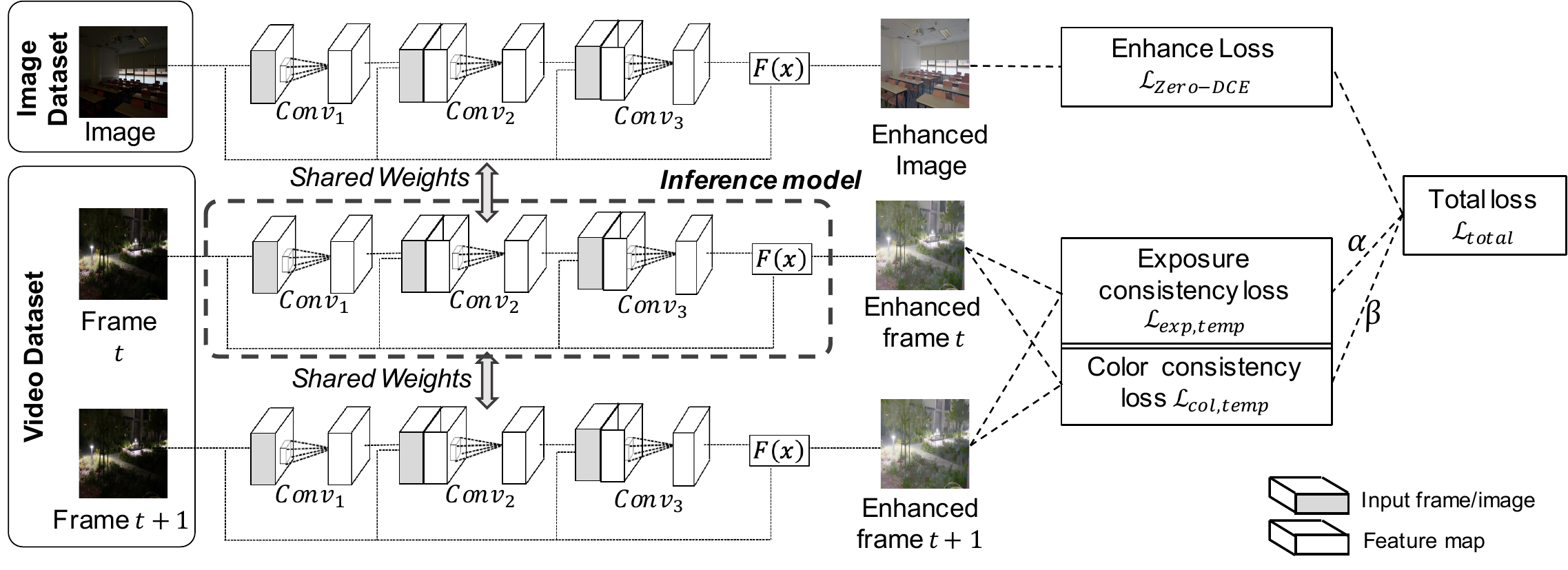}
  \caption{Model architecture and training scheme to learn the pixel-wise curve parameters in \equref{equ:ada:curve}. The model inside the  dashed box is used for inference.}
  \label{fig:nn}
  \vspace{-5mm}
\end{figure}

We apply a U-Net \cite{bib:MICCAI15:Ronneberger} to learn the parameters $\Gamma_n(\mathbf{x})$ in \equref{equ:ada:curve}.
For lower latency, we decrease the number of layers and the channels of the original U-Net.
We also change the contracting paths in the original U-Net to dense connections as DenseNet \cite{bib:CVPR17:Huang} for computation reuse (see \secref{subsubsec:algo:behaviour}).
\figref{fig:nn} shows the model architecture. 
\tabref{tab:nn} lists the detailed hyperparameters.
The hyperparameters are empirically tuned to trade off between enhancement quality and latency. 
The ablation studies on hyperparameters are in \secref{sec:exp_reuse}.
%为了在任意一个卷积块均可以进行输入原始图像以达到复用前层计算结果的效果，我们将U-Net中的跨层连接结构改为了密集连接结构

%%%%%%%%%%
% 跨层连接,跳跃连接在U-Net中的描述：a concatenation with the correspondingly cropped feature map from the contracting path,
% U-Net 网址https://arxiv.org/abs/1505.04597
% 密集连接,DenseNet https://arxiv.org/abs/1608.06993 将图像输入以密集连接的方式连接至模型的每一层
%%%%%%%%%%

% \begin{table}[t]
% \caption{Detailed hyperparameters of the neural network.} \label{tab:nn}
%     \centering
%     \begin{tabular}{cc}
%     \toprule
%     Layer & Parameters\\
%     \midrule
%     \multirow{2}{*}{Conv1} & K=1, Cin=3 Cout=1, S=1, G=1, P=0 \\
%     & K=3, Cin=1, Cout=1, S=1, G=1, P=0 \\
%     \midrule
%     \multirow{2}{*}{Conv2} & K=1, Cin=4, Cout=1, S=1, G=1, P=0 \\
%     & K=3, Cin=1, Cout=1, S=1, G=1, P=1 \\
%     \midrule
%     Conv3 & K=1, Cin=4, Cout=3, S=1, G=1, P=0 \\
%     \bottomrule
% \end{tabular}
% \end{table}

\begin{table}[t]
\centering
\scriptsize
\caption{Detailed hyperparameters of the neural network.} \label{tab:nn}
\vspace{-3mm}
\begin{tabular}{|c|c|c|c|c|c|}
\hline
\textbf{Layer} & \textbf{Parameters} & \textbf{Layer} & \textbf{Parameters} & \textbf{Layer} & \textbf{Parameters} \\ \hline
Conv\_1 & \begin{tabular}[c]{@{}c@{}}K=1, Cin=3, Cout=1, S=1, G=1, P=0 \\ K=3, Cin=1, Cout=1, S=1, G=1, P=0\end{tabular} & Conv\_2 & \begin{tabular}[c]{@{}c@{}}K=1, Cin=4, Cout=1, S=1, G=1, P=0 \\ K=3, Cin=1, Cout=1, S=1, G=1, P=1\end{tabular} & Conv\_2 & \begin{tabular}[c]{@{}c@{}}K=1, Cin=4, Cout=3\\ S=1, G=1, P=0\end{tabular} \\ \hline
\end{tabular}
\end{table}

We combine the loss function from Zero-DCE \cite{bib:CVPR20:Guo}, \ie image enhancement loss, with the two temporal consistency losses defined in \equref{equ:ada:exposure} and \equref{equ:ada:color} as the final loss function.
\begin{equation}\label{equ:ada:loss}
    \mathcal{L}_{total} = \mathcal{L}_{Zero-DCE} + \alpha \mathcal{L}_{exp, temporal} + \beta \mathcal{L}_{col, temporal}
\end{equation}
where $\alpha$ and $\beta$ can be empirically tuned.

\subsubsection{Training Strategy}
\label{subsubsec:algo:training}
Due to the lack of video datasets with evenly distributed exposure levels, it is challenging to train a network that minimizes \equref{equ:ada:loss}.
Therefore, we propose to learn the curve by training the image enhancement loss and the temporal consistency losses in a siamese mode as in \cite{bib:ICCV19:Chen, bib:CVPR21:Zhang}.
\figref{fig:nn} shows the training scheme. 
In each pass, we input one low-light image from the image dataset and two adjacent low-light video frames from the video dataset into the neural network in a siamese way \ie three neural networks share the same weights.
The low-light image will be used to calculate the image enhancement loss, whereas the two frames will be used to compute the two temporal consistency losses.

%%%%%%%%%%
% 孪生网络训练方式的具体含义：在训练时，我们将一张低照度图像和两张相邻低照度视频帧以孪生网络的方式输入至网络。我们计算增强后的低照度图像的零参考损失作为图像增强损失，计算相邻帧之间亮度、色彩差异作为视频增强的时域一致性损失。
% Loli-Phone的引文Low-Light Image and Video Enhancement Using Deep Learning: A Survey，网址：https://www.computer.org/csdl/journal/tp/5555/01/09609683/1yoxCAHSmWs

% 训练时确实是缩放到256*256*3，训练和测试的图像尺寸不同在低照度增强领域是较为常见的

% epochs我采用了50
%%%%%%%%%%

%尽管我们的损失函数设计均可实现无参考训练，但在训练过程中仍依赖于不同曝光等级的图像数据作为图像增强部分损失函数的驱动。而目前的视频数据集中的视频帧曝光等级分布不均匀，视频帧的质量较差，因此难以直接完全采用视频数据集进行训练。
%%%%%%%%%%
% The network takes a single frame as input.For training, two frames from a static sequence in DRV are sampled at random and are fed to the network in siamese mode.
% See Motion in the Dark中对采用孪生网络训练的描述
% SIGNATURE VERIFICATION USING A “SIAMESE” TIME DELAY NEURAL NETWORK
% See Motion in the Dark 关于孪生网络引用的文章
%  Given image pairs of original image and warped image, we train our network in a siamese way in which we feed them one by one to the network.
% Learning Temporal Consistency for Low Light Video Enhancement from Single Images 2021CVPR 中对孪生网络训练的描述

% 本网络以单帧图像作为输入.
%%%%%%%%%%
%为了解决该问题，我们采用了一个三孪生网络对我们的模型进行训练。

At inference time, the model takes a low-light video frame as input, and generates an enhanced frame in almost real-time ($0.05$ second delay or $20$ frames per second for $270 \times 480$ RGB images on the Raspberry Pi 4 (4GB) platform.
More quantitative evaluations on enhancement quality and latency are in \secref{subsec:exp_compare}.
    
\subsubsection{Towards Runtime Adaptation}
\label{subsubsec:algo:behaviour}
\begin{figure}[t]
  \centering
   \subfloat[]{\label{fig:behaviour:frame}
   \includegraphics[width=0.3\textwidth]{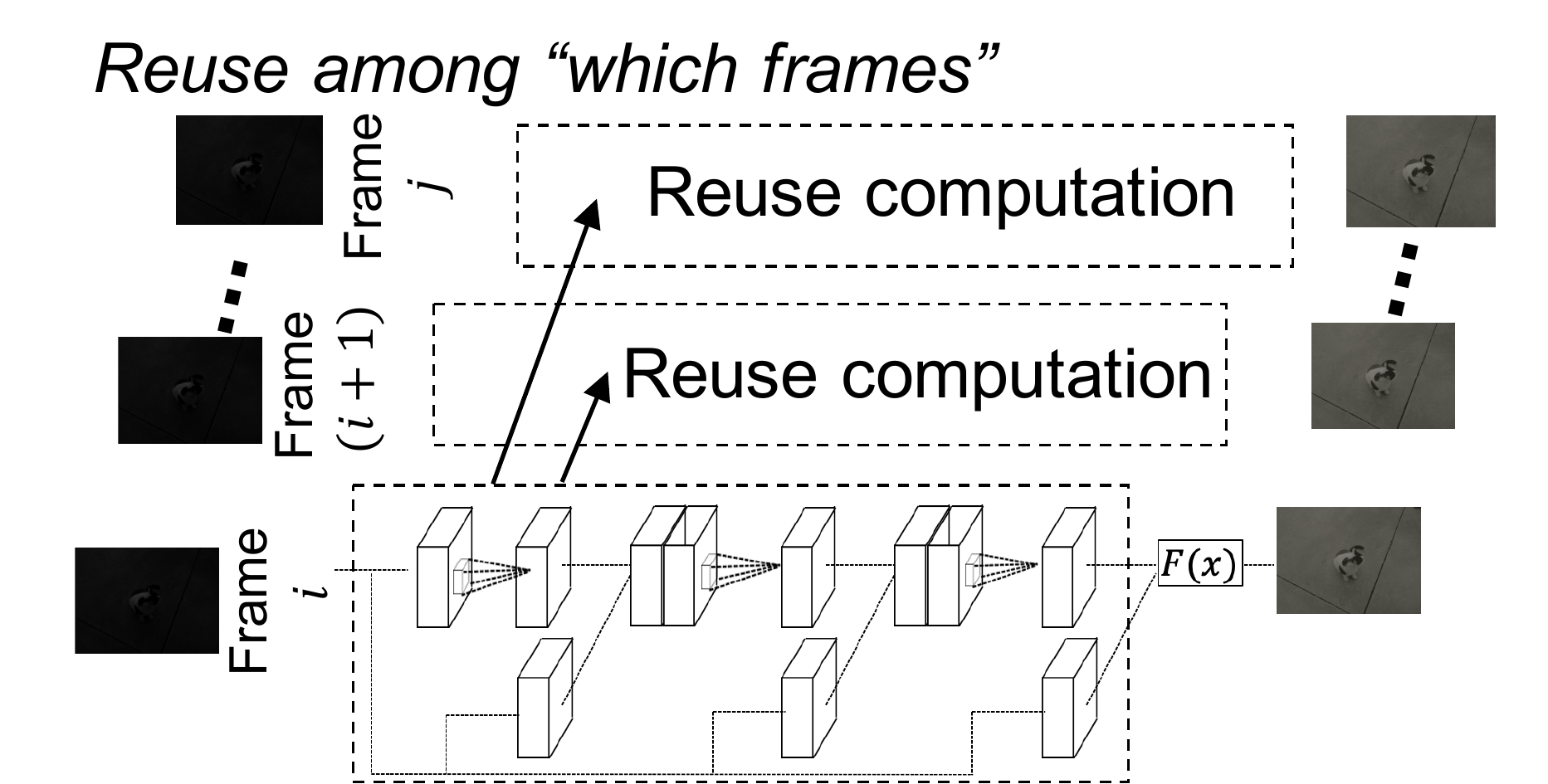}}
   \quad
   \subfloat[]{\label{fig:behaviour:layer}
   \includegraphics[width=0.3\textwidth]{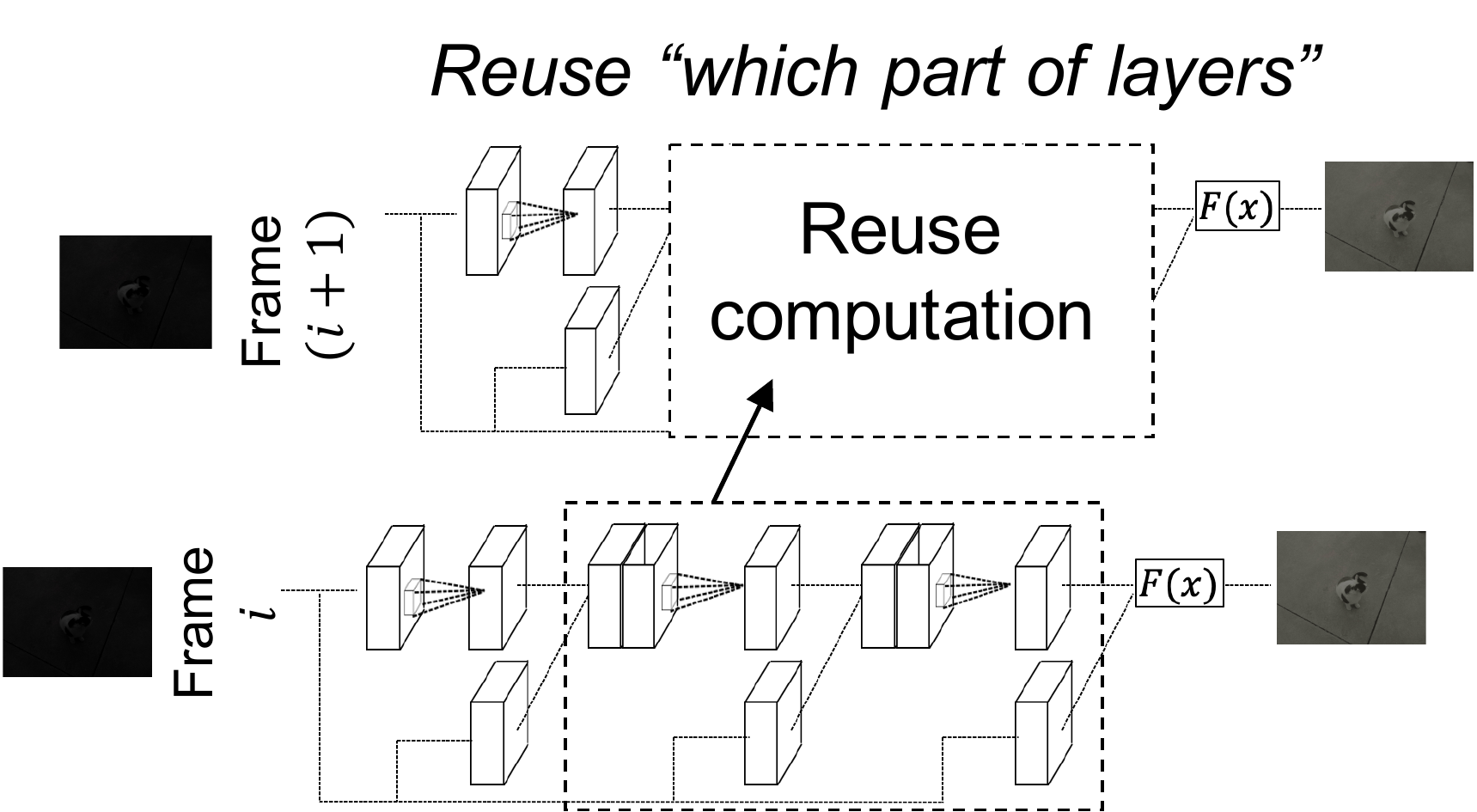}}
   \quad
   \subfloat[]{\label{fig:behaviour:resolution}
   \includegraphics[width=0.3\textwidth]{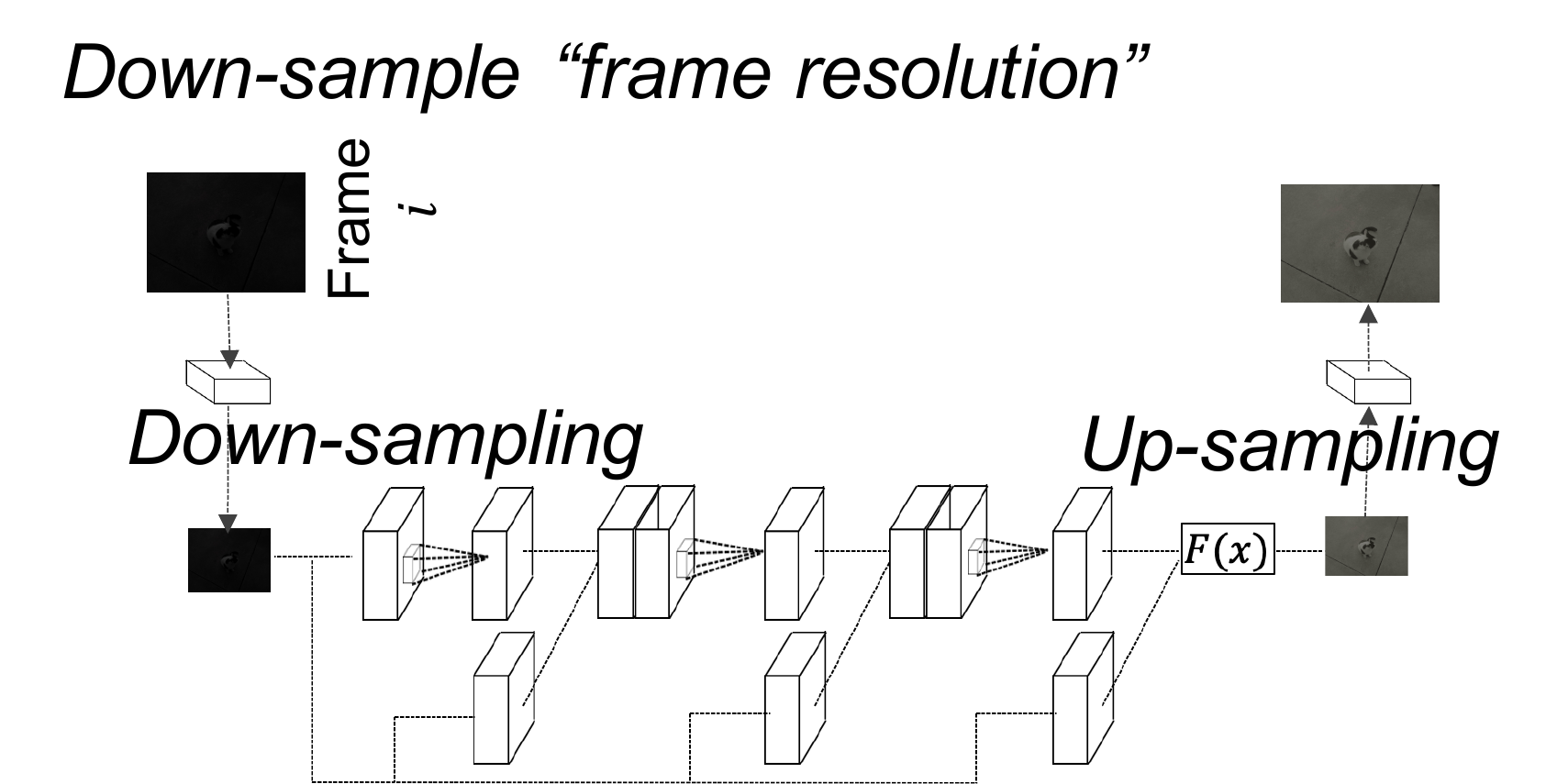}}
\caption{Illustrations of adjustable parameters of the video enhancement model: computation reuse as (a) frame level and (b) layer level; (c) down-sample the resolution of input frames.}
\label{fig:behaviour}
\vspace{-3mm}
\end{figure}

Although our low-light video enhancement module achieves near real-time performance, there are other requirements \eg energy, which the users may trade with enhancement quality.
Our video enhancement module supports such runtime adaptation by adjusting the parameters below.
\begin{itemize}
    \item \textit{Computation Reuse}.
    Recall that our video enhancement model consists of three convolutional layers. 
    Due to temporal similarities between adjacent frames, we can further reduce the computations by enforcing computation reuse during frame enhancement.
    We allow reuse at both the frame level (see \figref{fig:behaviour:frame}), \ie reuse the enhanced output of frame $t$ for frame $t+1$, and the layer level (see \figref{fig:behaviour:layer}), \ie reuse the intermediate results and skip some layers during enhancement.
    \item \textit{Frame Resolution}.
    Down-sampling the resolution of the input frame naturally reduces the computation of the video enhancement process (see \figref{fig:behaviour:resolution}).  
\end{itemize}
We utilize these parameters to trade off between enhancement quality and energy at runtime, as explained next.

% To realize the prompt reaction to the time-dependent context changing, we also design  the
% 
% \textit{dynamically adjustable behaviours} within the video enhancement model which are effective for tuning the trade-off between visual quality and energy demands.
% %
% These behaviors are required to be retraining-free, so that we can directly scale them during the inference time.
% %

%As shown in \figref{fig_parameter}, we present them as follows:
%\textit{(i) the coarse-grained temporal consistency parameter} is the updating frequency of inter-frame reusable computation, \ie the number of successive frames which reuse computations;
%
%\textit{(ii) the fine-grained temporal consistency parameter} is the architecture scale of inter-frame reusable computations, \ie the part of layers at which the computations are reused among successive frames;
%
%and \textit{(iii) the down-sampling rate} denotes the different sampling rate of the input frame, which re-samples the video frame at the new resolution and thereby scale the complexity of video enhancement process.
%
% As we will show in Sec. \ref{}, all of these three parameters can effectively tune the trading of enhancement quality for resource consumption at run-time.
% %
% And we concentrate on the above three adjustable decision variables to make the adaptation decision (Sec. \ref{sec_optimizer}).
\section{Energy-aware Adaptation Controller}
\label{sec:controller}
%Controller设计

This section presents the \sysnameposs energy-aware adaptation controller to dynamically adjust the above mentioned video enhancement model's behaviors for energy conservation. 
The focus of this controller is to optimize the trade-off between energy and the video quality, because the above design (\secref{subsec:algo:gamma}) has already guaranteed the near real-time video processing. 
We first introduce the design of the overall controller (\secref{subsec:design}), then the runtime energy profiler (\secref{subsec:profiler}), and finally the auxiliary modules to automate the control loop (\secref{subsec:loop}).

\subsection{Controller Design}
\label{subsec:design}
At a high level, the energy-aware adaptation controller of \sysname is a runtime heuristic optimizer for a multi-objective optimization problem, as formulated below.
\subsubsection{Formulation}
\label{subsubsec:formulation}
Mathematically, the controller aims to continuously optimize two metrics, \ie the visual quality $Q_{video}$ of the output video, and the energy consumption $E_{enhancer}$ of the video enhancement model.
We formulate it as the following time-varying optimization problem.
\begin{eqnarray}\label{equ_opt}
    \mathop{arg max}_{\theta_l, \theta_f, \theta_d}  &  Q_{video} = q(\theta_l, \theta_f, \theta_d, video(t)) ,\,\,
    \mathop{arg min}_{\theta_l, \theta_f, \theta_d}  &  E_{enhancer} = e(\theta_l, \theta_f, \theta_d, resource(t)) 
%\nonumber 
%\\%\text{s.t.} & T_{enhancer} \leq T_{bgt}
\end{eqnarray}
where $\theta_l$, $\theta_f$, and $\theta_d$ are three dynamically adjustable parameters of the video enhancement model, \ie the amount of layer-level computation reuse, the amount of frame-level computation reuse, and the down-sampled video frame resolution (see \secref{subsubsec:algo:behaviour}).
It is intractable to obtain a closed-form solution to the above dynamic multi-objective optimization problems because dealing with two distinct yet related spaces, \ie \textit{objective space} and \textit{variable space}, is challenging.
Specifically, the \textit{objective space} consisting of the visual metric $Q_{video}$ and the energy metric $E_{enhancer}$.
Regarding the \textit{variable space} related to the objective performance, some variables (\ie $\theta_l$, $\theta_f$, and $\theta_d$) are available for objective optimization, we regard them as \textit{decision variables}.
Other variables (\eg the available execution resource $resource(t)$, the video stream $video(t)$) are imposed by the external context, which are time-varying but independent from the optimization objective variables.

\subsubsection{Optimizer}
\label{subsubsec:optimizer}
To solve the above problem, we propose a heuristic optimizer.
Specifically, we solve the original optimization problem in two stages, which correspond to two typical dynamic optimization problems~\cite{bib:TEC2004:farina}.
In the offline stage, we regard this problem as a static optimization problem and adopt existing evolutionary algorithm to find a widely distributed set of solutions and derive the Pareto-optimal front.
When looking for the Pareto frontier, we do not set any relative importance coefficients for multiple optimization objectives (\ie $Q$ and $E$), because we need to get the unbiased opinion from this set of solutions.
In the online stage, the optimal decision variables (\ie $\theta_l$, $\theta_f$, $\theta_d$) changes, whereas the optimal objective space (\ie $Q, E$) does not change, which is the Type 2 dynamic multiple-objective optimization problem~\cite{bib:TEC2004:farina}.
Therefore, we leverage an analytical hierarchy process to set the dynamic importance coefficients $\lambda$ of the different criteria.
Then we multiply the respective performance of each solution with that coefficients, to get a total score for each solution, \ie $\lambda E_{enhancer} +(1-\lambda)Q_{video}$, indicating which suits best the current wishes.
For example, if the the enhancement model's demand stray beyond resource levels, the controller re-selects the dynamic importance coefficients to pick a sole optimal solution (\ie $\theta_l, \theta_f, \theta_d$) from the Pareto frontier at runtime. 

\begin{figure}[t]
  \centering
  \includegraphics[width=.80\textwidth]{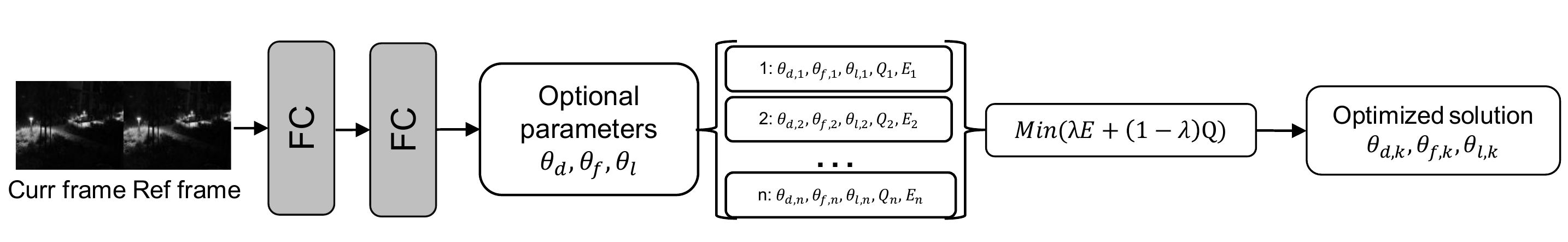}
  \caption{\rev{Workflow of the optimizer in the energy-aware adaption controller.}}
  \label{fig:optimizer}
  \vspace{-4mm}
\end{figure}
\subsection{Runtime Energy Profiler}
\label{subsec:profiler}
For the controller to function properly, it is essential to provide an accurate and timely estimate of $E_{enhancer}$ to the controller.
In \sysname, we propose an energy model for the video enhancement model that estimates its energy consumption given the adjustable parameters $\theta_l$, $\theta_f$, and $\theta_d$ at runtime.

However, the neighbor solutions on the Pareto frontier in the objective space are not always neighbors in the decision variable space, which makes the search of decision variable time-consuming.
In view of this challenge, we design an extra parametric neural network based regression module for building the map between the state (\ie the input video frame and the previous reference frame), decision variables (\ie $DO, FO, DR$), and the objective outcomes (\ie $Q, E$).
The network consists of two fully-connected layers.
The input of the neural network is \rev{the splicing matrix containing the pixel matrix of the input frame and the previous reference frame.
The output is the vector of $\theta_l$, $\theta_f$, $\theta_d$ as well as the objective outcomes (\ie $Q, E$).}
We collect $400$ samples to train this regression network.
\rev{At run-time, the network outputs all the optional parameters $\theta_{f}$, $\theta_{d}$, $\theta_{l}$ and the corresponding $Q, E$. 
The decision space built by $\theta_{f} \in \{0,1,2...10\}$, $\theta_{d} \in \{1,1/2,1/3\}$, $\theta_{l} \in \{0,1,2,3\}$ is limited.
Therefore, together with the dynamic importance coefficient $\lambda$, it is easy to get best computation reuse parameters by calculating $\lambda E_{enhancer} +(1-\lambda)Q_{video}$.
\figref{fig:optimizer} briefly describes the optimizer workflow.}

\subsubsection{Principles}
\begin{figure*}[t]
  \centering
   \subfloat[]{\label{fig:cache:a}
   \includegraphics[height=0.18\textwidth]{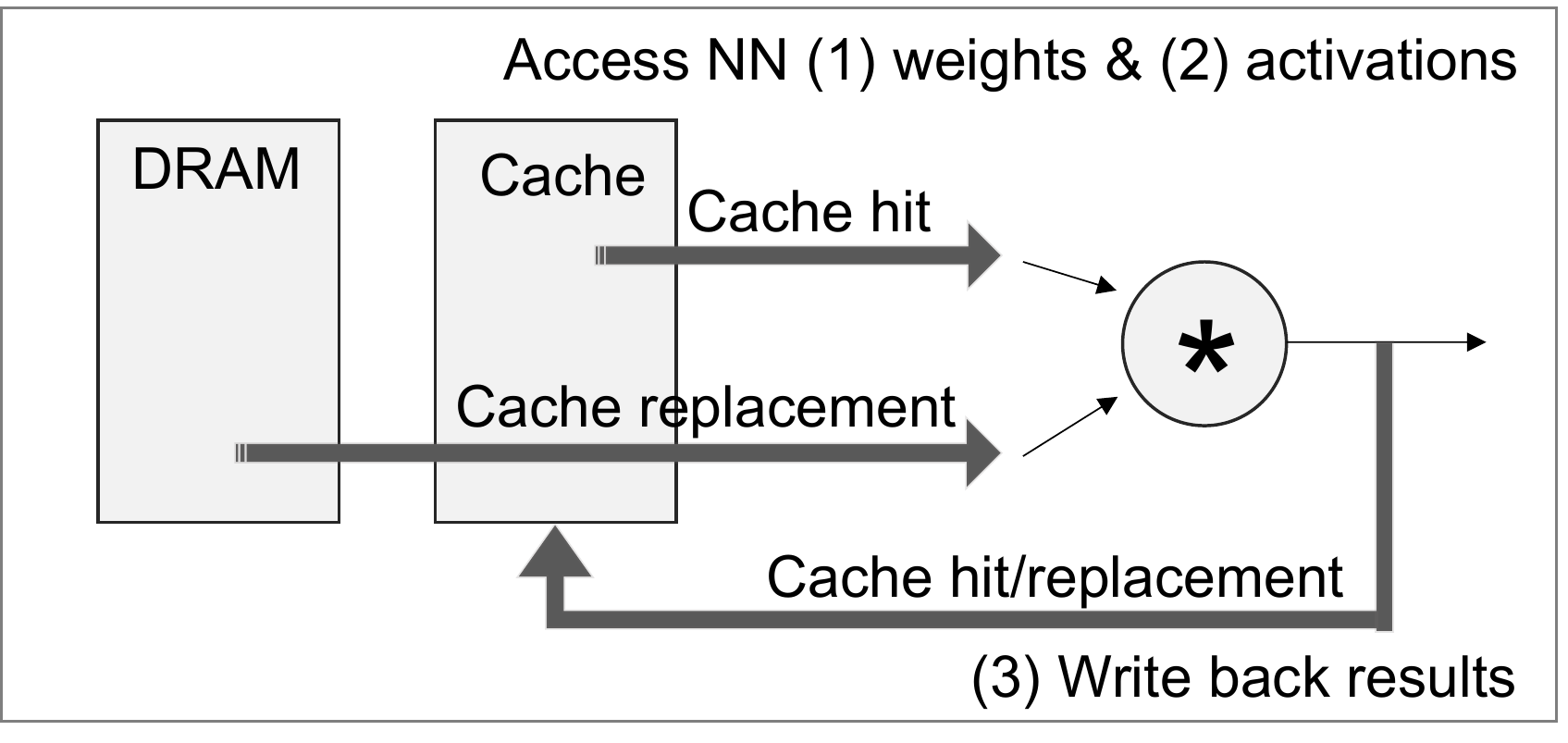}}
   \hspace{15pt}
   \subfloat[]{\label{fig:cache:b}
   \includegraphics[height=0.18\textwidth]{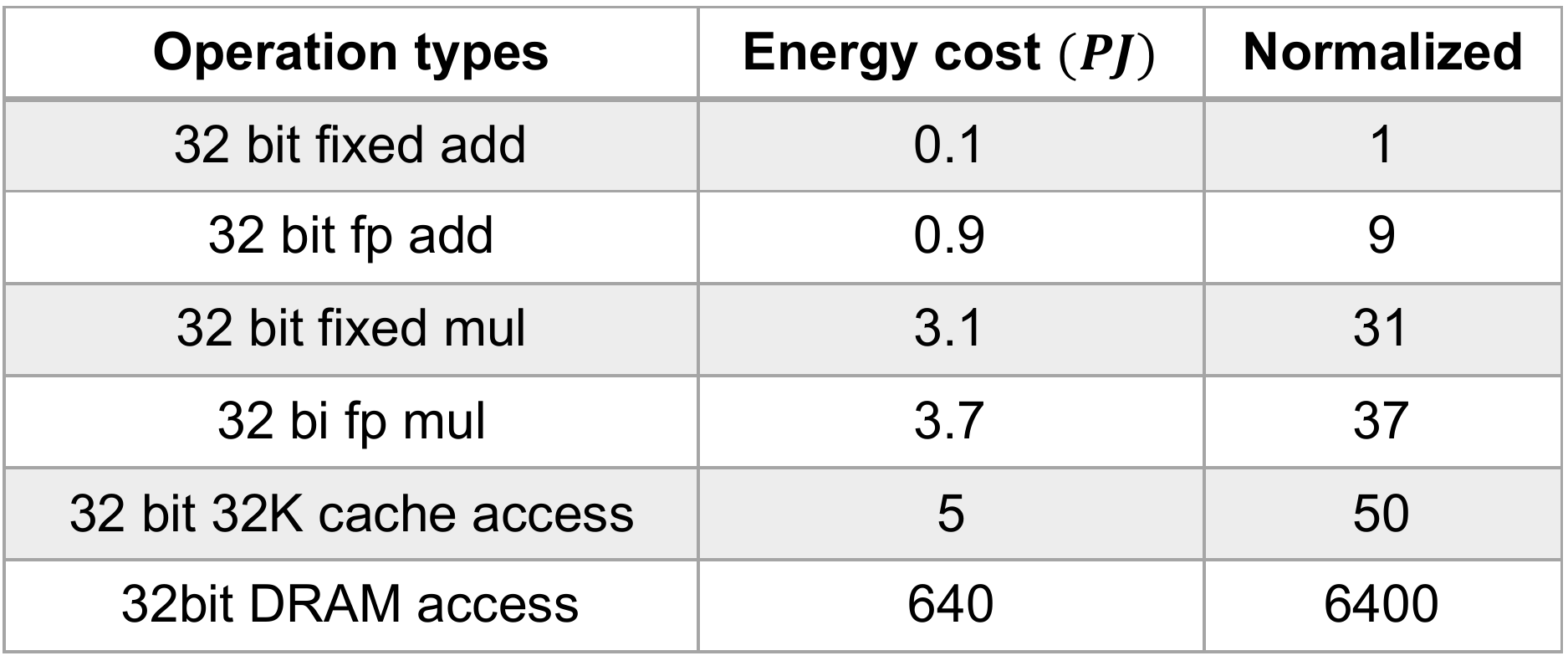}}
\caption{(a) Memory access and (b) energy cost of a convolutional operation on a mobile device with 45nm CMOS chip.}
\label{fig:cache}
\vspace{-3mm}
\end{figure*}

As discussed in \secref{subsec:related:energy}, it is challenging to model the energy consumption of a neural network because the energy cost is tightly coupled with the available execution resources of the given platform. 
The principles of our runtime energy profiler are two-fold.
\begin{itemize}
    \item 
    To estimate the energy cost of a neural network with dynamically adjustable hyperparameters, we decompose the network into layers and strive to model the energy cost on a layer basis.
    The layer-wise decomposition is reasonable because mobile devices with CPU/GPU tend to execute a neural network layer-by-layer due to limited resources \cite{bib:rtas2020:lee}.
    \item
    To estimate the energy cost in presence of time-varying execution resources, we convert the resource dynamics into a single parameter, the cache-hit-rate, which is directly measurable at runtime.
    This is because we empirically observe that the energy cost of memory accesses is dominated by the off-chip memory accesses, which can be estimated by the cache-hit-rate.
    \figref{fig:cache} shows an example of to execute a convolution operation on a mobile platform (\eg with a $45$nm CMOS ARM chip).
    There are three types of memory accesses for loading the weights and activations of a neural network and writing back results (see \figref{fig:cache:a}), and the DRAM access dominates the energy cost (see \figref{fig:cache:b}).
    Since DRAM (off-chip) accesses are only necessary when the on-chip cache misses, we can use the cache-hit-rate as the multiplier to estimate the energy cost when executing a layer.
\end{itemize}
These principles reduce the overhead to estimate the energy cost to execute a neural network on a platform.

\begin{figure}[t]
  \centering
  \includegraphics[width=.9\textwidth]{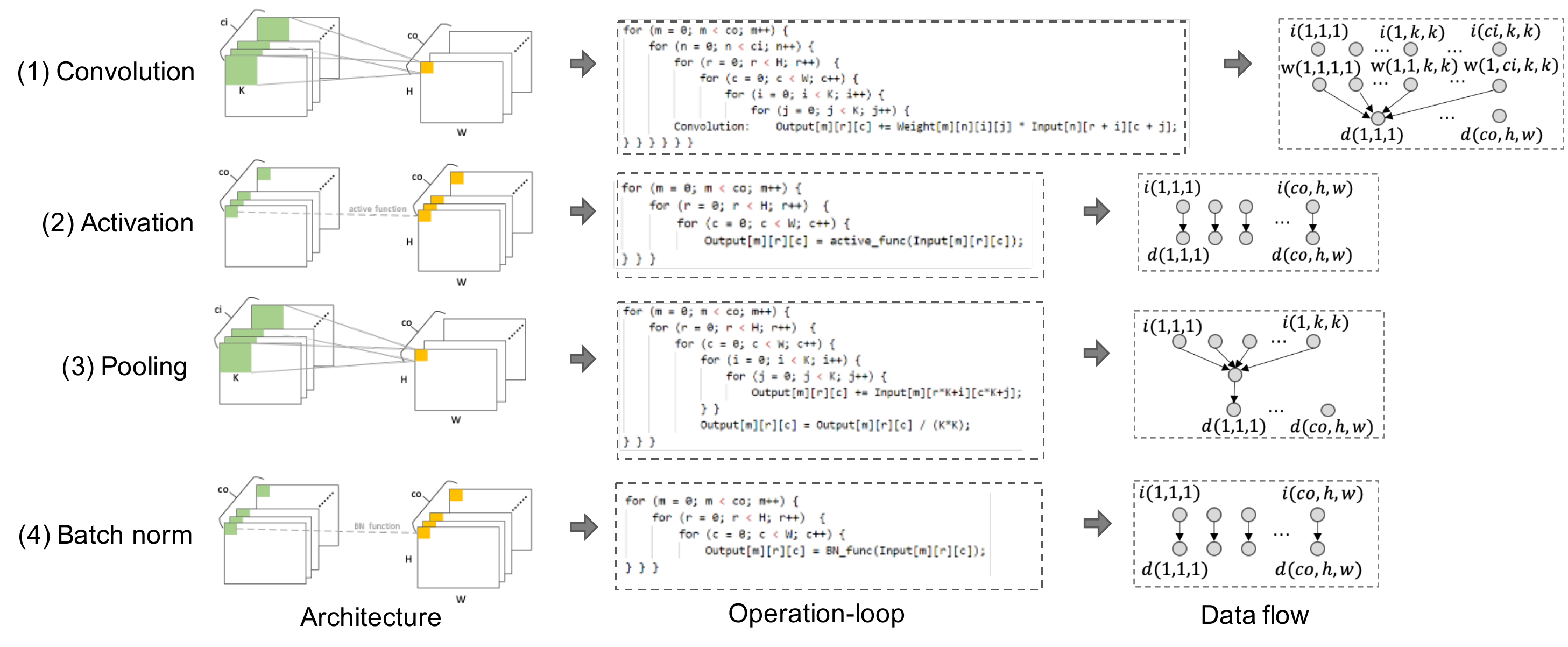}
  \caption{Counting the memory accesses from the underlying operation loops and data flow of different layers.}
  \label{fig:dataflow}
  \vspace{-3mm}
\end{figure}

\subsubsection{Energy Model}
We first propose the energy model to execute a single layer $l$ \rev{on mobile CPU/GPU platforms}.
\rev{
\begin{equation}\label{equ_energy}
    E_l = \delta_C \times C_l + \epsilon \times \delta_{cache}\times M_l + (1-\epsilon) \times \delta_{DRAM} \times M_l + M_l \times \delta_{SM}
\end{equation}
}
where $\delta_C$, $\delta_{cache}$, $\delta_{DRAM}$, \rev{and $\delta_{SM}$} are the unit energy cost of MAC computation, cache access, DRAM access, \rev{and the shared memory access};  $C_l$ and $M_l$ are the total number of MAC and memory access of layer $l$, respectively; and $\epsilon$ is the runtime cache-hit-rate.
For a given hardware platform, these parameters can be derived as follows.
\begin{itemize}
    \item \textit{Unit Energy Cost $\delta_C$, $\delta_{cache}$,  $\delta_{DRAM}$}, and $\delta_{SM}$.
    These parameters can be measured offline. 
    \rev{We empirically set $\delta_C : \delta_{cache}: \delta_{DRAM} : \delta_{SM} = 1:6:200:2$ for mobile GPU platforms.}
    \rev{As for mobile CPU platforms, $\delta_{SM} =0$ since they do not have such shared memory space,} and thus $\delta_C : \delta_{cache}: \delta_{DRAM} = 1:6:200$. 
    \rev{According to our evaluations, these parameters work with different DNN inference frameworks, \eg Raspberry Pi 4B (CPU) + NCNN, Hornor 9 (CPU) + Pytorch Mobile, and Nvidia Jetson Nano (GPU). }
    \item \textit{Number of MAC Computation $C_l$}.
    The amount of MAC can be directly derived from the layer architecture of layer $l$, \eg $C_{conv_l}=K^2* C_o*C_i* H* W$.
    \item \textit{Number of Memory Accesses $M_l$}.
    The memory access during the tensor computation consists of the accesses of layer parameters, input feature maps, and intermediate iterations. 
    We compute $M_l$ for a given layer type according to the data flow mapping to the given hardware platform.
    \figref{fig:dataflow} illustrates the data flow operations of different layer types used by our video enhancement model.
    For example, the total amount of memory accesses $M_l$ for a convolution layer $l$ can be calculated by the six-fold operation loops, \ie $M_{conv_l} = 2*C_o*C_i*H*W*K^2 + C_o*H*W$. 
    \item \textit{Runtime Cache-Hit-Rate $\epsilon$}.
    It accounts for the dynamics of the execution resources and thus is measured at runtime.
    We measure $\epsilon$ as the ratio of the actual MAC execution amount per unit time (\ie $ms$) to the MAC amount with the simulated $100\%$ cache-hit-rate.
    The amount of MAC executed with a $100\%$ cache-hit-rate is profiled offline for the given platform. 
    \rev{Observing that framework/compiler-level optimization such as operator fusion may affect the cache hit rate $\epsilon$ even for the same model~\cite{bib:mobisys21:zhang, bib:osdi2018:chen}, we measured the amount of MAC executed with a $100\%$ cache-hit-rate for multiple mainstream frameworks.} 
\end{itemize}

Given the energy model for a single layer $l$ as in \equref{equ_energy}, we can estimate the total energy consumption of $n_{frame}$ video frames, with adjustable parameters $\theta_l$, $\theta_f$, and $\theta_d$ as follows.
\begin{equation}\label{equ_total_energy}
    E_{enhancer} = \sum_{f=1}^{n_{frame}} \sum_{l=1}^{L}{\theta_{f_{(index_f)}} \theta_{l_{(index_l)}} E_l \theta_{d_f} }
\end{equation}
where $\theta_{f_{(index_f)}}$ and $\theta_{l_{(index_j)}}$ are Boolean value, and 0 means skipping the computation of frame $f$ or layer $l$. 
And $\theta_d$ is the down-sampling rate ($\%$) that affects the resolution of the input frame $f$ by $\theta_{d_f}*H*W$.

\subsubsection{Profiler Workflow}
The energy profiler in \sysname works in two stages, \textit{offline} and \textit{online}.
\begin{itemize}
    \item \textit{Offline Stage}.
    The unit energy cost $\delta_C, \delta_{cache}$,  $\delta_{DRAM}$, \rev{$\delta_{SM}$}, as well as the MAC throughput of the simulated $100\%$ cache-hit-rate are measured offline for the given platform.
    Specifically, during the offline data collection stage, we use a digital power monitor, \ie Monsoon AAA10F, to sample the ground truth of power consumption by profiling the device through its external power input.
    The energy cost of accessing the Cache, DRAM, and shared memory normalized to that of a MAC operation is determined by the platform based on the ground truth.
    The MAC execution amount per second in the simulated 100\% cache-hit-rate is the hardware-specified computation frequency times the computational parallelism (\eg 16 bits).
    The result is an energy profile with a sequence samples $\{catch-hit-rate/power~ drawn \}$.
    \item \textit{Online Stage}.
    During the online profiling stage, the profiler takes the current model hyperparameters and the runtime cache-hit-rate $\epsilon$ as input, and predicts the energy demand using the energy model in \equref{equ_total_energy}.
\end{itemize}
\rev{Note that the primary goal of the profiler is to ensure consistent \textit{ranking} between the \textit{estimated} and the \textit{actual} energy cost tested on the mobile device.
The consistency in ranking suffices to provide accurate feedback to dynamically adjust the video enhancement model’s hyperparameters for energy conservation.}

% \rev{We note that the ranking of the energy costs estimated by the profiler is consistent with the ranking of actual costs tested on mobile devices, which can provide an accurate feedback for the adaptation controller to dynamically adjust (\ie enlarge or reduce) the video enhancement model’s hyperparameters for energy conservation.}

\subsection{Automated Adaptation Loop}
\label{subsec:loop}

\begin{figure}[t]
  \centering
  \includegraphics[width=.83\textwidth]{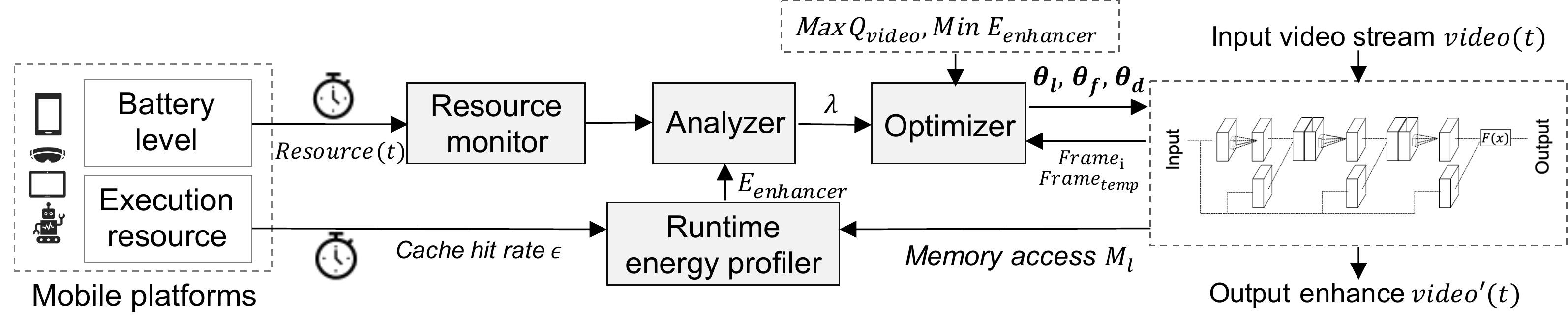}
  \caption{Energy-aware adaptation controller workflow.}
  \label{fig:loop}
  \vspace{-3mm}
\end{figure}

\figref{fig:loop} shows the overall workflow of the energy-aware adaptation controller.
The adaptation loop consists of the \textit{resource monitor}, the \textit{runtime energy profiler}, the \textit{analyzer}, and the \textit{optimizer}.
The \textit{resource monitor} tracks the energy supply of the platform, while the \textit{energy profiler} (\secref{subsec:profiler}) predicts the energy demand of the video enhancement process with the current configurations.
If the energy demand exceeds the supply, the \textit{analyzer} notifies the \textit{optimizer}, which then re-selects $\theta_l, \theta_f, \theta_d$ as in \secref{subsubsec:optimizer}.
The control loop routinely checks for changes in the system and performs adaptations at a pre-defined frequency (\eg 1 second).
%(\TODO{concrete number in our case}).

%\TODO{other implementation details about the resource monitor and the analyzer, or the controller as a whole}

\section{Evaluation}
\label{sec:experiment}
%This section presents the evaluations of \sysname. 

\subsection{Experiment Setups}
\label{sec_exp_setup}
%This subsection presents the settings for our evaluation.

\subsubsection{Datasets.}
\label{subsec:datasets}
We experiment with seven datasets, six open low-light image/video enhancement benchmark datasets (SCIE~\cite{bib:tip2018:cai}, Loli-Phone~\cite{bib:TPAMI21:Li}, LOL~\cite{bib:arXiv2018:wei}, VV~\cite{data:vv}, LIME~\cite{bib:TIP16:Guo},  NightOwls~\cite{data:NightOwls}) and one self-collected dataset (MobileScene).
\tabref{tb_dataset} lists the details of these datasets.

%%%%%%%%%%%%%%%%%二、晓晨：列出采用的数据集。
%1，训练数据集1，SCIE数据集，引用：https://ieeexplore.ieee.org/abstract/document/8259342，image数据，SCIE数据集包含了多种曝光等级的图像，我们使用了其中2002张图像，和Zero-DCE的release版本一致。我们使用该部分数据用于训练单帧图像增强后的曝光等级，平滑性，邻域性，色彩均衡。

%2，训练数据集2，Loli-Phone数据集，引用：https://arxiv.org/abs/2104.10729，video数据，Loli-Phone数据集包含了多条手机拍摄的视频，我们仅使用了其中的一条包含300帧的视频。我们使用该数据集用于训练模型时域上色彩和曝光等级的一致性。

%3，测试数据集1，VV数据集，图像数据集，引用：https://sites.google.com/site/vonikakis/datasets,仅有网址，未给出cite。24张图片，无参考

%4，测试数据集2，LIME数据集，图像数据集，引用：https://ieeexplore.ieee.org/abstract/document/7782813。10张图片，无参考

%5，测试数据集3，NightOwls数据集，视频数据集，但采用图像评价指标进行评价。引用：https://www.nightowls-dataset.org/download/。无参考，我们随机选取了500视频帧用于测试

%6，测试数据集4，LOL数据集，引用：https://arxiv.org/abs/1808.04560。图像数据集，500对低/正常光照图像数据。有参考

%7，测试数据集5：NightOwls，数据集多场景视频。我们从中抽取了40个视频段，每个视频段包含了15帧视频帧。这40个视频段根据场景划分为静态镜头+静态场景，动态镜头+静态场景，静态镜头+动态场景，动态镜头+动态场景四类，用于衡量视频的时域稳定性(此处衡量采用视频评价指标进行评价，MABD，TSSIM)

%8. 下游任务1：人脸探测，Dark Face数据集，引用：https://ieeexplore.ieee.org/document/9049390，测试算法，DSFD，引用，https://arxiv.org/abs/1810.10220。数据集包含了6000张夜间场景图像，以及对应的人脸位置标注。我们采用DSFD算法对增强前后的图像进行人脸探测，结果显示增强后的图像对人脸探测算法更加友好。

%9. 下游任务2：目标检测，COCO数据集，2017-test部分，引用https://arxiv.org/abs/1405.0312。测试算法，Detectron2，实现的Faster-RCNN，引用,https://github.com/facebookresearch/detectron2, (detectron2),https://arxiv.org/abs/1506.01497, (Faster-RCNN).为了模拟低照度环境，我们对原始图像进行了伽马校正预处理。处理公式为Low = A*Raw^{\gamma}，其中A,\gamma均为常数，分别取值为1，5。处理后我们得到了模拟低照度的图像结果。随后我们对低照度图像结果进行增强，并采用Faster-RCNN分别对增强前后的图像进行目标检测，可以看到低照度增强技术对目标检测技术同样有着较好的辅助效果。

We use SICE \cite{bib:tip2018:cai} and Loli-Phone~\cite{bib:TPAMI21:Li} for training.
SCIE contains images of different exposure levels.
We use it to train the single-frame image's enhancement performance of exposure level, smoothness, neighborhood, and color balance.
Loli-Phone contains multiple video clips taken by smartphones, and we adopt it for training the exposure and color consistency of the video enhancement model.
Specifically, we use SICE as the image training dataset as Zero-DCE \cite{bib:CVPR20:Guo}, and Loli-Phone as the video training dataset for the fast low-light video enhancement model of \sysname.
All the images and video frames are resized as $256\times256\times3$. 
We use Adam as the optimizer and train the model for $100$ epochs. 
The learning rate is set to $0.0001$ with the batch size of $4$.

The remaining datasets are for testing.
We consider datasets with both paired and unpaired samples.
VV~\cite{data:vv} and LIME~\cite{bib:TIP16:Guo} are unpaired low-light image datasets, NightOwls~\cite{data:NightOwls} is an unpaired low-light video dataset, and LOL~\cite{bib:arXiv2018:wei} is a paired low-light image dataset.
MobiScene is a self-collected dataset including four scenarios (\ie moving/static cameras with moving/static objects), which is used to measure the enhanced video's temporal stability.
The concrete performance metrics of each dataset are explained in the setups of each experiment below.

%And we use a self-collected dataset (\eg moving/static cameras for moving/static objects) to test \sysnameposs performance over diverse mobile scenarios.
\begin{table}[t]
\scriptsize
\caption{Overview of low-light image and video datasets.}
\vspace{-3mm}
\begin{tabular}{|l|l|l|l|}
\hline
\multicolumn{1}{|c|}{\textbf{Datasets}} & \multicolumn{1}{c|}{\textbf{Data type}} & \multicolumn{1}{c|}{\textbf{Sample number}} & \multicolumn{1}{c|}{\textbf{Description}} \\ \hline
\textbf{1. SCIE~\cite{bib:tip2018:cai}} & Images & 2002 pieces & Diverse exposure levels \\ \hline
\textbf{2. Loli-Phone~\cite{bib:TPAMI21:Li}} & Videos  & 300 frames & Taken by real-world smartphones \\ \hline
\textbf{3. VV~\cite{data:vv}} & Unpaired images & 24 pieces & Low-light images \\ \hline
\textbf{4. LIME~\cite{bib:TIP16:Guo}} & Unpaired images & 10 pieces & Low-light images \\ \hline
\textbf{5. NightOwls~\cite{data:NightOwls}} & Unpaired videos & 500 piece &  Low-light videos \\ \hline
\textbf{6. LOL~\cite{bib:arXiv2018:wei}} & Paired images & 500 pairs & Low-light images with the corresponding high-light images \\ \hline
\textbf{7. MobiScene} & Unpaired low-light videos & 40 pieces & Self-collected videos with moving/static cameras and moving/static objects \\ \hline
\vspace{-3mm}
\end{tabular}
\label{tb_dataset}
\end{table}

%(\ie MABD and TSSIM metrics). 
%It is worth noting that none of the paired low-light video datasets are needed to train \sysnameposs video enhancement model.
% In addition, we employ \TODO{}xx and \TODO{}xx datasets, respectively, to train the models of the downstream on-device video applications.

%%%%%%%%%
% TODO: to Xiaochen: is our model trained only on this dataset? From my understanding, the model is tested on different datasets. Maybe we should move this to the evaluat
% 我们的训练只用到了SCIE数据集和Loli-Phone数据集，其它数据集均为评价使用
%%%%%%%%%

\subsubsection{Baselines} 
\label{sec:baseline}
%%%%%%%%%%%%%%%%%三、晓晨：列出采用的数据集
%1. MBLLEN，引用，http://bmvc2018.org/contents/papers/0700.pdf，MBLLEN提出了一种端到端的多分枝网络结构，特别考虑了图像的噪声。此外，作者提出通过将2D卷积替换为3D卷积可以可以实现对视频的处理，但由于2D卷积带来的资源消耗已经过大，因此我们未进一步对比其3D卷积版本
%2. Zero-DCE Zero-DCE的主要结构为一个神经网络模型和一个可迭代曲线，模型训练时采用了零参考(Zero-Reference)的训练策略，实现了采用非成对数据(unpaired)进行网络训练
%3. Zero-DCE++ Zero-DCE++主要在Zero-DCE的基础上进行了剪枝(缩小模型输入尺寸，减少模型输出层通道数)操作，并将原有的卷积替换为了深度可分离卷积。
%4. StableLLVE ，引用，https://openaccess.thecvf.com/content/CVPR2021/html/Zhang_Learning_Temporal_Consistency_for_Low_Light_Video_Enhancement_From_Single_CVPR_2021_paper.html，StableLLVE提出了一种视频增强模型。它通过光流模型从单张图像中提取时域信息，避免了训练过程中对视频数据的依赖，提高了增强后视频的时域稳定性。
%5. Ours，我们的模型在Zero-DCE的基础上进行了剪枝(减少模型层数和通道数至3层4通道)，并采用可迭代的伽马校正替代原有的可迭代曲线
%6. release

%对比了哪些性能：参数量，处理延迟
%%%%%%%%%%%%%%%%%
We compare the performance of \sysname with the following baselines.
\begin{itemize}
    \item \textbf{Zero-DCE}~\cite{bib:CVPR20:Guo}.
    It is a lightweight low-light image enhancement scheme which adopts zero-shot learning to avoid training with paired data and employs an iterative curve-based architecture to map a low-light image to its enhanced version.
    \item \textbf{Zero-DCE++}~\cite{bib:TPAMI21:Li2}.
    It is the more resource-efficient version of Zero-DCE by compressing the neural network to learn the curve parameters by depth-wise separable convolutions and curve parameter map reuse.
    %  TODO: image or video 
    % 这里应当还是把MBLLEN看作图像增强，但是较Zero-DCE额外考虑了空域上的噪声，由于空域噪声也对时域有着影响，因此在时域降噪上也有不错的表现
    \item \textbf{MELLEN}~\cite{bib:BMVC18:Lv}.
    It is a recent low-light image enhancement method that achieves high enhancement quality.
    It extracts rich features via multiple sub-networks and generates the output with multi-branch fusion.
    \item \textbf{StableLLVE}~\cite{bib:CVPR21:Zhang}.
    It is a state-of-the-art low-light video enhancement model which enforces the temporal stability among frames.
    \item \textbf{\sysname Basic}.
    It is the variant of \sysname without incorporating the temporal consistency loss.
    \item \rev{\textbf{\sysname W/O\_Align}. 
    It is the variant of \sysname without the optical flow alignment pre-process.}
\end{itemize}
%Zero-DCE~\cite{bib:CVPR20:Guo} and Zero-DCE++~\cite{bib:TPAMI21:Li2} are among the most lightweight low-light image enhancement schemes.
%StableLIVE~\cite{bib:CVPR21:Zhang} is dedicated to low-light video enhancement.
%We include AdaEnlight\_no to highlight the importance of the temporal consistency loss for video enhancement.

% Zero-DCE~\cite{bib:CVPR20:Guo} and Zero-DCE++~\cite{bib:TPAMI21:Li2} provide a high speed standard for the image (\ie single frame) enhancement process.
% In particular, Zero-DCE is a state-of-the-art lightweight deep model for low-light image enhancement. 
% And Zero-DCE++ is the compressed version of Zero-DCE with lower model complexity. 
% MELLEN~\cite{bib:BMVC18:Lv} establishes a high visual quality basic, although its high computational complexity makes it unsuitable for resource-constrained mobile platforms.
% We employ the Zero-DCE, Zero-DCE++, and MELLEN methods to enhance the input ubiquitous video frame by frame.
% StableLIVE~\cite{bib:CVPR21:Zhang} provides a strict benchmark against which we can validate \sysnameposs advantage in the output video's temporal stability.
% And we further use the variant AdaEnlight\_no to validate the \sysnameposs benefit in temporal stability.

%  TODO: did we implement the baselines? in TODO: language
% Zero-DCE,Zero-DCE++和StableLLVE采用Pytorch进行了实现，MBLLEN采用Tensorflow进行了实现。此外，我们还在树莓派上用NCNN对Zero-DCE和Zero-DCE++进行了部署
\subsubsection{Implementation}
We implement all the compared methods in Pytorch. 
Training of the algorithms is performed on a server with NVIDIA GeForce RTX 2070 GPU and Intel Core i5-10400F CPU at 2.90GHz and CUDA 10.0.
We deploy each method to mobile devices for low-light enhancement.
%Additionally, we deployed Zero-DCE and Zero-DCE++ with NCNN on Raspberry Pi.
Specifically, we test three commercial mobile camera-embedded platforms, including a personal smartphone, \ie \rev{Honor 9 with Octa-core CPU processor (device 1), an embedded development board, \ie raspberry Pi 4B with Quad-core CPU (device 2), and a mobile platform, \ie NVIDIA Jetson Nano with 128-core NVIDIA Maxwell GPU (device 3)}.
For mobile deployment of \sysname, \rev{we use the Pytorch Mobile on the Android 12.0 platform (device 1) the NCNN framework on Raspberry Pi OS (device 2), and Pytorch 1.4 on Ubuntu 18.04 platform (device 3).} 
\rev{For platforms with both mobile CPU and GPU, we configure the computing tasks to execute on either the GPU or CPU to avoid excessive GPU-CPU communication.}
We use OpenCV to invoke the embedded camera for video shooting at a rate of 20 frames per second.
The video frames are resized to $270\times{480}$ and fed into each algorithm for enhancement.

\subsection{Low-light Video Enhancement Performance}
\label{subsec:exp_compare}
% 画三个图
%1. 对比ours 和其他baseline的 运行 时延---树莓派上
%2. 对比ours 和其他baseline的 参数量 [huag]
%3. 对比ours 和其他baseline的 视觉质量，两个数据集
This section presents the results on low-light enhancement compared with the baselines.

\subsubsection{Performance on Open Benchmarks}
This experiment compares the performance of \sysname and different baseline methods on open low-light enhancement datasets.
%%%%%%%%
% TODO: 60 in each dataset or in tota
% TODO: explain the metrics for visual quality
% VV(24张)，LOL(500张)，LIME（10张）
% NIQE指标是无参考指标，用于衡量图像/视频帧的自然程度，越低的NIQE分数代表了越高的图像质量
% SSIM指标衡量了两张图像之间的结构相似性。越高的SSIM分数代表了两张图象之间有着越高的相似度。
% PNSR指标衡量了两张图像之间的峰值信噪比的相似度。越高的PNSR分数代表了两张图像之间有着越高的相似度
%　MAE指标衡量了两张图象之间的平均绝对误差，越低的MAE分数代表了两张图象有着越高的相似度
%　此处三个有参考指标均用于衡量增强后的低照度图像与良好光照图像之间的相似度

%%%%%%%%
\textbf{Setups}.
We use three low-light image datasets, \ie VV (task3), LIME (task4), and LOL (task6) to assess the enhancement performance on a single frame, and 10 video streams from the low-light video dataset NightOwls (task5).
We compare the visual quality of the enhanced images or videos and the latency to process each image or video frame.
We use four visual quality metrics, \ie structural similarity (SSIM)\cite{bib:TIP2004:wang}, peak signal-to-noise ratio (PSNR), mean absolute error (MAE), and natural image quality evaluator (NIQE)\cite{bib:SPL2012:mittal}. 
SSIM, PSNR, and MAE measure the visual disparity of low-light images/videos from the labeled high-quality ones, while NIQE is a no-reference metric to measure the deviations from statistical regularities observed in natural images, without training on paired data. 
The experiments are conducted on Raspberry Pi 4B \rev{(with CPU, device 2)}.
We run the measurements for six times and report the average values.

\begin{table}[t]
\centering
\scriptsize
\caption{Performance comparison of \sysname with four baselines in terms of parameter size, latency, and visual quality.}
\vspace{-3mm}
\begin{tabular}{|c|c|c|cccccc|}
\hline
 &  &  & \multicolumn{6}{c|}{\textbf{Output visual quality}} \\ \cline{4-9} 
 &  &  & \multicolumn{3}{c|}{\textbf{NIQE $\downarrow$}} & \multicolumn{1}{c|}{\textbf{SSIM $\uparrow$}} & \multicolumn{1}{c|}{\textbf{PSNR $\uparrow$}} & \textbf{MAE $\downarrow$} \\ \cline{4-9} 
\multirow{-3}{*}{\textbf{Methods}} & \multirow{-3}{*}{\textbf{\begin{tabular}[c]{@{}c@{}}Parameter\\  size\end{tabular}}} & \multirow{-3}{*}{\textbf{\begin{tabular}[c]{@{}c@{}}Average latency \\ per frame (s)\end{tabular}}} & \multicolumn{1}{c|}{\textbf{VV (task3)}} & \multicolumn{1}{c|}{\textbf{LIME (task4)}} & \multicolumn{1}{c|}{\textbf{NightOwls (task5)}} & \multicolumn{3}{c|}{\textbf{LOL (task6)}} \\ \hline
\textbf{Zero-DCE~\cite{bib:CVPR20:Guo}} & 308.6KB & 11.743 & \multicolumn{1}{c|}{2.751} & \multicolumn{1}{c|}{{\color[HTML]{000000} 3.812}} & \multicolumn{1}{c|}{{\color[HTML]{000000} 3.541}} & \multicolumn{1}{c|}{{\color[HTML]{000000} 0.846}} & \multicolumn{1}{c|}{{\color[HTML]{000000} 14.132}} & {\color[HTML]{000000} 52.172} \\ \hline
\textbf{Zero-DCE++~\cite{bib:TPAMI21:Li2}} & 39.1KB & 0.193 & \multicolumn{1}{c|}{2.547} & \multicolumn{1}{c|}{{\color[HTML]{000000} 3.82}} & \multicolumn{1}{c|}{{\color[HTML]{000000} 3.159}} & \multicolumn{1}{c|}{{\color[HTML]{000000} 0.864}} & \multicolumn{1}{c|}{{\color[HTML]{000000} 14.596}} & {\color[HTML]{000000} 49.871} \\ \hline
\textbf{StableLLVE~\cite{bib:CVPR21:Zhang}} & 15.8MB & 1.2 & \multicolumn{1}{c|}{2.224} & \multicolumn{1}{c|}{{\color[HTML]{000000} 3.725}} & \multicolumn{1}{c|}{{\color[HTML]{000000} 3.793}} & \multicolumn{1}{c|}{{\color[HTML]{000000} 0.864}} & \multicolumn{1}{c|}{{\color[HTML]{000000} 17.142}} & {\color[HTML]{000000} 34.921} \\ \hline
\textbf{MBLLEN~\cite{bib:BMVC18:Lv}} & 1.7MB & 6.225 & \multicolumn{1}{c|}{2.76} & \multicolumn{1}{c|}{{\color[HTML]{000000} 3.763}} & \multicolumn{1}{c|}{{\color[HTML]{000000} 3.333}} & \multicolumn{1}{c|}{{\color[HTML]{000000} 0.899}} & \multicolumn{1}{c|}{{\color[HTML]{000000} 17.229}} & {\color[HTML]{000000} 34.296} \\ \hline
\textbf{AdaEnlight} & 0.17KB & 0.05 & \multicolumn{1}{c|}{\rev{2.57}} & \multicolumn{1}{c|}{{\color[HTML]{000000} \rev{3.723}}} & \multicolumn{1}{c|}{{\color[HTML]{000000} \rev{2.882}}} & \multicolumn{1}{c|}{{\color[HTML]{000000} \rev{0.854}}} & \multicolumn{1}{c|}{{\color[HTML]{000000} \rev{17.232}}} & {\color[HTML]{000000} \rev{33.784}} \\ \hline
\end{tabular}
\label{tb_compare}
\vspace{-3mm}
\end{table}

%%%%%%%%%
% TODO: metrics quite confusing, why NIQE is used for both video and image? why SSIM, PSNR, MAE only for a single iamge dataset LO
% 由于我们的视频数据集没有标签，因此我们采用NIQE对增强后的视频帧质量进行逐一评价
%　SSIM，PSNR，MAE评价方法必须依赖于有参考的数据集，而我们使用的数据集中，仅LOL为有参考数据集
%%%%%%%%%
\textbf{Results}.
\tabref{tb_compare} summarizes the results.
\sysname outperforms the baselines in execution latency.
In fact, only \sysname achieves real-time low-light enhancement. 
The average delay of \sysname is about $0.05s$ per frame, which is smaller than the input streaming speed (\ie 0.1s per frame for the 10FPS video stream from NightOwls~\cite{data:NightOwls}).
In comparison, the delays of StableLIVE, MBLLEN, Zero-DCE, and Zero-DCE++ are $1.2$, $6.225s$, $11.743s$, and $0.193s$ per frame.
The low latency of \sysname attributes to the non-iterative enhancement with Gamma correction-based curve design (see \secref{subsec:algo:gamma}).
%Also, \sysname achieves the lowest number of parameters. It reduces the parameter size to $0.17KB$, a $230\times{}$ reduction over the most lightweight baseline Zero-DCE++.
Despite its low latency, \sysname achieves competitive visual quality.
Specifically, \sysname achieves the best NIQE on NightOwls (video) and LIME (image), and is at least as good as Zero-DCE++ in the NIQE metric for the VV task~\cite{data:vv}. 
\rev{Also, \sysname achieves the best PSNR and MAE on LOL \cite{bib:TIP16:Guo}.}
These outcomes validate the effectiveness of the proposed curve-based enhancement. 

\subsubsection{Performance on Real-world Mobile Scenarios}
\label{exp_video}
% 画四个图，一个场景一个图
%一维： 1.动态镜头+动态场景	2.动态镜头+静态场景	3.静态镜头+动态场景	 4.静态镜头+静态场景
%二维：四个baseline，以及没有temporal loss的情况
%三维：视觉指标1，指标2
% 结论：稳定性
This experiment assesses the temporal stability of the videos enhanced by \sysname and different baseline methods across four typical real-world mobile scenarios. 
Note that both the camera and the objects in videos may be in motion in mobile shooting scenarios, making it challenging for a low-light enhancement algorithm to deliver stable video streams.
% TODO: full name, ref
% TSSIM, temporal SSIM, cite "See Motion in the Dark"
% MABD . mean absolute brightness differences, cite Learning Temporal Consistency for Low Light Video Enhancement From Single Images

\textbf{Setups}.
We test four real-world mobile scenarios: 
\textit{Scenario A:} a moving camera capturing videos of moving objects, 
\textit{Scenario B:} a moving camera capturing videos of static objects, 
\textit{Scenario C:} a static camera capturing videos of moving objects, 
and \textit{Scenario D:} a static camera capturing videos of static objects.
We use our self-collected video dataset MobiScene and videos from NightOwls \rev{on Raspberry Pi 4B (with CPU, device 2)}.
We quantify the temporal stability of videos by temporal structural similarity index measure (TSSIM) and mean absolute brightness differences (MADB).
TSSIM is a common metric for the temporal structural similarity of adjacent frames in the video stream, and MADB measures the pixel difference between adjacent frames.
% \rev{We conduct this thread of experiments on device 2 using CPU processor.}

% This section evaluates the \sysnameposs performance under four typical mobile scenarios: \textit{(i) Scenario A:} the moving camera captures the video of objects in motion, \textit{(ii) Scenario B:} the moving camera captures the video of static objects, \textit{(iii) Scenario C:} the static camera captures the video of objects in motion, and \textit{(iv) Scenario D:} the static camera captures the video of static objects.
% We set up this thread of experiments because both the camera-embedded mobile platforms (\eg smartphone, robot) and the objects (\eg road shoulder, pedestrian) are always unpredictably stationary and moving in practice.
% And in these diverse scenarios, the difficulty of eliminating the flicking problem (see Sec. \ref{subsec:algo:temporal}) between consecutive frames in the video enhancement process is also different.
% We employ the self-collected videos and the derived videos from the NightOwls dataset to test the enhanced video's temporal stability, \ie TSSIM for video's frame structural similarity and MADB for pixel difference.
%晓晨：两个指标分别测试什么视觉语义信息？
% TSSIM计算了两个连续帧之间的SSIM分数，MABD计算了两个连续帧之间像素差分绝对值。两个指标均能够在一定程度上反应增强后的视频帧在时域上的稳定性。

\begin{figure}[t]
  \centering
   \subfloat[\rev{Moving cam, moving obj}]{\label{fig:t_scene1}
   \includegraphics[width=0.245\textwidth]{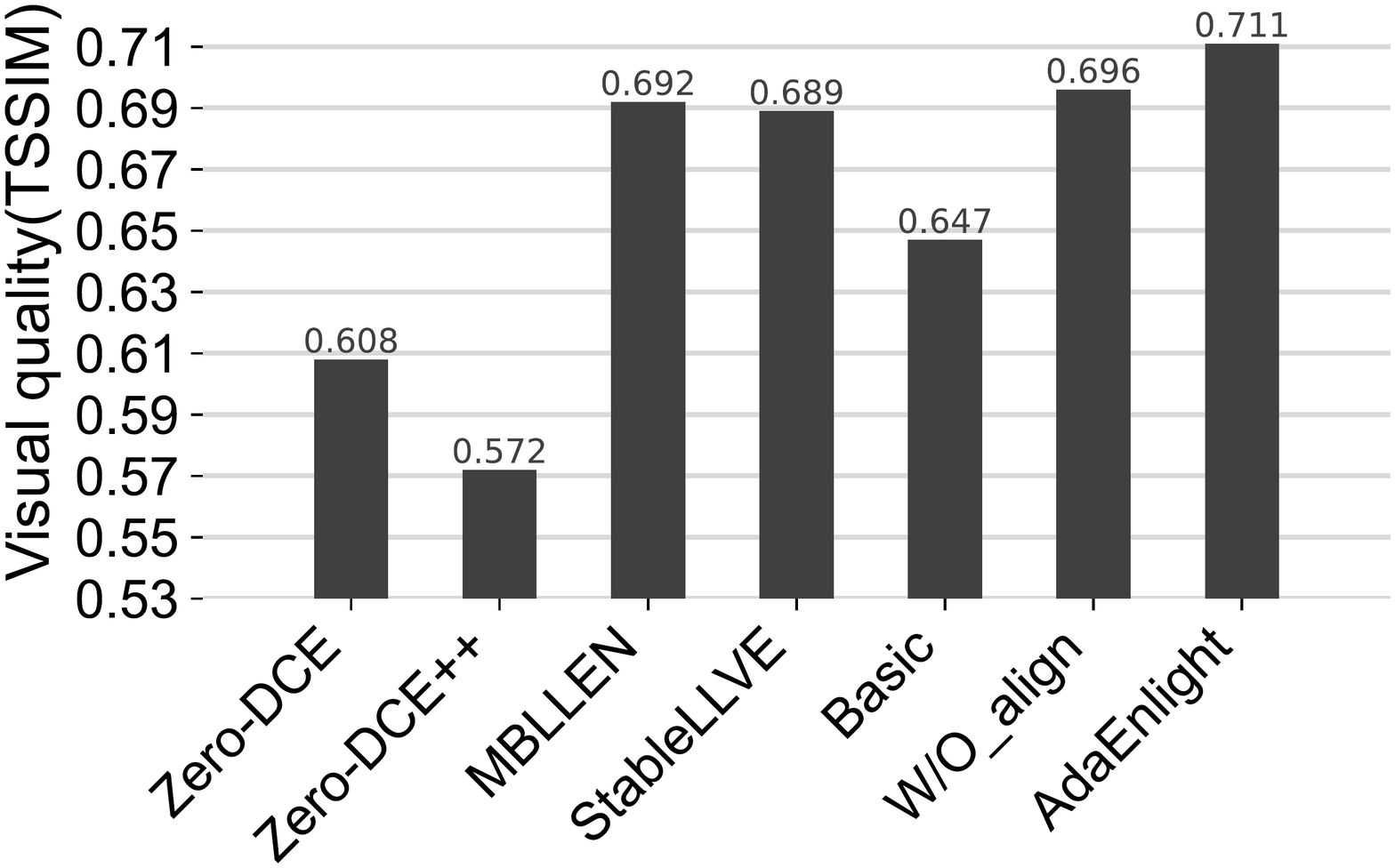}}
   \subfloat[\rev{Moving cam, static obj}]{\label{fig:t_scene2}
   \includegraphics[width=0.245\textwidth]{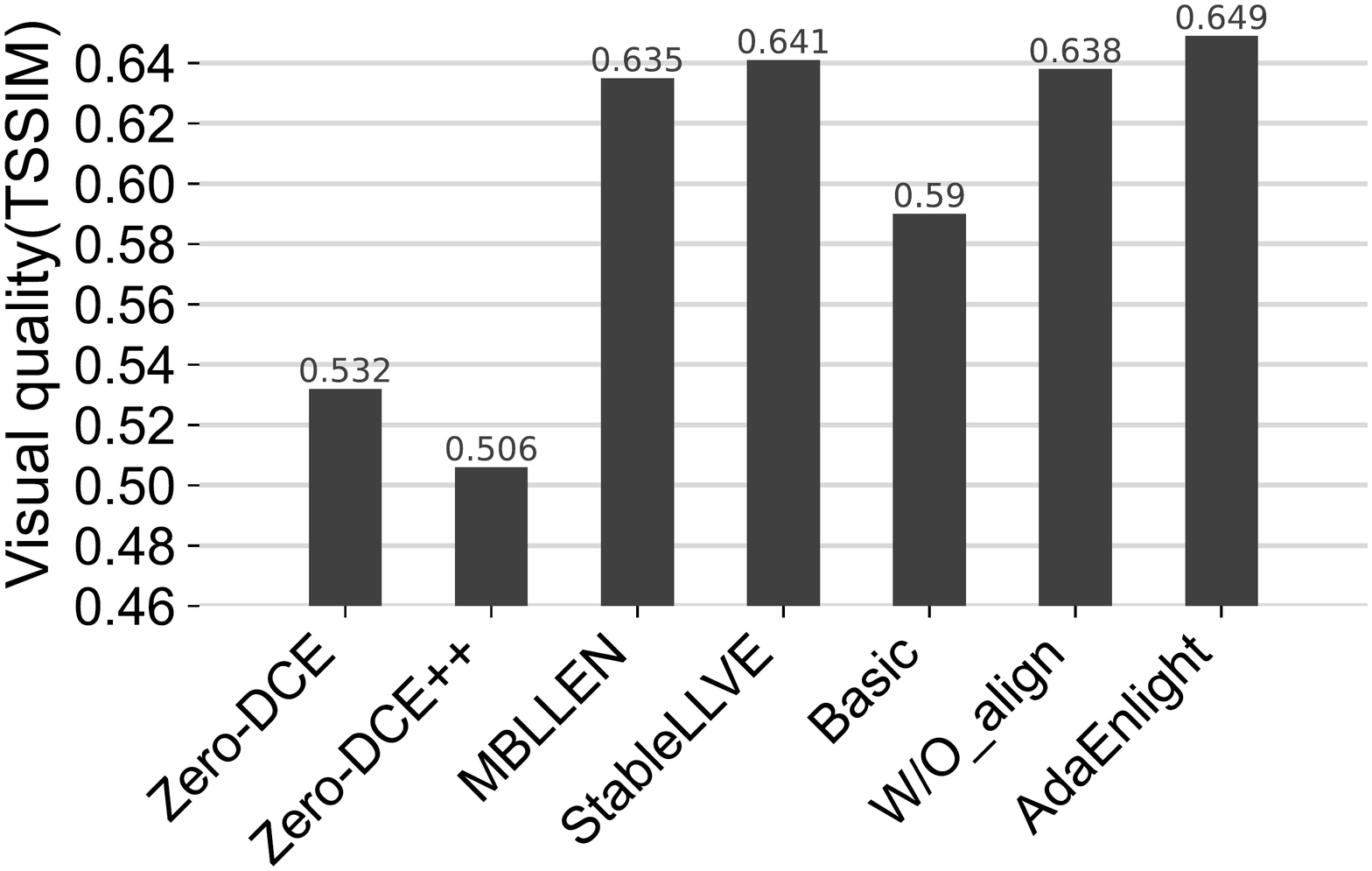}}
   \subfloat[\rev{Static cam, moving obj}]{\label{fig:t_scene3}
   \includegraphics[width=0.245\textwidth]{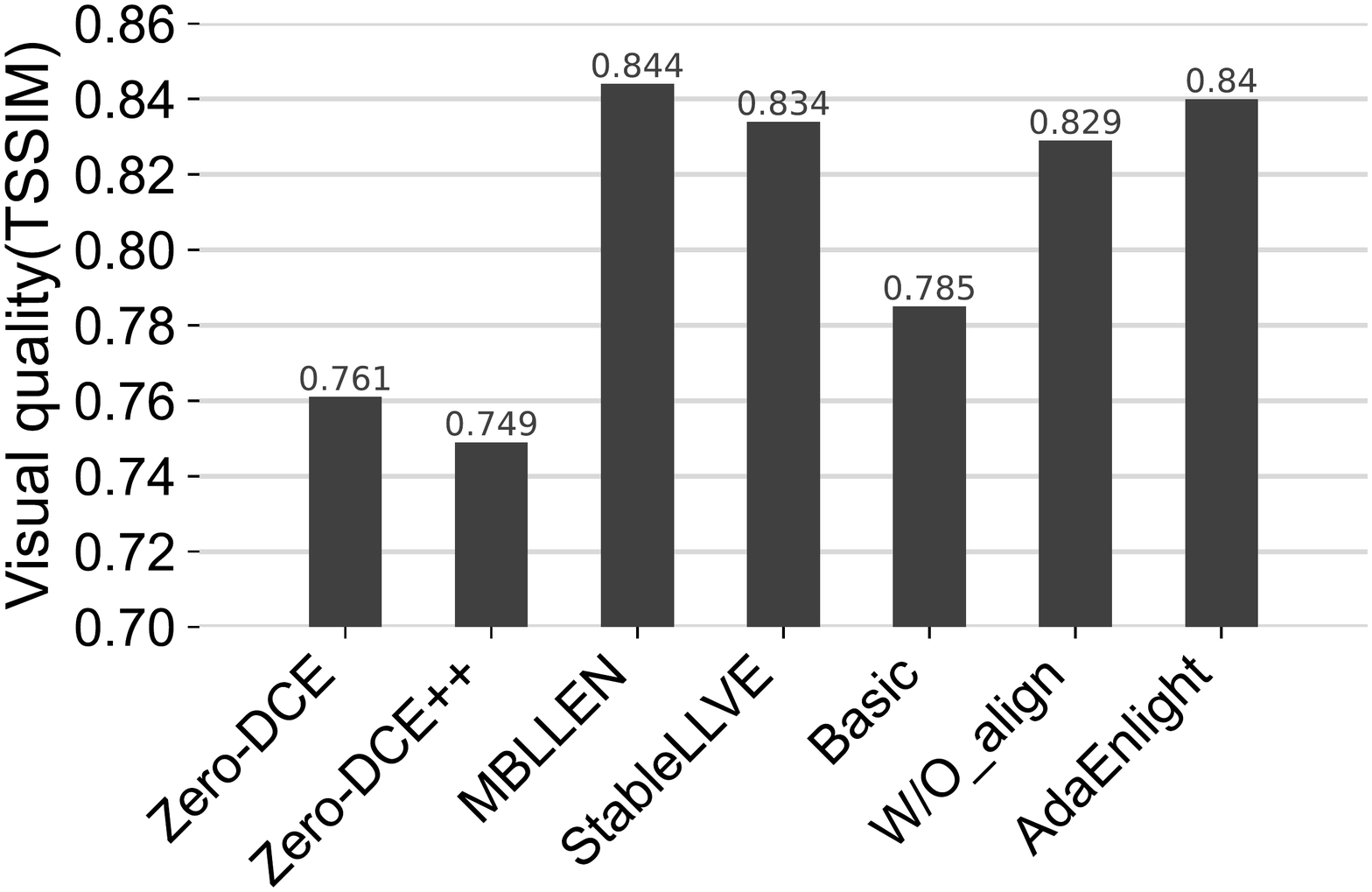}}
   \subfloat[\rev{Static cam, static obj}]{\label{fig:t_scene4}
   \includegraphics[width=0.245\textwidth]{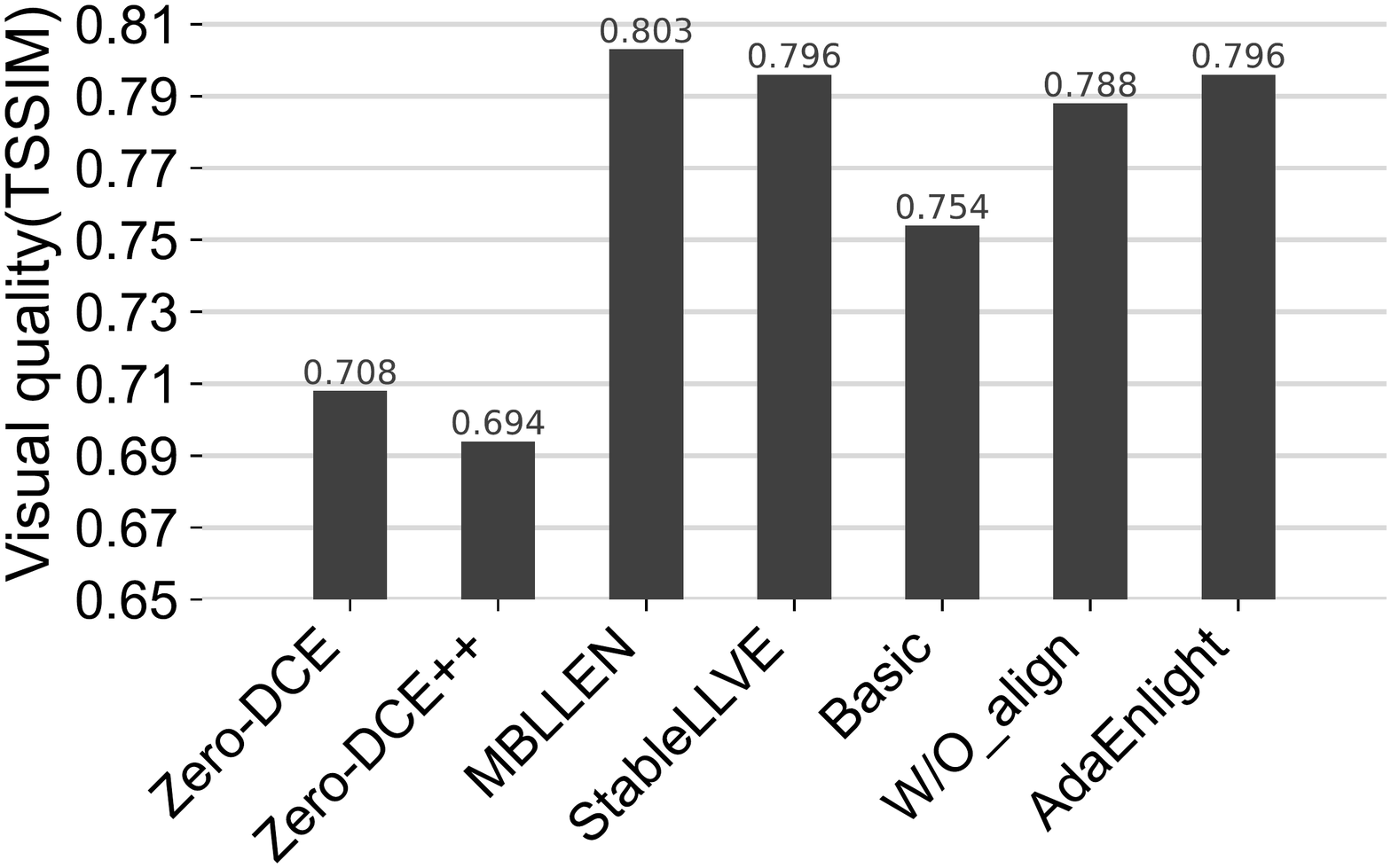}}\\
   \subfloat[\rev{Moving cam, moving obj}]{\label{fig::m_scene1}
   \includegraphics[width=0.245\textwidth]{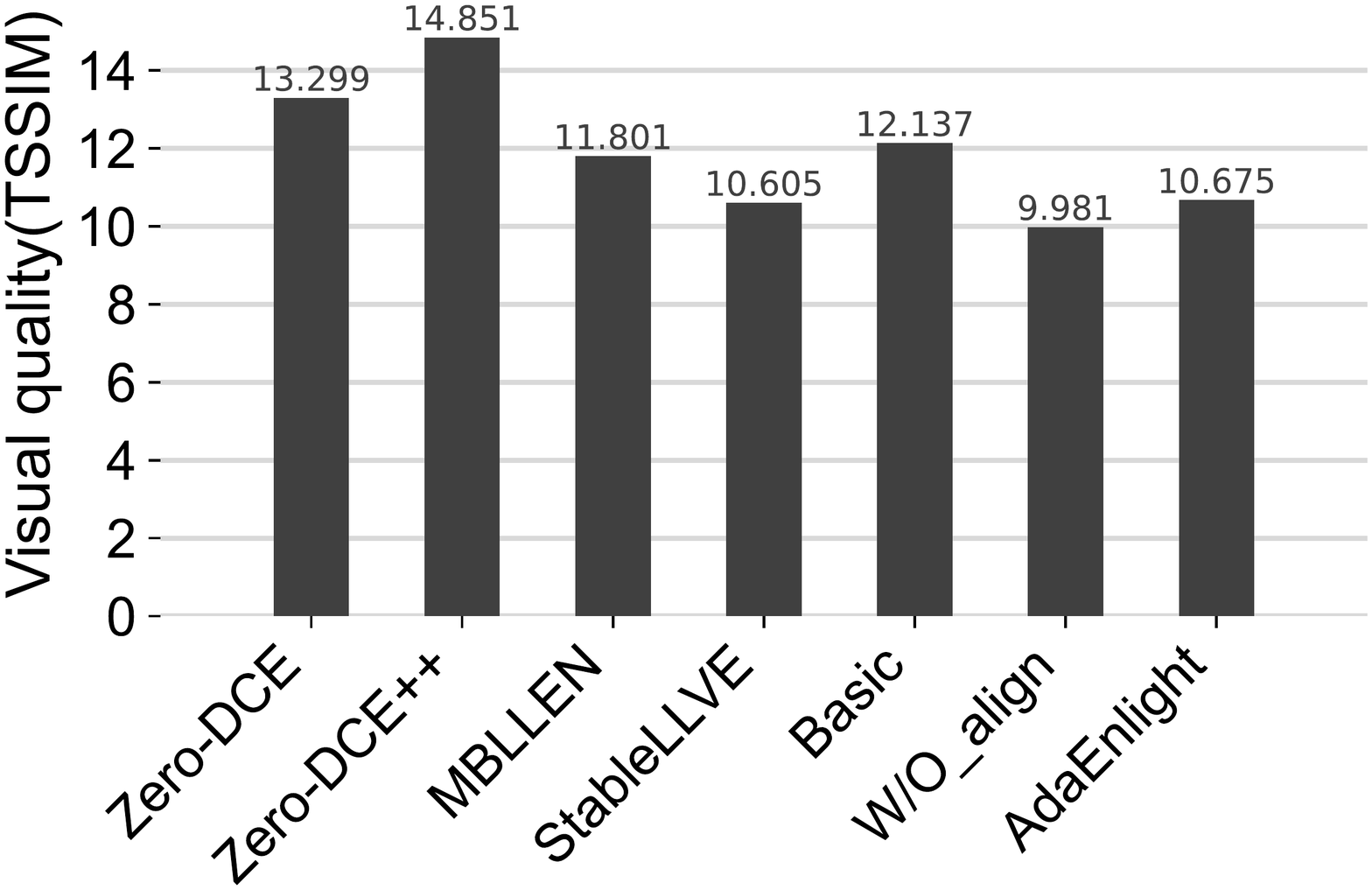}}
   \subfloat[\rev{Moving cam, static obj}]{\label{fig::m_scene2}
   \includegraphics[width=0.245\textwidth]{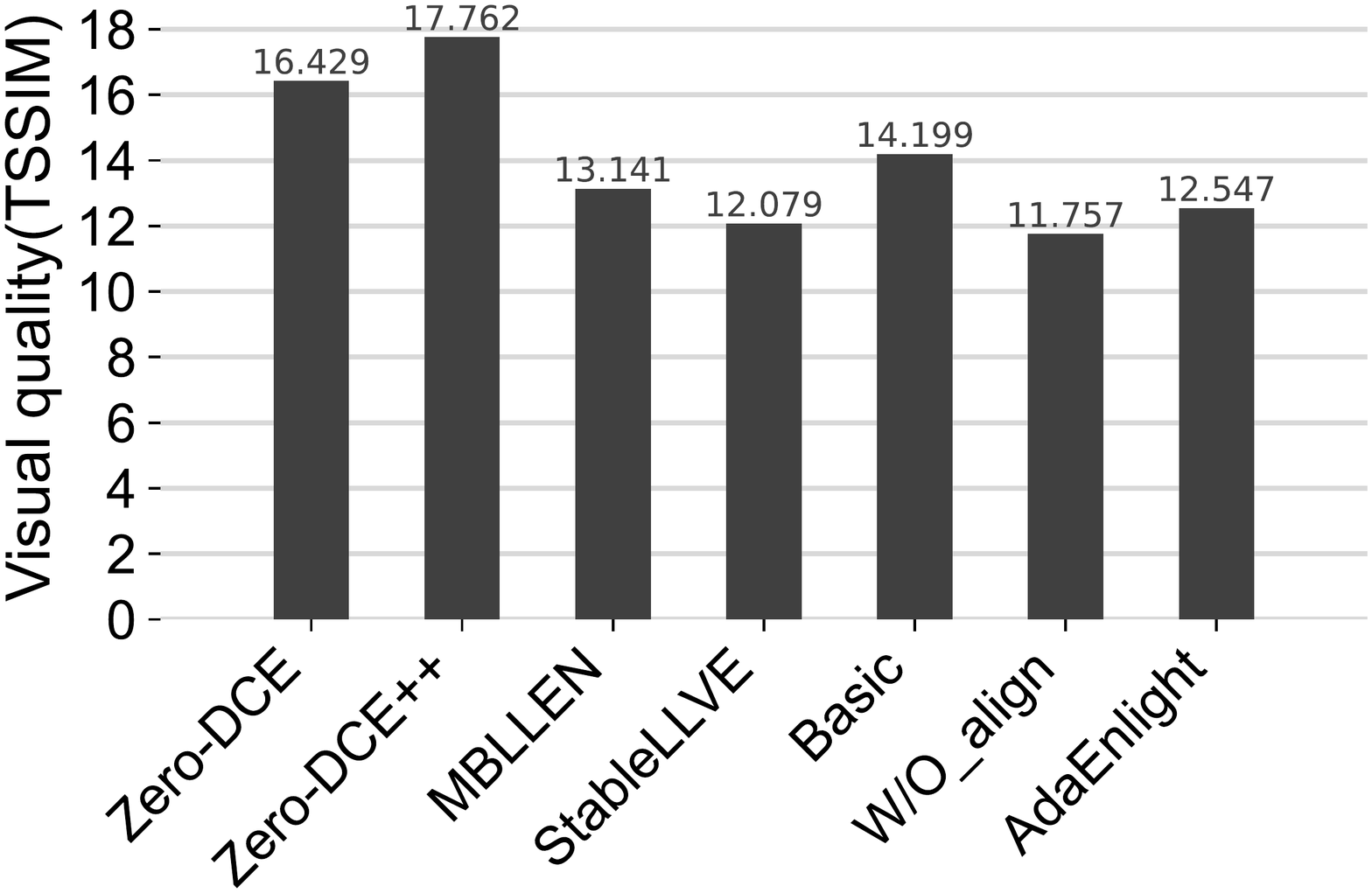}}
   \subfloat[\rev{Static cam, moving obj}]{\label{fig::m_scene3}
   \includegraphics[width=0.245\textwidth]{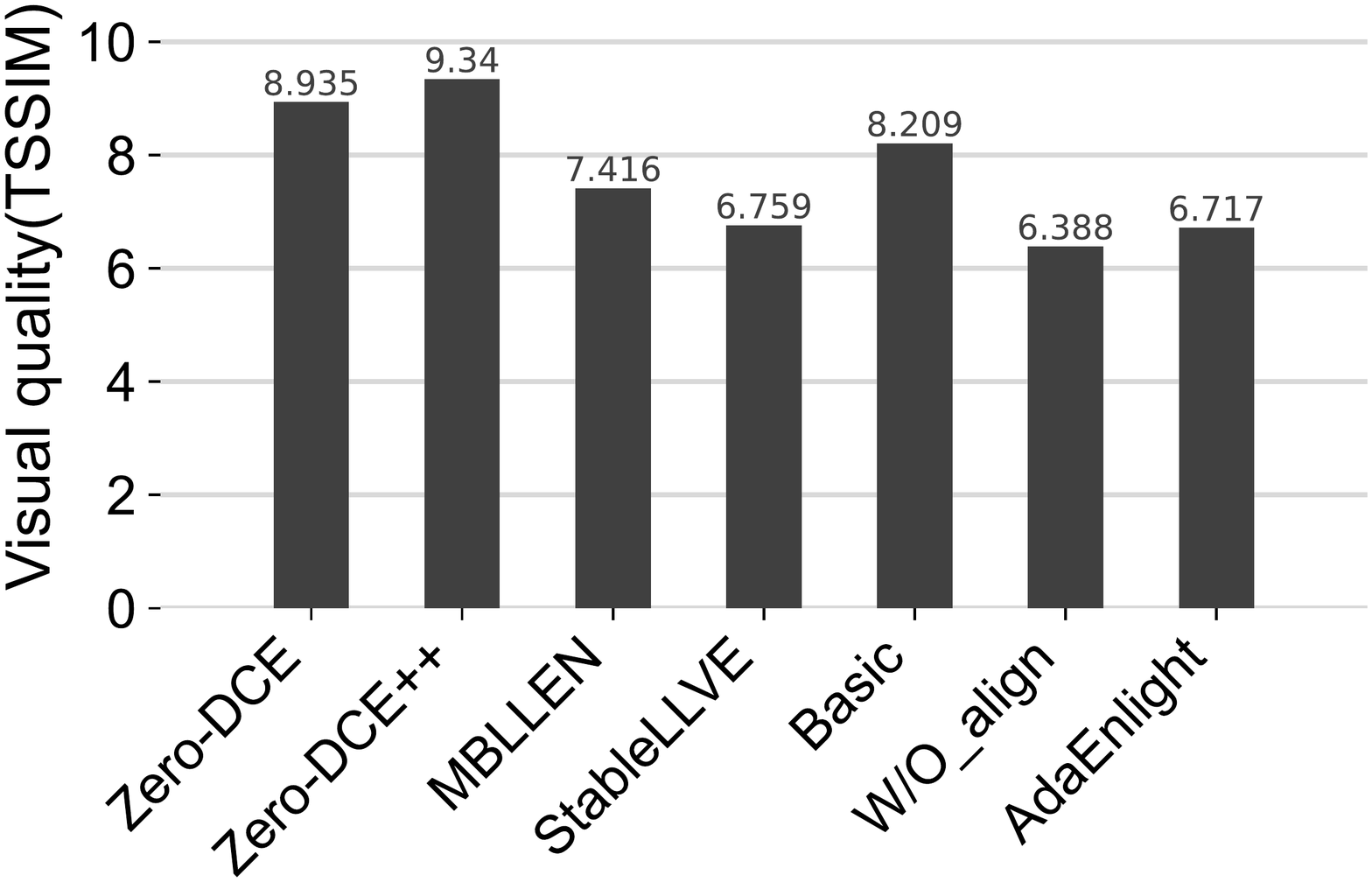}}
   \subfloat[\rev{Static cam, static obj}]{\label{fig::m_scene4}
   \includegraphics[width=0.245\textwidth]{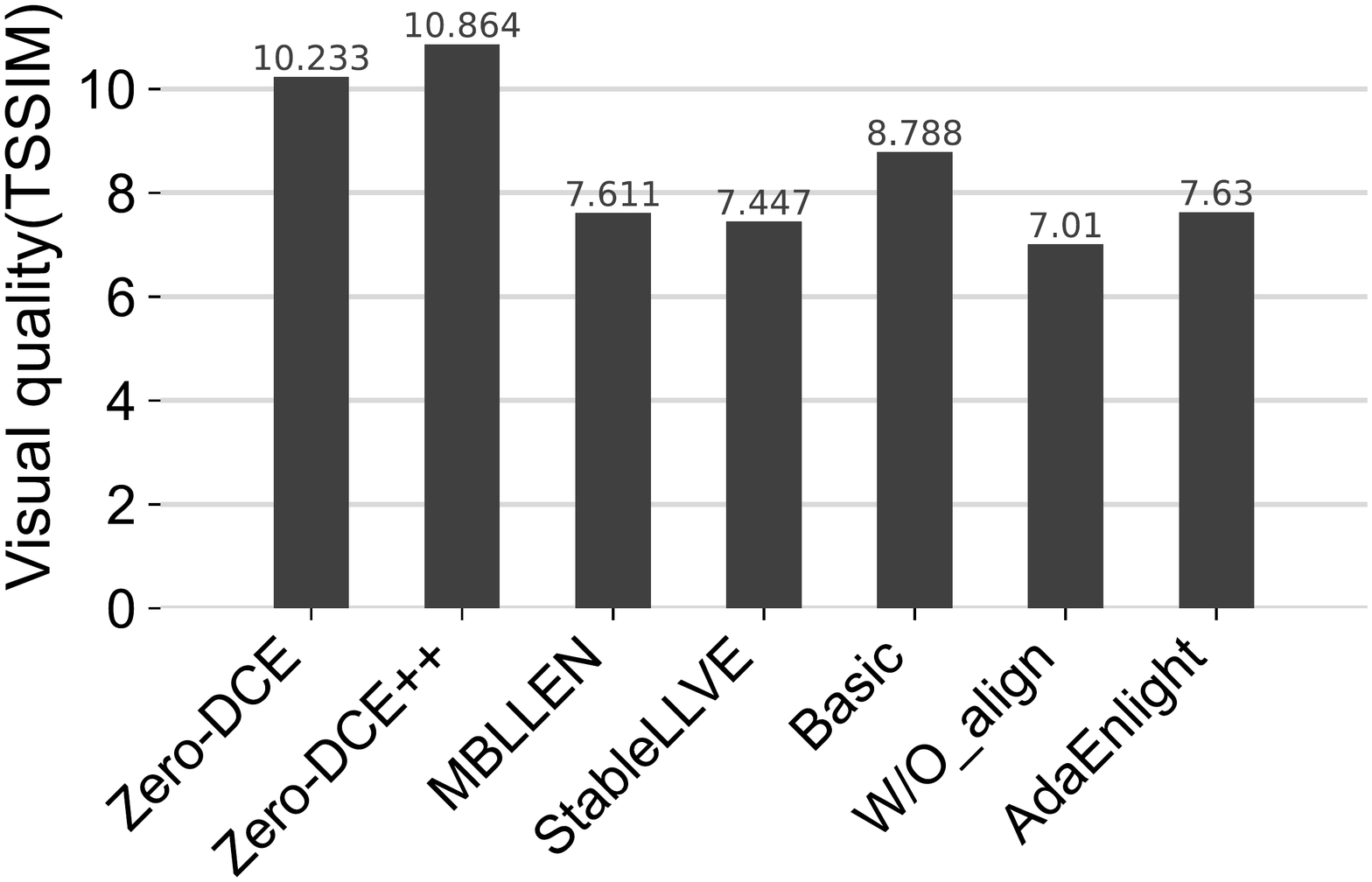}}
\caption{\rev{Temporal stability quantified by TSSIM and MABD, in four mobile scenarios. Scenario A: moving camera and moving objects \{(a),(e)\}, Scenario B: moving camera and static objects \{(b),(f)\}, Scenario C: static camera and moving objects \{(c),(g)\}, and Scenario D: static camera and static objects \{(d),(h)\}. A high TSSIM and a low MABD mean good temporal stability.}}
\label{fig:scene}
\vspace{-3mm}
\end{figure}

\textbf{Results}.
\figref{fig:scene} compares TSSIM and MABD of different algorithms in all the four scenarios.
\rev{Compared with prior methods, \sysname achieves the best results in (a), (b), (g) and the second best results in (c), (d), (e), (f). 
Furthermore, \sysname outperforms its \textbf{Basic} variant in all the four scenarios, which validates the necessity of the temporal consistency loss.
Compared with the variant \textbf{W/O\_align}, \sysname achieves better results in the TSSIM metric, although \sysname is slightly worse than \textbf{W/O\_align} in the MABD metric. 
The results show that \sysname achieves a better structural-level stability due to the optical flow alignment.} %both Scenario A and Scenario B.
 
%  Figure \ref{fig:scene} compares the performance of \sysname with four baseline methods (see Sec. \ref{sec:baseline}) and a \sysnameposs variant under these four scenarios. %without considering temporal consistency loss.
% First, \sysname achieves the best overall output visual quality under four scenarios. %both Scenario A and Scenario B.
% Specifically, the visual quality metric MABD of \sysname is the best (\ie $6.388 \sim 9.981$) in all four scenarios, compared to the other four baseline methods and the variant. 
% Regarding the TSSIM metric, \sysname outperforms the two image-specific baselines, \ie Zero-DCE and Zero-DCE++, across four scenarios.  
% And it also surpasses the three video-based baselines, \ie MBLLEN, StableLLVE, and AdaEnlight\_no, in Scenario A and B with moving cameras.

% \textbf{Summary}. The experimental outcomes demonstrate that the output visual performance of \sysname is the most robust in diverse mobile scenarios.
% It is because \sysname accounts for the temporal consistency, \ie the exposure consistency loss and color consistency loss, between two video frames when designing the loss function to train the video enhancement model.

\subsubsection{Performance on Downstream On-device Tasks: Face Detection}
% TODO: which dataset
% 画个示意图.下游任务1：人脸探测，Dark Face数据集，引用：https://ieeexplore.ieee.org/document/9049390，测试算法，DSFD，引用，https://arxiv.org/abs/1810.10220。数据集包含了6000张夜间场景图像，以及对应的人脸位置标注。我们采用DSFD算法对增强前后的图像进行人脸探测，结果显示增强后的图像对人脸探测算法更加友好。
% 目前尚不存在低照度的人脸探测视频数据集

This experiment shows the performance gains of \sysname for downstream vision tasks such as face detection.

\textbf{Setups}.
We use the widely used face detection algorithm, DSFD~\cite{bib:cvpr2019:li}, to detect human faces in the raw low-light frames and those enhanced by Zero-DCE, Zero-DCE++, and \sysname, respectively.
We use the dark face~\cite{bib:top20:yang} dataset to simulate the raw low-light frames.
The experiment was conducted \rev{on Raspberry Pi 4B (with CPU, device 2)} and we measure the precision and recall of face detection.

% \rev{The platform is (device 2) using CPU processor.}

\begin{figure*}[t]
  \centering
\includegraphics[height=.22\textwidth]{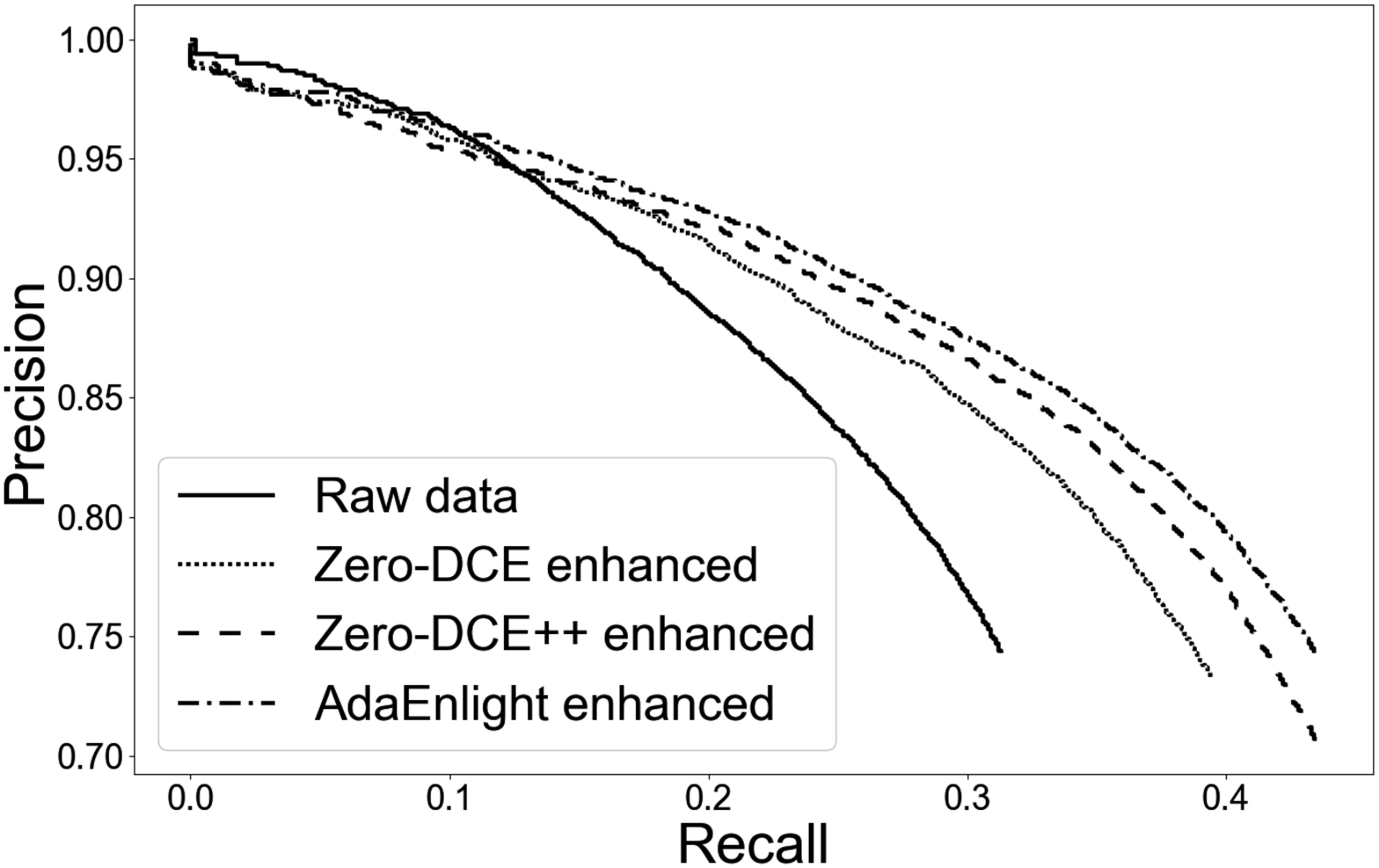}
  \caption{Performance comparison for the downstream on-device video application of mobile face detection.}
  \label{fig_ficial}
  \vspace{-3mm}
\end{figure*} 

\textbf{Results}.
\figref{fig_ficial} compares the precision-recall trade-off of different algorithms.
\sysname shows the highest area under the precision-recall curve, representing the best recall and the highest precision.
Specifically, the high precision and recall relate to a low false-positive rate and a low false-negative rate. 
We can conclude that the enhanced video frames can improve facial detection performance than the raw low-light frame \sysname brings better facial detection benefits than Zero-DCE and Zero-DCE++. 

%High scores for both show that the facial detector based on the frame enhanced by \sysname outperforms the raw frame and other baselines.

%This section compares the visual enhancement benefits of \sysname and two baselines (\ie Zero-DCE and Zero-DCE++) to the downstream on-device application, \ie facial detection.

%%%%%%%%%%
% PC上专业软件：Adobe Premiere Pro，实时，专家依赖且为后处理（Video post-processing，指拍摄完成后再进行处理）
% 手机软件：iPhone 13 camera，需要硬件权限(camera support)，端侧进行，在较暗场景下，时延1s+(照片拍摄，视频拍摄未提供该功能)
%%%%%%%%%%

\begin{table}[t]
\scriptsize
\caption{Comparison of \sysname with other commercial solutions.}
\vspace{-3mm}
\begin{tabular}{|c|c|c|c|c|c|}
\hline
\textbf{Solutions} & \textbf{Special hardware} & \textbf{Enhancement} & \textbf{Internet} & \textbf{Latency} & \textbf{Usage shortcoming} \\ \hline
\textbf{Night vision camera} & Infrared camera & On-device & None & Real-time & Lost color \\ \hline
\textbf{Adobe premiere pro} & None & PC & None & 0.04s/frame & Expert only \\ \hline
\textbf{Baidu contrast enhance API} & None & Cloud server & Dependent & 1.8s/frame & Unstable \\ \hline
\textbf{iPhone 13 camera} & None & On-device & None & 1.3s/image & No video support \\ \hline
\textbf{AdaEnlight} & None & On-device & None & 0.03s/frame & None \\ \hline
\end{tabular}
\label{tb_exp_comp_sys}
\end{table}
% TODO: how are the numbers derived? by our own test? need more details here
%　不同平台的测试均由自行测试得出

\begin{figure*}[t]
  \centering
   \subfloat[On-device video enhancement]{
   \includegraphics[height=0.115\textwidth]{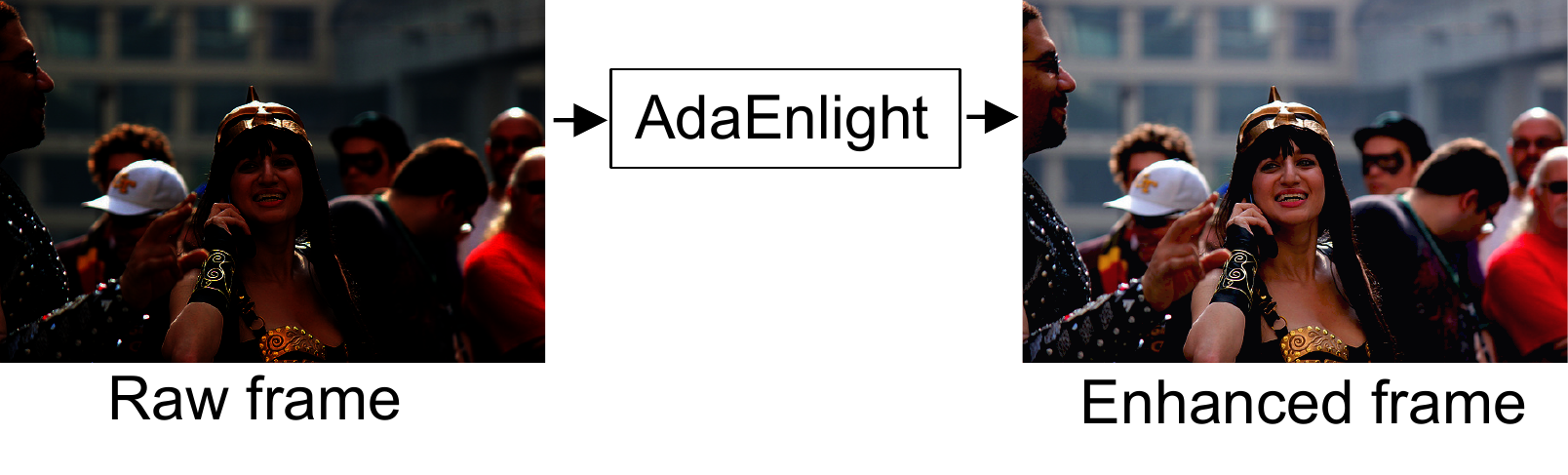}}
   \quad
   \subfloat[Remote video enhancement after internet streaming]{
   \includegraphics[height=0.115\textwidth]{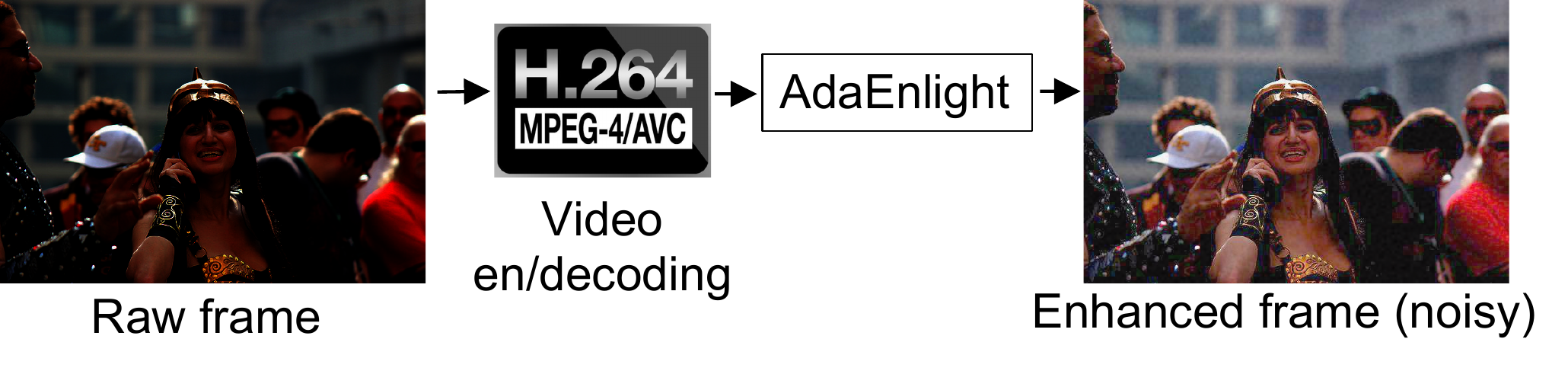}}
   %\hfill
   \vspace{-3mm}
\caption{Visual quality comparison of the enhanced frame before and after commercial video encoding technique.}
\label{fig:sys_encode}
\vspace{-3mm}
\end{figure*}

\subsubsection{Comparison with Commercial Low-light Enhancement Solutions.}
\label{sec:exp_sys}
\tabref{tb_exp_comp_sys} summarizes \sysnameposs main characteristics compared with other commercial low-light enhancement solutions.
\sysname requires no special devices, and conducts video enhancement without network dependency.
%\TODO{how are the numbers derived? by our own test? need more details here}

\figref{fig:sys_encode} illustrates that the output visual quality of locally enhanced video, before encoding (\eg H.264 encoding), is better than the remotely enhanced one, with the intermediate encoding/decoding interference.
The video encoding (\eg H.264 encoding) technique is widely used to stream video over the internet.
It verifies \rev{the \textit{visual benefit of on-device video enhancement}, demonstrating the need for on-device video enhancement.} 
%\TODO{there is no quantitative measure here, not sure if necessary}
Specifically, the natural image quality metric (NIQE) value of the locally enhanced and remotely enhanced video frame is 3.0 and 3.7, respectively.
The smaller the NIQE score, the higher the quality.
% TODO: there is no quantitative measure here, not sure if necessar
% 评价指标NIQE，得分越小代表质量越高
% 压缩前增强质量：3.00
% 压缩后增强质量：3.72

\subsection{System Performance}
\label{sec:exp_system}
This subsection presents the evaluations of \sysname in terms of various system performance metrics.

\subsubsection{Adaptation to Dynamic Energy Budgets of Mobile Platforms}
\label{sec:exp_budget}

\begin{table}[t]
\centering
\scriptsize
\caption{\sysnameposs performance with diverse mobile energy budgets \rev{on Honor 9 (with CPU, device 1)}.}
\vspace{-3mm}
\begin{tabular}{|c|c|c|c|c|c|c|ccc|}
\hline
\multicolumn{1}{|l|}{\multirow{2}{*}{\textbf{No.}}} & \multirow{2}{*}{\textbf{\begin{tabular}[c]{@{}c@{}}Diverse \\ energy supply\end{tabular}}} & \multirow{2}{*}{\textbf{\begin{tabular}[c]{@{}c@{}}Power\\ (W)\end{tabular}}} & \multirow{2}{*}{\textbf{\begin{tabular}[c]{@{}c@{}}Energy per \\ frame(mJ)\end{tabular}}} & \multirow{2}{*}{\textbf{\begin{tabular}[c]{@{}c@{}}Power cost \\ per video (mAh)\end{tabular}}} & \multirow{2}{*}{\textbf{\begin{tabular}[c]{@{}c@{}}Visual quality\\ (VMAF)\end{tabular}}} & \multirow{2}{*}{\textbf{\begin{tabular}[c]{@{}c@{}}Latency\\ per frame (ms)\end{tabular}}} & \multicolumn{3}{c|}{\textbf{Optimal Hyperparameters}} \\ \cline{8-10} 
\multicolumn{1}{|l|}{} &  &  &  &  &  &  & \multicolumn{1}{c|}{\textbf{$\theta_f$}} & \multicolumn{1}{c|}{\textbf{$\theta_l$}} & \textbf{$\theta_d$} \\ \hline
1 & 89\% & 4.1 & 200 & 1.7 & 99.8 & 49 & \multicolumn{0}{c|}{0} & \multicolumn{1}{c|}{0} & 1 \\ \hline
2 & 78\% & 3.8 & 191 & 1.5 & 93.2 & 47 & \multicolumn{1}{c|}{1} & \multicolumn{1}{c|}{1} & 1 \\ \hline
3 & 69\% & 3.7 & 182 & 1.4 & 90.2 & 45 & \multicolumn{1}{c|}{1} & \multicolumn{1}{c|}{1} & 1/2 \\ \hline
4 & 50\% & 3.5 & 175 & 1.3 & 86.2 & 42 & \multicolumn{1}{c|}{2} & \multicolumn{1}{c|}{2} & 1 \\ \hline
5 & 48\% & 3.2 & 138 & 1.1 & 73.7 & 37 & \multicolumn{1}{c|}{3} & \multicolumn{1}{c|}{1} & 1/2 \\ \hline
6 & 32\% & 3.2 & 93 & 0.8 & 72.1 & 30 & \multicolumn{1}{c|}{3} & \multicolumn{1}{c|}{3} & 1/3 \\ \hline
\end{tabular}
\label{tb_energy}
\end{table}

\begin{table}[t]
\centering
\scriptsize
\caption{\rev{\sysnameposs performance with diverse mobile energy budgets on Jetson Nano (with GPU, device 3)}.}
\vspace{-3mm}
\begin{tabular}{|c|c|c|c|c|c|c|ccc|}
\hline
\multicolumn{1}{|l|}{\multirow{2}{*}{\rev{\textbf{No.}}}} & \multirow{2}{*}{\rev{\textbf{\begin{tabular}[c]{@{}c@{}}Diverse \\ energy supply\end{tabular}}}} & \multirow{2}{*}{\rev{\textbf{\begin{tabular}[c]{@{}c@{}}Power\\ (W)\end{tabular}}}} & \multirow{2}{*}{\rev{\textbf{\begin{tabular}[c]{@{}c@{}}Energy per \\ frame(mJ)\end{tabular}}}} & \multirow{2}{*}{\rev{\textbf{\begin{tabular}[c]{@{}c@{}}Power cost \\ per video (mAh)\end{tabular}}}} & \multirow{2}{*}{\rev{\textbf{\begin{tabular}[c]{@{}c@{}}Visual quality\\ (VMAF)\end{tabular}}}} & \multirow{2}{*}{\rev{\textbf{\begin{tabular}[c]{@{}c@{}}Latency\\ per frame (ms)\end{tabular}}}} & \multicolumn{3}{c|}{\rev{\textbf{Optimal Hyperparameters}}} \\ \cline{8-10} 
\multicolumn{1}{|l|}{} &  &  &  &  &  &  & \multicolumn{1}{c|}{\rev{\textbf{$\theta_f$}}} & \multicolumn{1}{c|}{\rev{\textbf{$\theta_l$}}} & \rev{\textbf{$\theta_d$}} \\ \hline
\rev{1} & \rev{95\%} & \rev{3.3} & \rev{82.3} & \rev{0.68} & \rev{100} & \rev{25} & \multicolumn{0}{c|}{\rev{0}} & \multicolumn{1}{c|}{\rev{0}} & \rev{1} \\ \hline
\rev{2} & \rev{81\%} & \rev{3.2} & \rev{66.2} & \rev{0.55} & \rev{96.9} & \rev{20} & \multicolumn{1}{c|}{\rev{1}} & \multicolumn{1}{c|}{\rev{1}} & \rev{1} \\ \hline
\rev{3} & \rev{76\%} & \rev{3.2} & \rev{57.6} & \rev{0.48} & \rev{94.2} & \rev{18} & \multicolumn{1}{c|}{\rev{4}} & \multicolumn{1}{c|}{\rev{1}} & \rev{1} \\ \hline
\rev{4} & \rev{40\%} & \rev{3.0} & \rev{48.1} & \rev{0.40} & \rev{75.1} & \rev{12} & \multicolumn{1}{c|}{\rev{0}} & \multicolumn{1}{c|}{\rev{0}} & \rev{1/2} \\ \hline
\end{tabular}
\label{tb_energy_nano}
\end{table}

This experiment evaluates how \sysname optimizes visual quality under different energy budget.
Given a specific energy budget, \sysnameposs controller actively adjusts its behaviors (\ie the frame-level computation reuse, the layer-level computation reuse, and the frame resolution) by re-selecting the suitable hyperparameters $\theta_f$, $\theta_l$, and $\theta_d$.

\textbf{Setups}. We leverage the NightOwls video dataset (task 5) for evaluation. 
We normalize the resolution of all video frames into the same size, \ie 270P, for fair comparison.
And we sample \rev{diverse stations of energy supply (\ie the remaining battery) on Honor 9 (with CPU, device 1) and Jetson Nano (with GPU, device 3)}, to evaluate \sysnameposs hyperparameter selection on computation reuse and down-sampling.
The computation reuse and down-sampling operations for energy conservation will bring about a decline in visual quality.
Thus, we adopt a full-reference video quality metric, \ie video multimethod assessment fusion (VMAF) to quantify the quality decrease of computation reuse in different levels.
%, and the reference index VMAF can effectively measure the quality of the distorted video compared with the original video. 
%And we set the video stream as three frame rates, \ie 10FPS, 20FPS, and 30FPS, to test the.
%%%%%%%%%%
%TODO: more details here: which dataset for test; how you set the energy supply, how you measure the energy consumption, how many tests you did, why you use another metric for visual quality
% 能源这块我不太清楚怎么描述
%我们采用了NightOwls数据集进行测试，在测试时，视频帧尺寸被resize至270P,　帧率根据需求调节至10,20,30帧，在AGX上进行测试时，尺寸被resize至1080P，帧率同上
%%%%%%%%%%
%　指标换用VMAF，我们认为不进行计算复用时视频质量是最高的，进行计算复用会带来视觉质量的下降，而有参考指标VMAF可以有效的衡量失真视频较原视频的质量下降程度，因此此处采用VMAF指标量化评价计算复用前后的视频质量关系
%\TODO{more details here: which dataset for test; how you set the energy supply, how you measure the energy consumption, how many tests you did, why you use another metric for visual quality?}

\textbf{Results}.
\tabref{tb_energy} and \tabref{tb_energy_nano} summarize the performance and the corresponding hyperparameters with different energy supply.
We make the following observations.
First, \sysname can select suitable hyperparameters with various energy budgets.
For example, \rev{at $78\%$ energy supply on Honor 9 (with CPU, device 1)}, the optimal hyperparameter is to reuse 1 frame and 1 layer without down-sampling. 
At $48\%$ energy supply, it becomes reusing 3 frames and 3 layers and down-sampling by $1/3$.
Second, different hyperparameters setups of $\theta_f$, $\theta_l$, and $\theta_d$ can dynamically tune the trade-off between the output video quality and the energy consumption.
Third, the computation reuse and down-sampling behaviors can further reduce the execution latency of the video enhancement model, ranging from $49ms$ to $30ms$ per frame. 
Therefore, all these hyperparameters ensure real-time constraints. 
\rev{Finally, for completeness, we run \sysname on Honor 9 (device 1, with $3200$ mAH battery) with different computation reuse parameters to simulate diverse user preference on visual quality  and energy consumption. 
Experiments show that \sysname will drain the battery in $2.5h \sim 4h$ under these settings.}

% \textbf{Summary}. \sysname is able to automatically select the proper
% hyperparameters that meet various budgets on visual quality and energy constraints. Besides, we also verify \sysnameposs fidelity for dynamic computation adaptation, which yields significant energy savings. 

\begin{figure}[t]
  \centering
   \subfloat[Latency per frame]{\label{fig:fps_latency_frame}
   \includegraphics[width=0.2\textwidth]{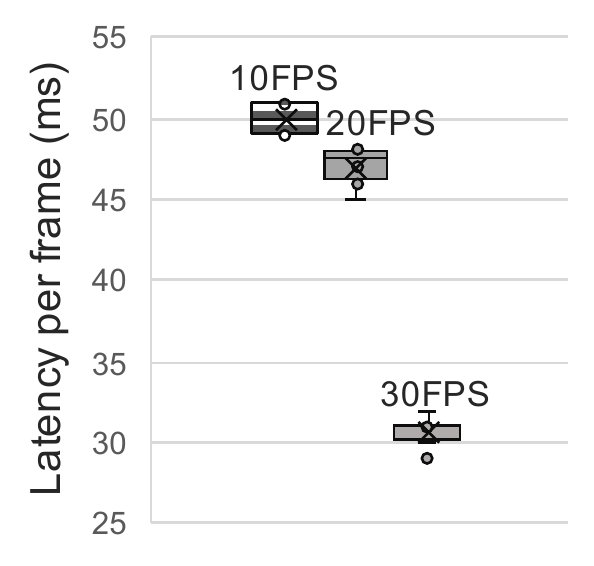}}
   \hfill
   \subfloat[Latency per video]{\label{fig:fps_latency_video}
   \includegraphics[width=0.2\textwidth]{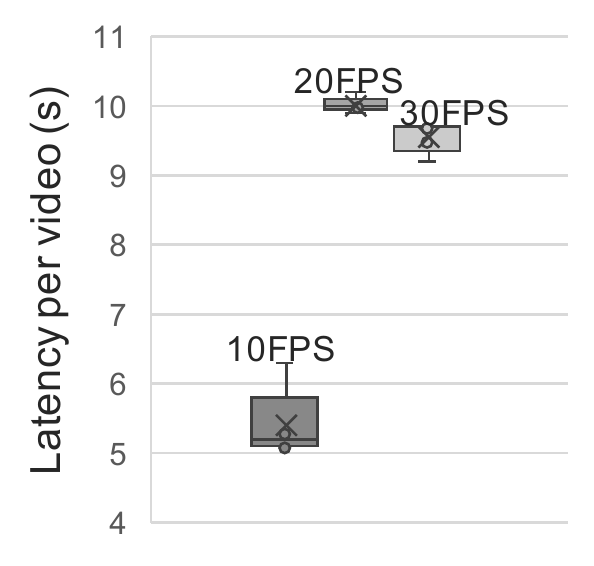}}
   \hfill
   \subfloat[Power]{\label{fig:fps_power}
   \includegraphics[width=0.2\textwidth]{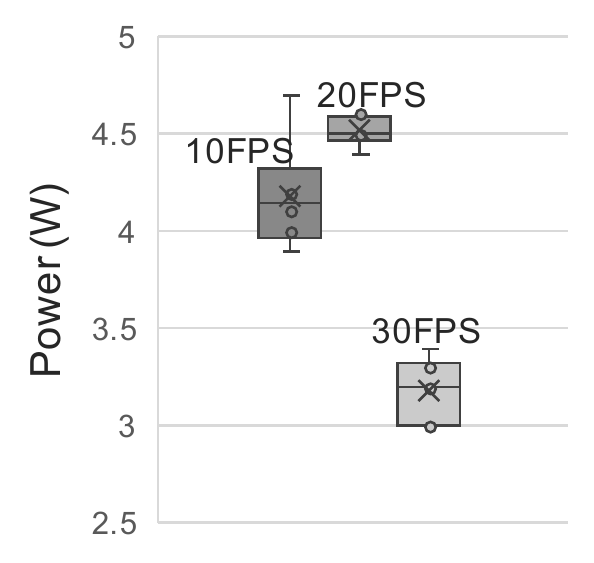}}
    \hfill
   \subfloat[Energy cost per frame]{\label{fig:fps_energy}
   \includegraphics[width=0.2\textwidth]{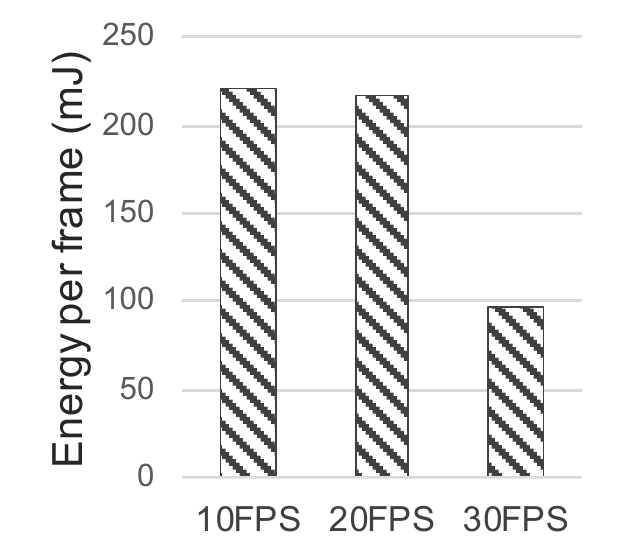}}
   \vspace{-3mm}
\caption{\sysnameposs performance across diverse inputs with varying video frame rates (FPS), \ie 10 FPS, 20FPS, and 30FPS.}
\label{fig:fps}
\vspace{-3mm}
\end{figure}

\begin{figure*}[t]
  \centering
  \includegraphics[height=0.29\textwidth]{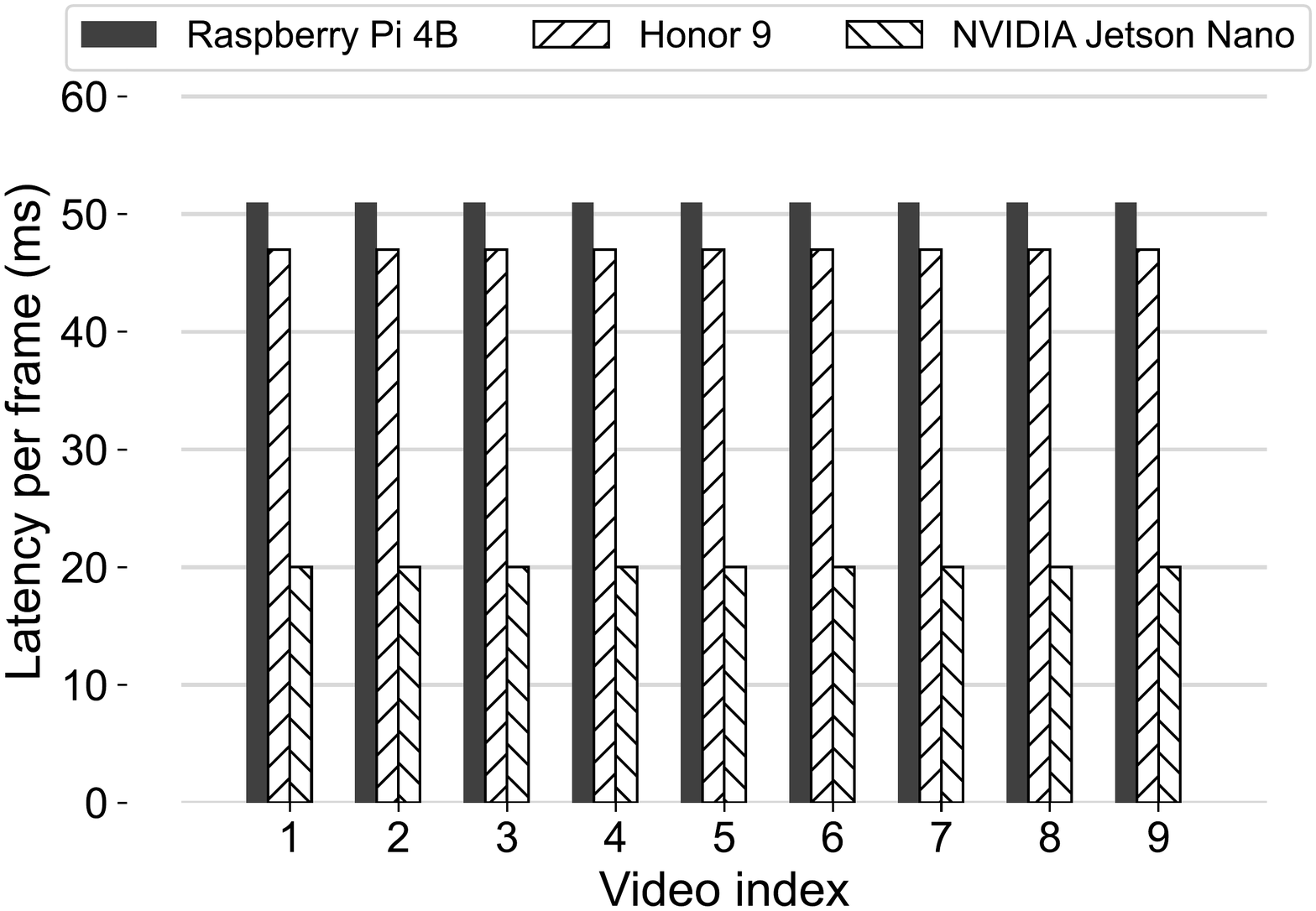}
  \caption{\rev{\sysnameposs performance on three typical camera-embedded platforms.}}
  \label{fig_device}
  \vspace{-3mm}
\end{figure*} 

\subsubsection{Adaptation to Input Video Frame Rates}
We now test \sysnameposs performance over diverse  video streams.

\textbf{Setups}.
We use $20$ video clips of $10s$ duration with three different video frame rates, \ie 10 frame per second (FPS), 20FPS, and 30 FPS.
We test the \sysnameposs execution latency on \rev{Honor 9 (with CPU, device 1)}.
We use the digital power monitor of Monsoon AAA10F to power the smartphone and measure its energy cost.

%In this part of the experiments, we use the real-world camera-embedded mobile platform, \ie Honor 9 smartphone, powered by the $3200 mAh$ battery.
%And we leverage the , to test the \sysnameposs energy cost and power through its external power input.

\textbf{Results}.
\figref{fig:fps} shows the performance of \sysname across diverse input video streams.
First, \sysname achieves real-time processing for all frame rates.
Specifically, for the 10FPS, 20FPS, and 30FPS input video streams, the average processing latency per frame is $50ms$, $47ms$, and $31ms$, respectively.
Furthermore, the average delay of \sysname to process the entire video clip with these three frame rates is $\leq 10s$, which is smaller than the streaming speed of the input video, thereby ensuring real-time video stream enhancement.
Second, the energy consumption of \sysname on the Honor 9 platform is adjustable for diverse input videos with the same energy budget.
Specifically, the energy cost of \sysname for the 10FPS, 20FPS, and 30FPS input videos is $227 mJ$, $158 mJ$, and $97mJ$, respectively. 
For video streams over 30FPS, we can further reduce the execution latency and energy consumption by adjusting the computation reuse and frame resolution hyperparameters.

% \textbf{Summary}. \sysname can guarantee real-time video processing (\eg in millisecond level).
% %for videos with $\leq 30FPS$ frame rate. 
% Also, the \sysnameposs energy consumption is adjustable to satisfy the energy budgets for diverse video frame rates.

%%%%%%%%%%
% TODO: dataset?
%%%%%%%%%%

% \begin{figure*}[t]
%   \begin{minipage}{0.43\linewidth}
%   \centering
%   \includegraphics[height=0.63\textwidth]{image/various_device.pdf}
%   \caption{\sysnameposs performance on three typical mobile camera-embedded platforms.}
%   \label{fig_device}
%   \end{minipage}
%   %\hfill
%   \quad \quad 
%   \begin{minipage}{0.43\linewidth}
%   \centering
% \includegraphics[height=.63\textwidth]{image/PR_figure.pdf}
%   \caption{Performance comparison for the downstream on-device video task, \ie facial detection.
%   }
%   \label{fig_ficial}
%   \end{minipage}
% \end{figure*} 

\subsubsection{Performance on Diverse Mobile Platforms}
\label{subsec:exp_device}
This experiment evaluates \sysname on different  devices.

% 替换掉AGX
\textbf{Setups}.
We test three platforms: Honor 9 \rev{(using CPU, device 1)}, Raspberry Pi 4B \rev{(with CPU, device 2)}, and \rev{NVIDIA Jetson Nano (with GPU, device 3)}. 
Different devices have diverse resource availability (\eg computation throughput and memory hierarchy) and \rev{frameworks (Pytorch mobile, NCNN, Pytorch)}, which lead to different execution latency.
%Raspberry Pi 4B and Honor 9 are equipped with mobile CPUs, while AGX has a mobile GPU processor.
%\TODO{dataset?}
%我们采用了NightOwls数据集进行测试，在树莓派和Honor 9上进行测试时，视频帧尺寸被resize至270P,　帧率根据需求调节至10,20,30帧，在AGX上进行测试时，尺寸被resize至1080P，帧率同上
We use the NightOwls dataset to extract video samples as the input of \sysname.
We use nine input videos of 10s at 270P, and a frame rate of 10FPS for testing to obtain the overall processing delay.

\textbf{Results}.
\figref{fig_device} compares the \sysname performance on these three different platforms.
On all three resource-constrained mobile devices, \sysname achieves real-time processing with $20ms \sim 51.5ms$ execution latency.

%\textbf{Summary}. \sysname demonstrate the real-time processing performance on different resource-constrained mobile platforms with CPU/GPU processors. And GPUs in high-end mobiles can further accelerate the execution.

\subsection{Ablation Studies}
\label{subsec:exp:bench}

This section evaluates the impact of various setups on the \sysnameposs performance.

\begin{figure*}[t]
  \centering
   \subfloat[$Ite=1$, Zero-DCE++]{
   \includegraphics[width=0.2\textwidth]{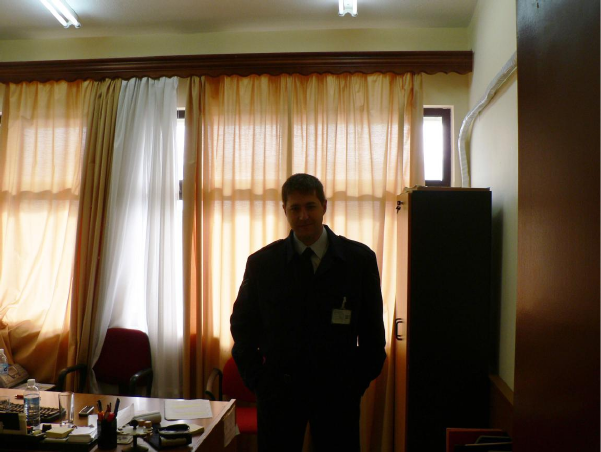}}
   \hfill
   \subfloat[$Ite=8$, Zero-DCE++]{
   \includegraphics[width=0.2\textwidth]{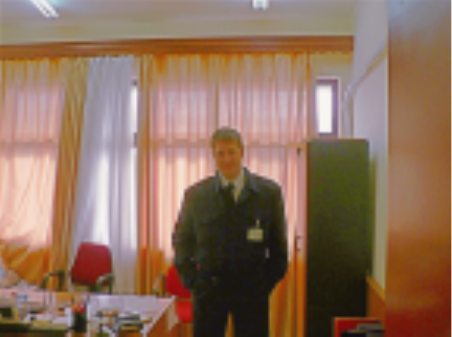}}
  \hfill
  \subfloat[\rev{$Ite=1$, \sysname}]{
  \includegraphics[width=0.2\textwidth]{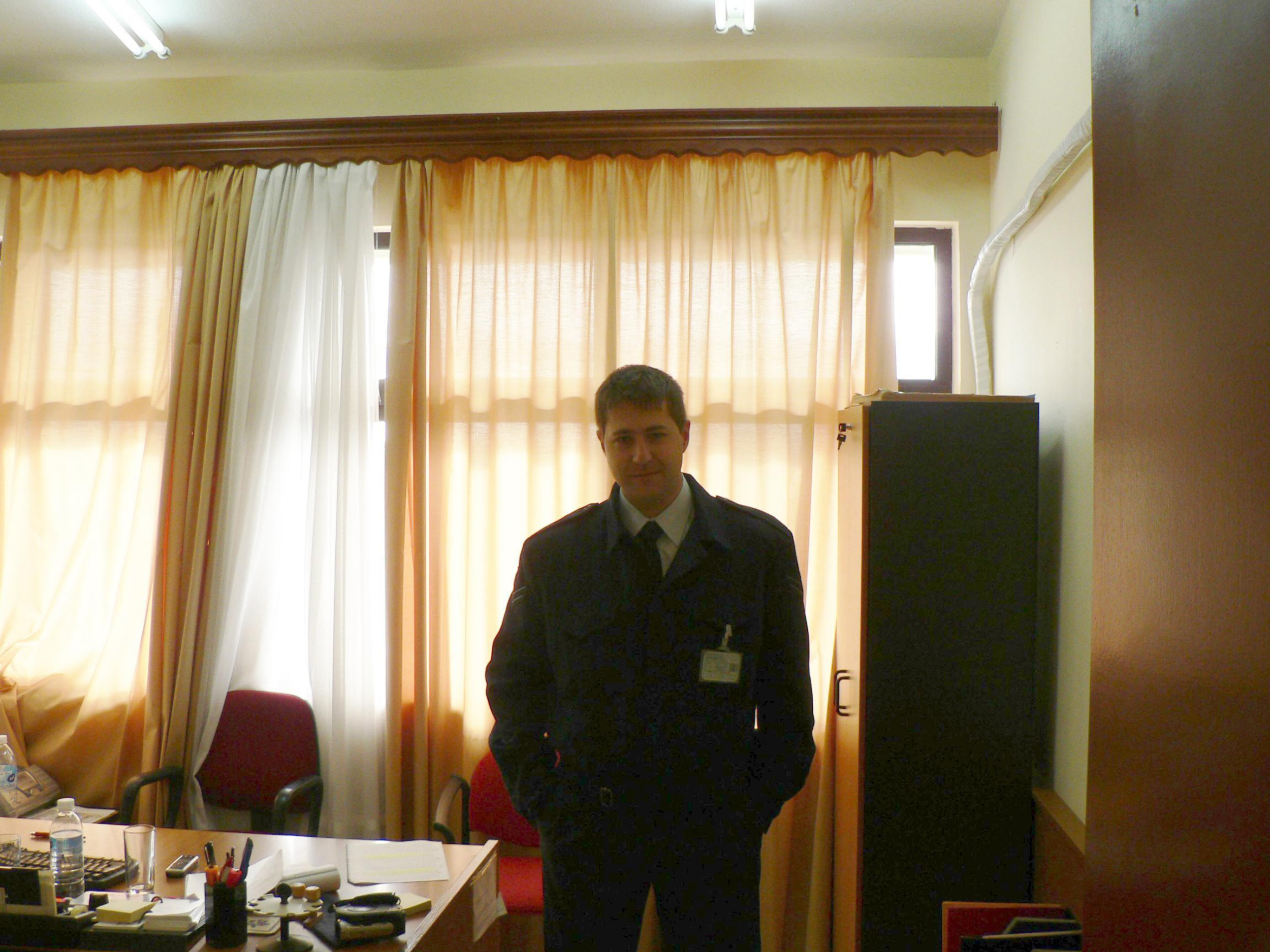}}
  \hfill
  \subfloat[\rev{$Ite=8$, \sysname}]{
  \includegraphics[width=0.2\textwidth]{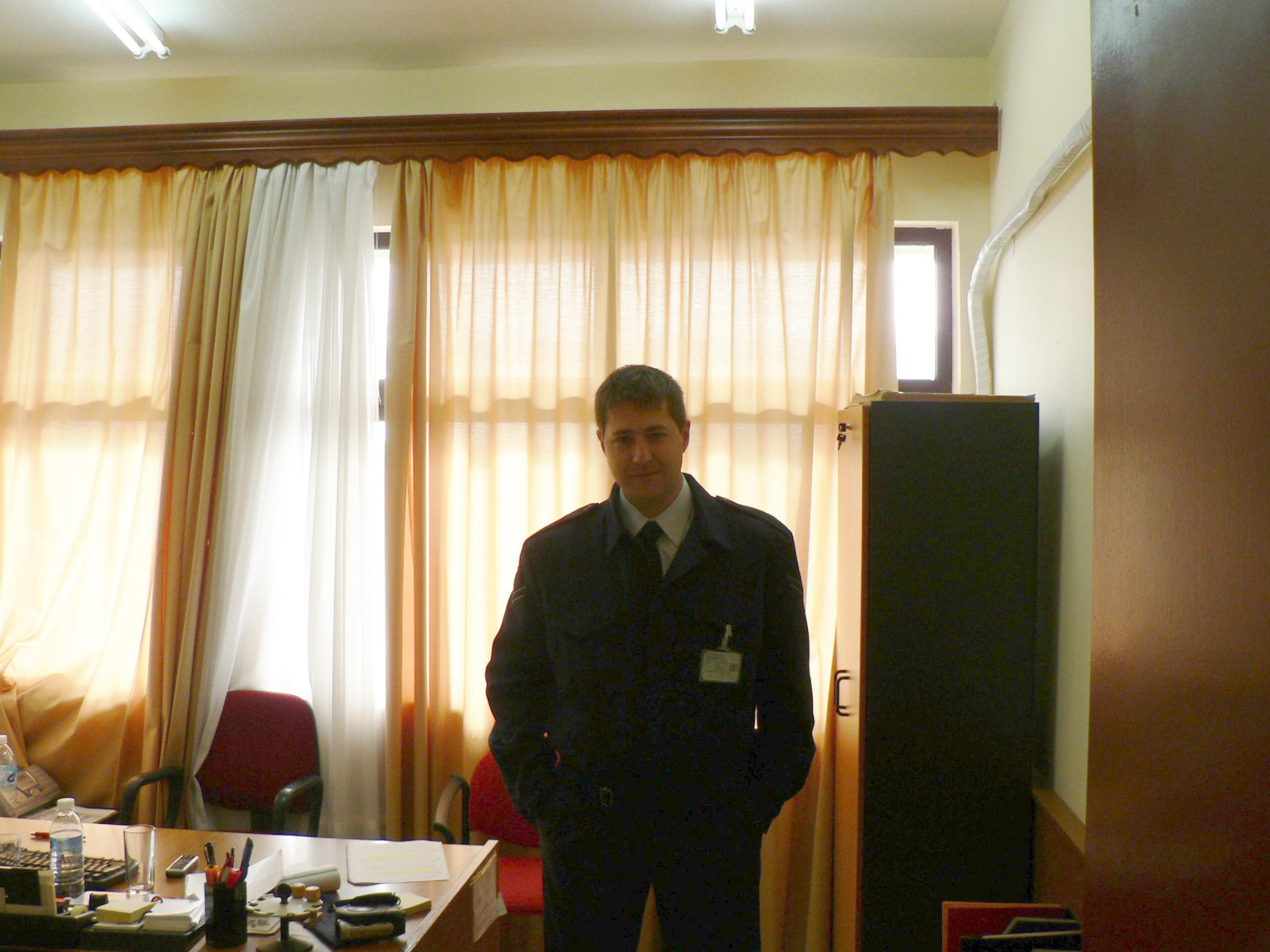}}
  \vspace{-3mm}
\caption{\rev{Impact of iteration numbers $Ite$ on  visual quality by Zero-DCE++ and our Gamma correction-based curve.}}
\label{fig:iterations}
\vspace{-3mm}
\end{figure*}

\begin{figure}[t]
  \centering
   \subfloat[Iteration number]{\label{fig_ite_num}
   \includegraphics[width=0.24\textwidth]{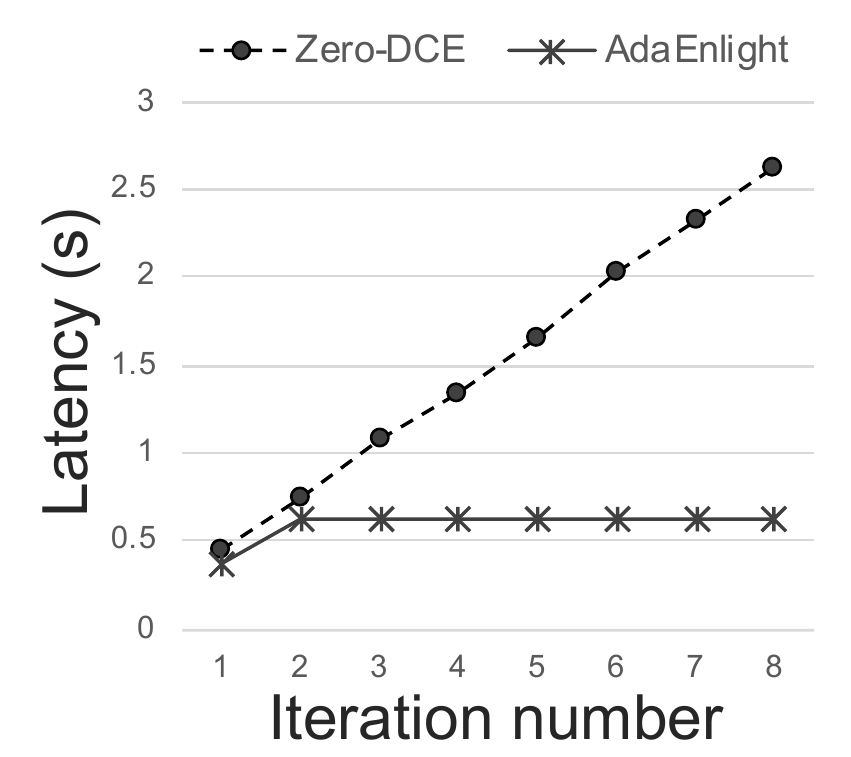}}
   \hfill
   \subfloat[Architecture]{\label{fig_layer_num}
   \includegraphics[width=0.24\textwidth]{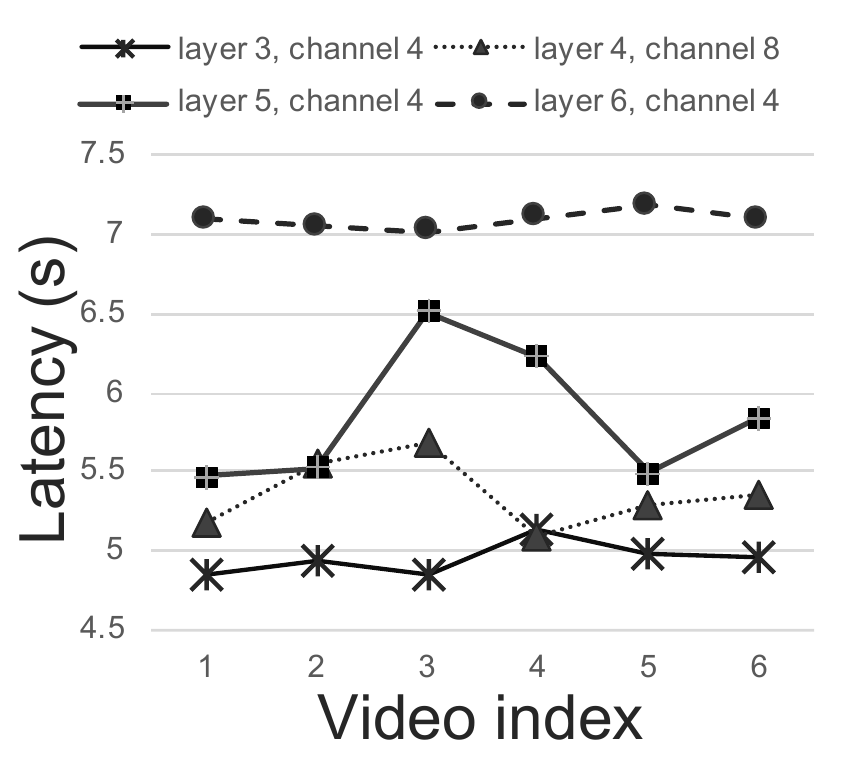}}
   \hfill
   \subfloat[Reuse computation]{\label{fig_reuse}
   \includegraphics[width=0.24\textwidth]{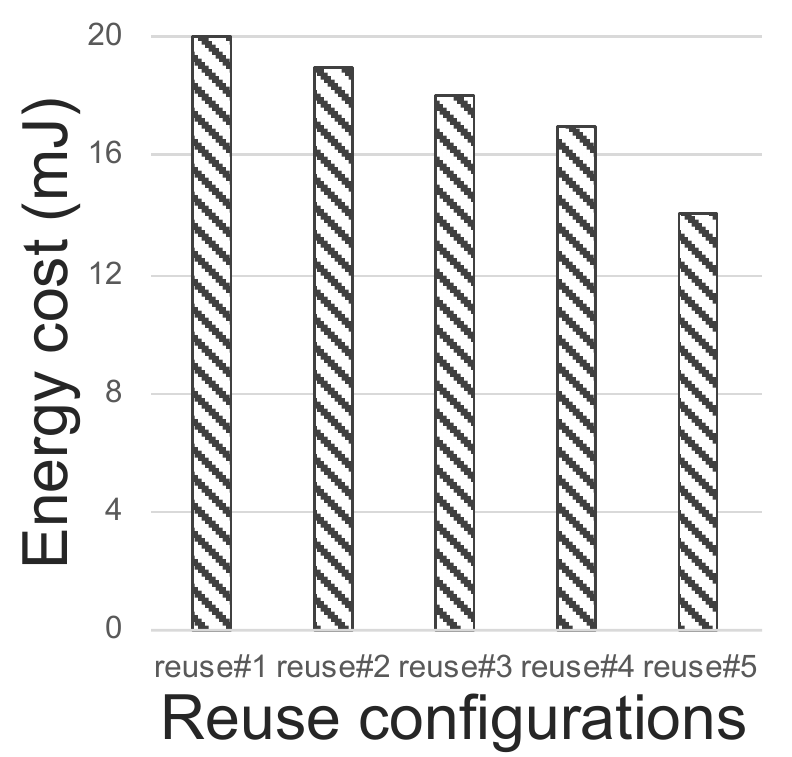}}
   \hfill
   \subfloat[frame resolution]{\label{fig_resolution}
   \includegraphics[width=0.24\textwidth]{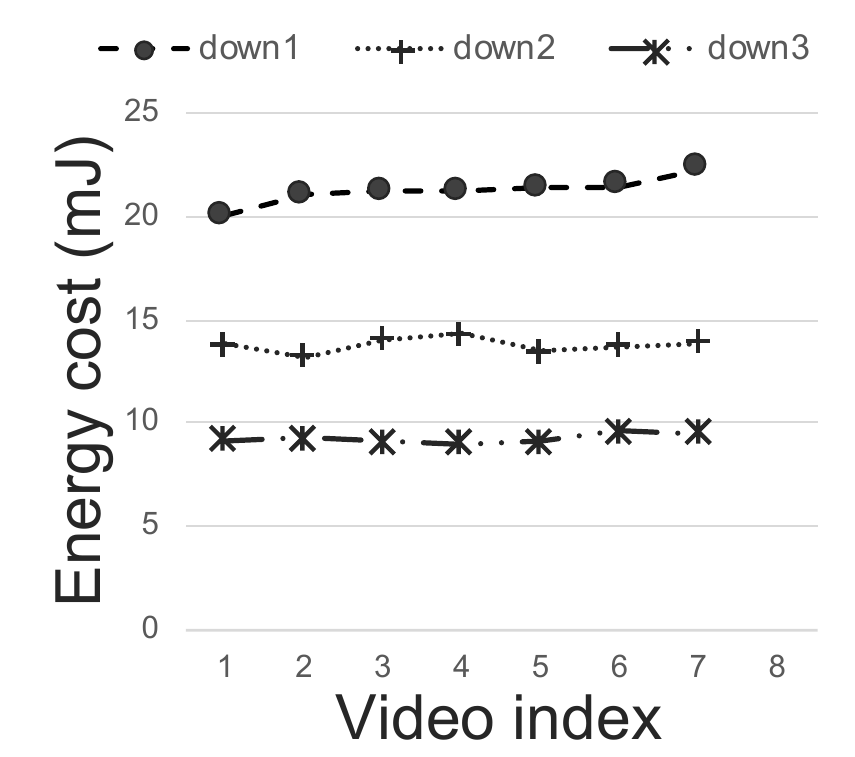}}
\caption{Benchmark of various parameters, \ie (a) iteration number, (b) layer and channel number, (c) frame-level and layer-level computation reuse, and (d) frame resolution, on \sysnameposs performance.}
\label{fig_bench}
\end{figure}

% TODO: which dataset
% NightOwls数据集，尺寸应为270*480
\subsubsection{Iteration Number in Gamma Correction-based Curve}
\label{subsec:exp_ite}
As discussed in \secref{subsec:algo:gamma}, we propose the non-iterative enhancement method to break the latency bottleneck of traditional curve-based image enhancement.
As shown in \figref{fig:iterations}, the visual quality enhanced by \sysname does not notably increase with more iterations.
Thus, we set the iteration number as 1, which is equivalent to a non-iterative model.
\figref{fig_ite_num} compares the execution latency of \sysnameposs enhancement model and Zero-DCE with different iteration numbers in their curve functions. 
\sysname avoids the long latency due to the iterative enhancement process in Zero-DCE.

\subsubsection{Layer and Channel Number}
To tune the architecture hyperparameters (\ie the number of layer and channel) of \sysnameposs enhancement model, we compare the performance of different architecture settings.
We randomly selected six video segmentations from NightOwls dataset with the frame resolution of $270\times{480}$ for testing, each video has a duration of 10S, a frame rate of 10FPS, and a frame resolution of $270\times{480}$.  
\figref{fig_layer_num} illustrates the model's execution latency with four settings. 
All the settings result in the competitive visual quality, while 3 layers with 4 channels has the lowest latency \rev{on Raspberry Pi 4B (with CPU, device 2)}. 
Thus, we set the layer number as 3 and the channel number as 6 in \sysnameposs enhancement model by default.

\subsubsection{Impact of Computation Reuse}
\label{sec:exp_reuse}
% 超参数：对应4.4.1的TODO
As discussed in \secref{subsubsec:algo:behaviour}, due to temporal similarities between adjacent video frames, we can dynamically adjust the model complexity by enforcing computation reuse.
\figref{fig_reuse} illustrates the energy consumption of \sysnameposs video enhancement model \rev{on Honor 9 (with CPU, device 1)}, under different frame and layer reuse settings.
We compare five options, \ie reuse\#1 (frame = 1 layer = 1),  reuse\#2 (frame = 1, layer = 2), reuse\#3 (frame = 1, layer = 3), and reuse\#4 (frame = 2, layer = 2), and reuse\#5 (frame = 2, layer = 3).
The outcomes verify that these computation reuse hyperparameters can effectively tune the model's energy cost.
\rev{It is worth mentioning that such computation reuse also reduces the overall latency by $10\% \sim 50\%$.
For example, given a fixed frame-level computation reuse parameter 5, if we set layer-level computation reuse parameter as 1,2 and 3, the latency decreases by 12.5\%, 35.1\% and 57.6\%, respectively.
However, since \sysname without computation reuse already achieves near real-time processing \ie 50ms per frame (shown in \tabref{tb_compare}), a reduction of $50\%$ in delay does not introduce perceivable improvement in user experience.
Therefore, we mainly focus on the improvement in energy consumption due to computation reuse.}

% \rev{Impact of computation reuse on latency shows a similar result($10\% \sim 50\%$ latency optimization). Considering that the maximum latency optimization of 50\% brings not obvious user experience, we mainly focus on energy consumption.}

\subsubsection{Impact of Frame Resolution}
% 超参数：对应4.4.1的TODO
This experiment verifies the effect of input frame resolution on the energy consumption of \sysnameposs video enhancement, \rev{on Honor 9 (with CPU, device 1)}.
\figref{fig_resolution} compares the energy cost of three down-sampling levels (\ie $1\sim 3$), representing the full resolution, down to $1/2$ resolution, and down to $1/3$ resolution, respectively.
The energy consumption increases linearly with the frame resolution.
Thus, down-sampling the resolution of the input video frame can naturally tune the model's energy demand.

\begin{table}[t]
\centering
\scriptsize
\caption{Cross-time and cross-model energy profiler measurement on Honor 9 (with CPU, device 1).}
\vspace{-3mm}
\begin{tabular}{|cccccc|}
\hline
\multicolumn{6}{|c|}{\textbf{Cross-time evaluation}} \\ \hline
\multicolumn{1}{|c|}{\textbf{\begin{tabular}[c]{@{}c@{}}Prediction at time 1\\ (resource state 1)\end{tabular}}} & \multicolumn{1}{c|}{\textbf{\begin{tabular}[c]{@{}c@{}}Measurement at time 1\\ (resource state 1)\end{tabular}}} & \multicolumn{1}{c|}{\textbf{Error}} & \multicolumn{1}{c|}{\textbf{\begin{tabular}[c]{@{}c@{}}Prediction at time 2\\ (resource state 2)\end{tabular}}} & \multicolumn{1}{c|}{\textbf{\begin{tabular}[c]{@{}c@{}}Measurement at time 2\\ (resource state 2)\end{tabular}}} & \textbf{Error} \\ \hline
\multicolumn{1}{|c|}{0.192mJ} & 
\multicolumn{1}{c|}{0.2mJ} & 
\multicolumn{1}{c|}{4\%} & 
\multicolumn{1}{c|}{0.141mJ} & 
\multicolumn{1}{c|}{0.138mJ} & 
\multicolumn{1}{c|}{1.4\%} \\ \hline
\multicolumn{6}{|c|}{\textbf{Cross-model evaluation}} \\ \hline
\multicolumn{1}{|c|}{\textbf{\begin{tabular}[c]{@{}c@{}}Prediction for model 1\\ (4 layers )\end{tabular}}} & \multicolumn{1}{c|}{\textbf{\begin{tabular}[c]{@{}c@{}}Measurement for model 1\\ (4 layers)\end{tabular}}} & \multicolumn{1}{l|}{\textbf{Error}} & \multicolumn{1}{c|}{\textbf{\begin{tabular}[c]{@{}c@{}}Prediction for model 2\\ (5 layers)\end{tabular}}} & \multicolumn{1}{c|}{\textbf{\begin{tabular}[c]{@{}c@{}}Measurement for model 2\\ (5 layers)\end{tabular}}} & \multicolumn{1}{l|}{\textbf{Error}} \\ \hline
\multicolumn{1}{|c|}{0.285mJ} & 
\multicolumn{1}{c|}{0.27mJ} & 
\multicolumn{1}{c|}{5.6\%} & 
\multicolumn{1}{c|}{0.368mJ} & 
\multicolumn{1}{c|}{0.38mJ} & 
\multicolumn{1}{c|}{3.2\%} \\ \hline
\end{tabular}
\label{tb_exp_general}
\end{table}

\begin{table}[t]
\centering
\scriptsize
\caption{\rev{Cross-time and cross-model energy profiler measurement on Jetson Nano (with GPU, device 3).}}
\vspace{-3mm}
\begin{tabular}{|cccccc|}
\hline
\multicolumn{6}{|c|}{\rev{\textbf{Cross-time evaluation}}} \\ \hline
\multicolumn{1}{|c|}{\rev{\textbf{\begin{tabular}[c]{@{}c@{}}Prediction at time 3\\ (resource state 1)\end{tabular}}}} & \multicolumn{1}{c|}{\rev{\textbf{\begin{tabular}[c]{@{}c@{}}Measurement at time 3\\ (resource state 1)\end{tabular}}}} & \multicolumn{1}{c|}{\rev{\textbf{Error}}} & \multicolumn{1}{c|}{\rev{\textbf{\begin{tabular}[c]{@{}c@{}}Prediction at time 4\\ (resource state 2)\end{tabular}}}} & \multicolumn{1}{c|}{\rev{\textbf{\begin{tabular}[c]{@{}c@{}}Measurement at time 4\\ (resource state 2)\end{tabular}}}} & \rev{\textbf{Error}} \\ \hline
\multicolumn{1}{|c|}{\rev{0.36mJ}} & \multicolumn{1}{c|}{\rev{0.38mJ}} & \multicolumn{1}{c|}{\rev{5.6\%}} & \multicolumn{1}{c|}{\rev{0.541mJ}} & \multicolumn{1}{c|}{\rev{0.574mJ}} & 
\rev{6.1\%} \\ \hline
\multicolumn{6}{|c|}{\rev{\textbf{Cross-model evaluation}}} \\ \hline
\multicolumn{1}{|c|}{\rev{\textbf{\begin{tabular}[c]{@{}c@{}}Prediction for model 1\\ (4 layers )\end{tabular}}}} & \multicolumn{1}{c|}{\rev{\textbf{\begin{tabular}[c]{@{}c@{}}Measurement for model 1\\ (4 layers)\end{tabular}}}} & \multicolumn{1}{l|}{\rev{\textbf{Error}}} & \multicolumn{1}{c|}{\rev{\textbf{\begin{tabular}[c]{@{}c@{}}Prediction for model 3\\ (3 layers)\end{tabular}}}} & \multicolumn{1}{c|}{\rev{\textbf{\begin{tabular}[c]{@{}c@{}}Measurement for model 3\\ (3 layers)\end{tabular}}}} & \multicolumn{1}{l|}{\rev{\textbf{Error}}} \\ \hline

\multicolumn{1}{|c|}{\rev{0.129mJ}} & \multicolumn{1}{c|}{\rev{0.128mJ}} & \multicolumn{1}{c|}{\rev{0.8\%}} & \multicolumn{1}{c|}{\rev{0.090mJ}} & \multicolumn{1}{c|}{\rev{0.093mJ}} & 
\rev{3.3\%} \\ \hline
\end{tabular}
\label{tb_exp_general_gpu}
\end{table}

\subsubsection{Generalization of Energy Profiler}
This experiment demonstrates the cross-time and cross-model generalization performance. 
We refer to the evaluation methodology in nn-Meter~\cite{bib:mobisys21:zhang}.
We use the existing energy profiler trained for Honor 9 \rev{(with CPU, device 1) and Jetson Nano (with GPU, device 3) to predict model energy consumption for four different times and three models. 
We evaluate our energy profiling for CPU at time 1 and time 2 with model 1 and model 2; while for GPU at time 3 and time 4 with model 1 and model 3.}  
At different times, the available execution resources of the platform differ, which will affect the cache hit rate and thus the energy demand.
The three models have different architectures: model 1 has 3 Conv and activation layers, model 2 has 4 Conv and activation layers, \rev{and model 3 has 2 Conv and activation layers.}
As shown in \tabref{tb_exp_general} and \tabref{tb_exp_general_gpu}, the energy profiler achieves \rev{$\geq 93.9\%$} accuracy for cross-time prediction, and \rev{$\geq 94.4\%$} for cross-model prediction. 
\rev{Importantly, the results show that our energy profiler ensures consistent ranking between the estimated and the actual energy cost, which is crucial for the controller to adapt the hyperparameters to the energy demand.}

\rev{\subsection{Summary of Main Experimental Results}}
\rev{We summarize the main results of our evaluations as follows.
\begin{itemize}
    \item \textit{Near real-time processing on diverse mobile devices}.
    \newrev{\sysnameposs enhancement model} achieves near real-time video enhancement on different mobile devices, \eg $51$ ms per frame on Raspberry Pi 4B, $47$ ms per frame on Honor 9, and $20$ ms per frame on NVIDIA Jetson Nano. 
    \newrev{This is because the proposed non-iterative model breaks the latency bottleneck of the state-of-the-art enhancement schemes}.
    %Above experiments prove that \sysname achieves real-time processing on mobile devices
    \item \textit{Better trade-off between processing delay and visual quality}.
    \newrev{\sysnameposs enhancement model} achieves the lower video enhancement latency while retaining competitive visual quality.
    \sysnameposs latency is only 1/4 of Zero-DCE++ \cite{bib:TPAMI21:Li2}, the fastest low-light enhancement scheme. 
    Meanwhile, \sysname still yields high visual quality.
    For example, \sysname achieves a PSNR of $17.232$, which is almost the same as MBLLEN \cite{bib:BMVC18:Lv}, the state-of-the-art video enhancement solution (with a PSNR of $17.229$).
    %Meanwhile, the visual quality of \sysname is still competitive, \ie  on PNSR metric, \sysnameposs 17.232 is similar to 17.229 of the best method(MBLLEN). 
    \sysname consistently outputs stable and high-quality videos on various datasets (LOL, NightOwls, LIME and VV), and different scenarios (static/mobile camera, static/mobile objects).
    \newrev{This is mainly due to the enforced temporal consistency between video.}
    %Furthermore, in video stability comparison, \sysname achieves three best and five second in eight scenes(\secref{exp_video}). 
    %Above experiments prove that \sysname achieves outstanding efficient processing while keep competitive visual quality.
    \item \textit{Adaptive to diverse energy supply}.
    %We evaluate the effectiveness of \sysnameposs system-level performance in \secref{sec:exp_system}. 
    \newrev{\sysnameposs energy-aware controller} is able to adapt the enhancement quality according to the energy budget at runtime.
    As a concrete example, when the energy supply is reduced from 95\% to 40\%, the power consumption of the \sysname is reduced from 0.68mAh to 0.4mAh (see \tabref{tb_exp_general_gpu}).
    \newrev{It proves the necessity of the energy-aware controller in mobile systems.}
    %take No.1 and No.4 as an example, as the energy supply is reduced from 95\% to 40\%,More cases in \secref{sec:exp_system} prove that the \sysname can adapt its enhancement quality according to the energy budget at runtime. 
\end{itemize}
}
\section{Related Work}
\label{sec:related}
%Our work is closely related to the following research.

\subsection{Low-light Image and Video Enhancement}
\label{subsec:related:low-light}

Low-light enhancement improves the perception or interpretability of images and videos captured in dim light \cite{bib:TPAMI21:Li, bib:imwut20:janveja}.
Most prior efforts focus on image enhancement \cite{bib:CVPR20:Guo, bib:TIP16:Guo, bib:TCE07:Ibrahim, bib:TIP97:Jobson, bib:TIP13:Lee, bib:TPAMI21:Li2} while video enhancement \cite{bib:ICCV19:Chen, bib:BMVC18:Lv, bib:ICCV19:Jiang, bib:CVPR21:Zhang} is regarded as an extension of image enhancement by avoiding the flicking problem between frames. 
%Conventional low-light image enhancement solutions are based on either histogram equalization \cite{bib:TCE07:Ibrahim, bib:TIP13:Lee} or the Retinex theory \cite{bib:TIP16:Guo, bib:TIP97:Jobson}.
%More recent low-light image enhancement proposals adopt deep learning for better accuracy and robustness \cite{bib:PR17:Lore, bib:CVPR20:Guo, bib:TPAMI21:Li2}.
%These methods differ in network structures and learning strategies, and 
We refer readers to \cite{bib:TPAMI21:Li} for a comprehensive review.
Of our particular interest is Zero-DCE \cite{bib:CVPR20:Guo} and its follow-up, Zero-DCE++ \cite{bib:TPAMI21:Li2}, a state-of-the-art zero-shot learning based low-light image enhancement scheme.
The reasons are two-fold.
\textit{(i)}
There lacks paired training data \ie low-light images or videos and the corresponding well-illuminated versions, in real-world mobile video applications.
Therefore, zero-shot learning based solutions \cite{bib:CVPR20:Guo, bib:TPAMI21:Li2} are preferable over supervised learning ones \cite{bib:PR17:Lore, bib:ICCV19:Chen} since the former requires neither paired nor unpaired training data. 
\textit{(ii)} 
Zero-DCE \cite{bib:CVPR20:Guo} and Zero-DCE++ \cite{bib:TPAMI21:Li2} apply a model architecture that consists of a deep neural network and an image-to-curve mapping, making it one of the most lightweight solution while achieving competitive image enhancement performance.

The primary challenge to extend low-light enhancement from images to videos is the flicking problem \cite{bib:TPAMI21:Li}. 
Some studies implicitly solve the problem by substituting the 2D network architectures in image enhancement models to the corresponding 3D versions \cite{bib:BMVC18:Lv, bib:ICCV19:Jiang}. 
However, they require specialized equipment to collect video datasets for training the new 3D architectures.
A more practical strategy is to incorporate temporal consistency into the loss function of image enhancement models \cite{bib:ICCV19:Chen, bib:CVPR21:Zhang}. 
For example, Chen \etal~\cite{bib:ICCV19:Chen} proposed a self-consistency loss to tolerate minor differences between inputs while keeping the output stable.
Zhang \etal \cite{bib:CVPR21:Zhang} enforces temporal consistency by ensuring stable optical flows estimated from the videos. 

To this end, the low-light video enhancement module of \sysname extends existing low-light enhancement solutions from images to videos. 
\rev{And \sysname advances the prior arts by introducing novel designs}.
\textit{(i)} 
\rev{It notably reduces the processing time of low-light enhancement schemes, by replacing the iterative image-curve mapping using a non-iterative mapping function. 
For example, 
\sysname dramatically outperforms Zero-DCE \cite{bib:CVPR20:Guo} in terms of processing latency, one of the most lightweight low-light image enhancement method.}
\textit{(ii)}
It designs the novel temporal consistency loss and its training scheme compatible to zero-shot learning based image enhancement schemes without the need for large-scale paired video datasets.

\subsection{Energy-awareness in DNNs}
\label{subsec:related:energy}

%Energy conservation is a vital problem for mobile computing systems because mobile devices are usually battery-powered, and many applications expect a long battery lifetime for continuous execution~\cite{ bib:SOSP99:Flinn, bib:todaes2000:benini,  bib:infocom2015:hu, bib:Mobisys17:Huynh}.
%Some conventional studies explicitly solve this problem.
%Jason \etal~\cite{bib:WMCSA99:Flinn} implement PowerScope, a tool for profiling energy usage, and use it to reduce the energy consumption of an adaptive video playing application.
%Based on this tool, Jason \etal ~\cite{bib:SOSP99:Flinn}further adjusts the mobile applications' data fidelity to select the correct tradeoff between energy conservation and application quality.
%Hui \etal~\cite{bib:icpp2012:chen} collect system states and estimate the application energy cost for mobile application design.
%Lide \etal~\cite{bib:CODES2010:zhang} present PowerBooter, a power model construction technique to monitor power consumption for controlling activity states of individual components.

%To close the gap between the recent DNN-powered system design and energy conservation, some studies explore the profiling methods of DNN's energy consumption. 
%Tien-Ju \etal~\cite{bib:CVPR17:Yang} propose the energy estimation methodology using parameters extrapolated from actual hardware measurements and use it to guide the energy-aware DNN pruning process.
%\sysname follows the trend of applying energy profiler to guide the energy consumption accurately optimization~\cite{bib:SOSP99:Flinn, bib:WMCSA99:Flinn, bib:CVPR17:Yang}.

Existing DNN energy profiling methods couple with either the DNN topology (\eg model structure, computation path)~\cite{bib:Mobisys18:Liu} or the hardware platform (\eg on-chip memory capability, memory bandwidth)~\cite{bib:CVPR17:Yang, bib:TCS20:Lee}.
Tien-Ju \etal~\cite{bib:ssc2017:yang} propose a layer-wise estimation method, which takes approximately 10 seconds to output the normalized energy consumption of a DNN based on its architecture, sparsity, and bitwidth.
Yannan \etal ~\cite{bib:ICCAD2019:wu}present the hardware architecture-dependent energy profiling method.
Some other works generate the design-specific tables with predefined attributes to estimate DNN energy cost~\cite{bib:islped2018:ke, bib:ISPASS2019:parashar}.
However, the above methods cannot profile the DNN energy usage for providing exact runtime feedback on \textit{dynamic resource availability} with \textit{agnostic DNN topology}, because the sampling for the specific DNN prototype or platform is costly.

Unlike existing methods, \sysname builds the one-fits-all profiler for DNN energy usage based on the following two observations: %experimental observations:
\textit{(i)} despite the dynamic topology of video processing models, the underlying tensor operation (\ie the energy-dominating operation loops) and the data flow are stable.
\textit{(ii)} Despite the dynamic hardware resource supply (\eg on-chip memory capability, computation bandwidth), the hit ratio of on-chip memory access shows up the stable rules and within a small sampling set.

\subsection{Adaptive Vision Tasks on Mobile Devices}
\label{subsec:related:vision_tasks}

Adaptive vision tasks involve image/video delivery between distributed mobiles or between the mobile and cloud.
Examples include live video streaming~\cite{bib:sigcomm2020:kim}, video-on-demand~\cite{bib:mobicom2019:baig}, and virtual reality~\cite{bib:hotcloud2019:wang}.
The key challenge for such tasks is the reliance on networking conditions.
%They need to maximize the user satisfaction with the delivered visual content and counteract the fluctuated networking conditions~\cite{bib:CSUR22:Lee}.
To tackle this challenge, researchers propose two categories of solutions, \ie adaptive bitrate~\cite{bib:ICDCS2019:chen,bib:SIGCOMM2016:sun, bib:hotcloud2019:wang}, and super-resolution~\cite{ bib:mobicom2019:lee,bib:tmc2020:yi, bib:IMWUT2020:liu}.
They co-design the task demands (\eg visual quality, latency, or inference accuracy) with the streaming process.
%Specifically, the bitrate adaptation is the dominant approach, which is also the main driver behind the adaptive video streaming ~\cite{bib:ICDCS2019:chen, bib:SIGCOMM2016:sun}.
In bitrate adaptation, each video is split into segments. 
These segments are encoded using multiple bitrates and streamed with the suitable bitrate based on the network conditions \cite{bib:hotcloud2019:wang}. 
%Yiding \etal ~\cite{bib:hotcloud2019:wang} present CloudSeg, an edge-to-cloud framework that co-designs the cloud-side inference with video streaming for optimizing the latency and inference accuracy.
% super-resolution
%Sun \etal ~\cite{bib:SIGCOMM2016:sun} realize the video bitrate adaptation based on the data-driven throughput prediction.
In super-resolution scheme, compact content with low resolution or quality is transmitted, which is then enhanced at the receiver~\cite{bib:mobicom2019:lee,bib:tmc2020:yi}.
For example, Yeo \etal \cite{bib:hotnet2017:yeo} leverages the super-resolution DNNs to enable the content-aware video delivery.
Lee \etal ~\cite{bib:mobicom2019:lee} present MobiSR to boost the performance of on-device super-resolution.
%Kim \etal ~\cite{bib:sigcomm2020:kim} employ the informed frame selection mechanism to provide the selective enhancement on the current live stream.

Nevertheless, the concerns of these efforts are coupled with other edge or cloud devices. \sysname serves as a locally deployed middleware to directly provide video enhancement services for local applications.

\section{Conclusion}
\label{sec:conclude}

This paper presents \sysname, an energy-aware low-light video enhancement system that achieves near real-time performance with competitive enhanced visual quality.
It consists of a novel model for fast low-light video enhancement as well as an agile energy-aware controller to dynamically adjust the enhancement behaviors for energy conservation.
And experimental results show that \sysname outperforms existing image and video enhancement methods in latency and temporal stability while retaining satisfactory system energy efficiency.

\newrev{
There are several limitations of our work that may need future exploration.
\textit{(i)}
Although \sysname achieves near real-time processing for RGB videos, the frames are resized to $270\times480\times3$.
To support videos with higher resolutions, a more lightweight and efficient video enhancement model is necessary.
We plan to exploit the fixed-point operations common on mobile platforms and explore quantization and other model compression techniques to further accelerate the video enhancement model.
An interesting follow-up is to extend \sysname to videos in raw format to handle extremely low-light conditions.
\textit{(ii)}
The proposed energy profiler can be integrated into other mobile systems.
The energy profiler of \sysname is built for mobile GPUs or CPUs only for tractability.
Also, more efforts are needed to design the modular and extensible energy profiler for devices with heterogeneous computation resources, \eg GPU-CPU co-execution.
}

\begin{acks}
This work was partially supported by the National Key R\&D Program of China (2019YFB1703901), National Science Fund for Distinguished Young Scholars (62025205, 61725205), the National Natural Science Foundation of China (No. 62102317, 62032020), and the China Postdoctoral Science Foundation (No. 2021M702671). 
The authors also thank the anonymous reviewers for their constructive feedback that has made the work stronger.
\end{acks}

\bibliography{acmart}


\begin{thebibliography}{61}

%%% the Web address.

\ifx \showCODEN    \undefined \def \showCODEN     #1{\unskip}     \fi
\ifx \showDOI      \undefined \def \showDOI       #1{#1}\fi
\ifx \showISBNx    \undefined \def \showISBNx     #1{\unskip}     \fi
\ifx \showISBNxiii \undefined \def \showISBNxiii  #1{\unskip}     \fi
\ifx \showISSN     \undefined \def \showISSN      #1{\unskip}     \fi
\ifx \showLCCN     \undefined \def \showLCCN      #1{\unskip}     \fi
\ifx \shownote     \undefined \def \shownote      #1{#1}          \fi
\ifx \showarticletitle \undefined \def \showarticletitle #1{#1}   \fi
\ifx \showURL      \undefined \def \showURL       {\relax}        \fi
% invisible to TeX
\providecommand\bibfield[2]{#2}
\providecommand\bibinfo[2]{#2}
\providecommand\natexlab[1]{#1}
\providecommand\showeprint[2][]{arXiv:#2}

\bibitem[\protect\citeauthoryear{Al-Obaidi, Al-Qassar, Nasser, Alkhayyat,
  Humaidi, and Ibraheem}{Al-Obaidi et~al\mbox{.}}{2021}]%
        {bib:IDST2021:al}
\bibfield{author}{\bibinfo{person}{Abdulkareem Sh~Mahdi Al-Obaidi},
  \bibinfo{person}{Arif Al-Qassar}, \bibinfo{person}{Ahmed~R Nasser},
  \bibinfo{person}{Ahmed Alkhayyat}, \bibinfo{person}{Amjad~J Humaidi}, {and}
  \bibinfo{person}{Ibraheem~K Ibraheem}.} \bibinfo{year}{2021}\natexlab{}.
\newblock \showarticletitle{Embedded design and implementation of mobile robot
  for surveillance applications}.
\newblock \bibinfo{journal}{\emph{Indonesian Journal of Science and
  Technology}} \bibinfo{volume}{6}, \bibinfo{number}{2} (\bibinfo{year}{2021}),
  \bibinfo{pages}{427--440}.
\newblock


\bibitem[\protect\citeauthoryear{Ananthanarayanan, Bahl, Bod{\'\i}k,
  Chintalapudi, Philipose, Ravindranath, and Sinha}{Ananthanarayanan
  et~al\mbox{.}}{2017}]%
        {bib:Computer17:Ananthanarayanan}
\bibfield{author}{\bibinfo{person}{Ganesh Ananthanarayanan},
  \bibinfo{person}{Paramvir Bahl}, \bibinfo{person}{Peter Bod{\'\i}k},
  \bibinfo{person}{Krishna Chintalapudi}, \bibinfo{person}{Matthai Philipose},
  \bibinfo{person}{Lenin Ravindranath}, {and} \bibinfo{person}{Sudipta Sinha}.}
  \bibinfo{year}{2017}\natexlab{}.
\newblock \showarticletitle{Real-time video analytics: The killer app for edge
  computing}.
\newblock \bibinfo{journal}{\emph{IEEE Computer}} \bibinfo{volume}{50},
  \bibinfo{number}{10} (\bibinfo{year}{2017}), \bibinfo{pages}{58--67}.
\newblock


\bibitem[\protect\citeauthoryear{Baig, He, Qureshi, Qiu, Chen, Chen, and
  Hu}{Baig et~al\mbox{.}}{2019}]%
        {bib:mobicom2019:baig}
\bibfield{author}{\bibinfo{person}{Ghufran Baig}, \bibinfo{person}{Jian He},
  \bibinfo{person}{Mubashir~Adnan Qureshi}, \bibinfo{person}{Lili Qiu},
  \bibinfo{person}{Guohai Chen}, \bibinfo{person}{Peng Chen}, {and}
  \bibinfo{person}{Yinliang Hu}.} \bibinfo{year}{2019}\natexlab{}.
\newblock \showarticletitle{Jigsaw: Robust live 4k video streaming}. In
  \bibinfo{booktitle}{\emph{Proceedings of the International Conference on
  Mobile Computing and Networking}}. \bibinfo{publisher}{ACM},
  \bibinfo{address}{New York, NY, USA}, \bibinfo{pages}{1--16}.
\newblock


\bibitem[\protect\citeauthoryear{Benini and Micheli}{Benini and
  Micheli}{2000}]%
        {bib:todaes2000:benini}
\bibfield{author}{\bibinfo{person}{Luca Benini} {and}
  \bibinfo{person}{Giovanni~de Micheli}.} \bibinfo{year}{2000}\natexlab{}.
\newblock \showarticletitle{System-level power optimization: techniques and
  tools}.
\newblock \bibinfo{journal}{\emph{ACM Transactions on Design Automation of
  Electronic Systems (TODAES)}} \bibinfo{volume}{5}, \bibinfo{number}{2}
  (\bibinfo{year}{2000}), \bibinfo{pages}{115--192}.
\newblock


\bibitem[\protect\citeauthoryear{Cai, Gu, and Zhang}{Cai et~al\mbox{.}}{2018}]%
        {bib:tip2018:cai}
\bibfield{author}{\bibinfo{person}{Jianrui Cai}, \bibinfo{person}{Shuhang Gu},
  {and} \bibinfo{person}{Lei Zhang}.} \bibinfo{year}{2018}\natexlab{}.
\newblock \showarticletitle{Learning a deep single image contrast enhancer from
  multi-exposure images}.
\newblock \bibinfo{journal}{\emph{IEEE Transactions on Image Processing}}
  \bibinfo{volume}{27}, \bibinfo{number}{4} (\bibinfo{year}{2018}),
  \bibinfo{pages}{2049--2062}.
\newblock


\bibitem[\protect\citeauthoryear{Chen, Chen, Do, and Koltun}{Chen
  et~al\mbox{.}}{2019a}]%
        {bib:ICCV19:Chen}
\bibfield{author}{\bibinfo{person}{Chen Chen}, \bibinfo{person}{Qifeng Chen},
  \bibinfo{person}{Minh~N Do}, {and} \bibinfo{person}{Vladlen Koltun}.}
  \bibinfo{year}{2019}\natexlab{a}.
\newblock \showarticletitle{Seeing motion in the dark}. In
  \bibinfo{booktitle}{\emph{Proceedings of the International Conference on
  Computer Vision}}. \bibinfo{publisher}{IEEE}, \bibinfo{address}{Piscataway,
  NJ, USA}, \bibinfo{pages}{3185--3194}.
\newblock


\bibitem[\protect\citeauthoryear{Chen, Moreau, Jiang, Zheng, Yan, Shen, Cowan,
  Wang, Hu, Ceze, et~al\mbox{.}}{Chen et~al\mbox{.}}{2018}]%
        {bib:osdi2018:chen}
\bibfield{author}{\bibinfo{person}{Tianqi Chen}, \bibinfo{person}{Thierry
  Moreau}, \bibinfo{person}{Ziheng Jiang}, \bibinfo{person}{Lianmin Zheng},
  \bibinfo{person}{Eddie Yan}, \bibinfo{person}{Haichen Shen},
  \bibinfo{person}{Meghan Cowan}, \bibinfo{person}{Leyuan Wang},
  \bibinfo{person}{Yuwei Hu}, \bibinfo{person}{Luis Ceze}, {et~al\mbox{.}}}
  \bibinfo{year}{2018}\natexlab{}.
\newblock \showarticletitle{$\{$TVM$\}$: An Automated $\{$End-to-End$\}$
  Optimizing Compiler for Deep Learning}. In
  \bibinfo{booktitle}{\emph{Proceedings of USENIX Symposium on Operating
  Systems Design and Implementation}}. \bibinfo{publisher}{USENIX},
  \bibinfo{address}{Berkeley, CA, USA}, \bibinfo{pages}{578--594}.
\newblock


\bibitem[\protect\citeauthoryear{Chen, Tan, and Cao}{Chen
  et~al\mbox{.}}{2019b}]%
        {bib:ICDCS2019:chen}
\bibfield{author}{\bibinfo{person}{Xianda Chen}, \bibinfo{person}{Tianxiang
  Tan}, {and} \bibinfo{person}{Guohong Cao}.} \bibinfo{year}{2019}\natexlab{b}.
\newblock \showarticletitle{Energy-aware and context-aware video streaming on
  smartphones}. In \bibinfo{booktitle}{\emph{Proceedings of the International
  Conference on Distributed Computing Systems}}. \bibinfo{publisher}{IEEE},
  \bibinfo{address}{Piscataway, NJ, USA}, \bibinfo{pages}{861--870}.
\newblock


\bibitem[\protect\citeauthoryear{Eric S.~Yuan}{Eric S.~Yuan}{2021}]%
        {app:zoom}
\bibfield{author}{\bibinfo{person}{etal Eric S.~Yuan, Ryan~Azus}.}
  \bibinfo{year}{2021}\natexlab{}.
\newblock \bibinfo{title}{Zoom Video Communications APP}.
\newblock
  \bibinfo{howpublished}{\url{https://support.zoom.us/hc/en-us/sections/200305413-Mobile}}.
\newblock


\bibitem[\protect\citeauthoryear{Fan, Liu, Su, Hui, and Niu}{Fan
  et~al\mbox{.}}{2020}]%
        {bib:percom2020:fan}
\bibfield{author}{\bibinfo{person}{Boyu Fan}, \bibinfo{person}{Xuefeng Liu},
  \bibinfo{person}{Xiang Su}, \bibinfo{person}{Pan Hui}, {and}
  \bibinfo{person}{Jianwei Niu}.} \bibinfo{year}{2020}\natexlab{}.
\newblock \showarticletitle{Emgauth: An emg-based smartphone unlocking system
  using siamese network}. In \bibinfo{booktitle}{\emph{Proceedings of the
  International Conference on Pervasive Computing and Communications}}.
  \bibinfo{publisher}{IEEE}, \bibinfo{address}{Piscataway, NJ, USA},
  \bibinfo{pages}{1--10}.
\newblock


\bibitem[\protect\citeauthoryear{Farina, Deb, and Amato}{Farina
  et~al\mbox{.}}{2004}]%
        {bib:TEC2004:farina}
\bibfield{author}{\bibinfo{person}{Marco Farina}, \bibinfo{person}{Kalyanmoy
  Deb}, {and} \bibinfo{person}{Paolo Amato}.} \bibinfo{year}{2004}\natexlab{}.
\newblock \showarticletitle{Dynamic multiobjective optimization problems: test
  cases, approximations, and applications}.
\newblock \bibinfo{journal}{\emph{IEEE Transactions on Evolutionary
  Computation}} \bibinfo{volume}{8}, \bibinfo{number}{5}
  (\bibinfo{year}{2004}), \bibinfo{pages}{425--442}.
\newblock


\bibitem[\protect\citeauthoryear{Farneb{\"a}ck}{Farneb{\"a}ck}{2003}]%
        {bib:farneback2003two}
\bibfield{author}{\bibinfo{person}{Gunnar Farneb{\"a}ck}.}
  \bibinfo{year}{2003}\natexlab{}.
\newblock \showarticletitle{Two-frame motion estimation based on polynomial
  expansion}. In \bibinfo{booktitle}{\emph{Scandinavian conference on Image
  analysis}}. Springer, \bibinfo{pages}{363--370}.
\newblock


\bibitem[\protect\citeauthoryear{Flinn and Satyanarayanan}{Flinn and
  Satyanarayanan}{1999}]%
        {bib:SOSP99:Flinn}
\bibfield{author}{\bibinfo{person}{Jason Flinn} {and} \bibinfo{person}{Mahadev
  Satyanarayanan}.} \bibinfo{year}{1999}\natexlab{}.
\newblock \showarticletitle{Energy-aware adaptation for mobile applications}.
  In \bibinfo{booktitle}{\emph{Proceedings of the Symposium on Operating
  Systems Principles}}. \bibinfo{publisher}{ACM}, \bibinfo{address}{New York,
  NY, USA}, \bibinfo{pages}{48--63}.
\newblock


\bibitem[\protect\citeauthoryear{Guo, Li, Guo, Loy, Hou, Kwong, and Cong}{Guo
  et~al\mbox{.}}{2020}]%
        {bib:CVPR20:Guo}
\bibfield{author}{\bibinfo{person}{Chunle Guo}, \bibinfo{person}{Chongyi Li},
  \bibinfo{person}{Jichang Guo}, \bibinfo{person}{Chen~Change Loy},
  \bibinfo{person}{Junhui Hou}, \bibinfo{person}{Sam Kwong}, {and}
  \bibinfo{person}{Runmin Cong}.} \bibinfo{year}{2020}\natexlab{}.
\newblock \showarticletitle{Zero-reference deep curve estimation for low-light
  image enhancement}. In \bibinfo{booktitle}{\emph{Proceedings of the
  Conference on Computer Vision and Pattern Recognition}}.
  \bibinfo{publisher}{IEEE}, \bibinfo{address}{Piscataway, NJ, USA},
  \bibinfo{pages}{1780--1789}.
\newblock


\bibitem[\protect\citeauthoryear{Guo, Li, and Ling}{Guo et~al\mbox{.}}{2016}]%
        {bib:TIP16:Guo}
\bibfield{author}{\bibinfo{person}{Xiaojie Guo}, \bibinfo{person}{Yu Li}, {and}
  \bibinfo{person}{Haibin Ling}.} \bibinfo{year}{2016}\natexlab{}.
\newblock \showarticletitle{LIME: Low-light image enhancement via illumination
  map estimation}.
\newblock \bibinfo{journal}{\emph{IEEE Transactions on Image Processing}}
  \bibinfo{volume}{26}, \bibinfo{number}{2} (\bibinfo{year}{2016}),
  \bibinfo{pages}{982--993}.
\newblock


\bibitem[\protect\citeauthoryear{Huang, Liu, Van Der~Maaten, and
  Weinberger}{Huang et~al\mbox{.}}{2017}]%
        {bib:CVPR17:Huang}
\bibfield{author}{\bibinfo{person}{Gao Huang}, \bibinfo{person}{Zhuang Liu},
  \bibinfo{person}{Laurens Van Der~Maaten}, {and} \bibinfo{person}{Kilian~Q
  Weinberger}.} \bibinfo{year}{2017}\natexlab{}.
\newblock \showarticletitle{Densely connected convolutional networks}. In
  \bibinfo{booktitle}{\emph{Proceedings of the Conference on Computer Vision
  and Pattern Recognition}}. \bibinfo{publisher}{IEEE},
  \bibinfo{address}{Piscataway, NJ, USA}, \bibinfo{pages}{4700--4708}.
\newblock


\bibitem[\protect\citeauthoryear{Ibrahim and Kong}{Ibrahim and Kong}{2007}]%
        {bib:TCE07:Ibrahim}
\bibfield{author}{\bibinfo{person}{Haidi Ibrahim} {and}
  \bibinfo{person}{Nicholas Sia~Pik Kong}.} \bibinfo{year}{2007}\natexlab{}.
\newblock \showarticletitle{Brightness preserving dynamic histogram
  equalization for image contrast enhancement}.
\newblock \bibinfo{journal}{\emph{IEEE Transactions on Consumer Electronics}}
  \bibinfo{volume}{53}, \bibinfo{number}{4} (\bibinfo{year}{2007}),
  \bibinfo{pages}{1752--1758}.
\newblock


\bibitem[\protect\citeauthoryear{Janveja, Nambi, Bannur, Gupta, and
  Padmanabhan}{Janveja et~al\mbox{.}}{2020}]%
        {bib:imwut20:janveja}
\bibfield{author}{\bibinfo{person}{Ishani Janveja}, \bibinfo{person}{Akshay
  Nambi}, \bibinfo{person}{Shruthi Bannur}, \bibinfo{person}{Sanchit Gupta},
  {and} \bibinfo{person}{Venkat Padmanabhan}.} \bibinfo{year}{2020}\natexlab{}.
\newblock \showarticletitle{Insight: monitoring the state of the driver in
  low-light using smartphones}.
\newblock \bibinfo{journal}{\emph{Proceedings of the ACM on Interactive,
  Mobile, Wearable and Ubiquitous Technologies}} \bibinfo{volume}{4},
  \bibinfo{number}{3} (\bibinfo{year}{2020}), \bibinfo{pages}{1--29}.
\newblock


\bibitem[\protect\citeauthoryear{Jiang and Zheng}{Jiang and Zheng}{2019}]%
        {bib:ICCV19:Jiang}
\bibfield{author}{\bibinfo{person}{Haiyang Jiang} {and}
  \bibinfo{person}{Yinqiang Zheng}.} \bibinfo{year}{2019}\natexlab{}.
\newblock \showarticletitle{Learning to see moving objects in the dark}. In
  \bibinfo{booktitle}{\emph{Proceedings of the International Conference on
  Computer Vision}}. \bibinfo{publisher}{IEEE}, \bibinfo{address}{Piscataway,
  NJ, USA}, \bibinfo{pages}{7324--7333}.
\newblock


\bibitem[\protect\citeauthoryear{Jobson, Rahman, and Woodell}{Jobson
  et~al\mbox{.}}{1997}]%
        {bib:TIP97:Jobson}
\bibfield{author}{\bibinfo{person}{Daniel~J Jobson}, \bibinfo{person}{Zia-ur
  Rahman}, {and} \bibinfo{person}{Glenn~A Woodell}.}
  \bibinfo{year}{1997}\natexlab{}.
\newblock \showarticletitle{A multiscale retinex for bridging the gap between
  color images and the human observation of scenes}.
\newblock \bibinfo{journal}{\emph{IEEE Transactions on Image Processing}}
  \bibinfo{volume}{6}, \bibinfo{number}{7} (\bibinfo{year}{1997}),
  \bibinfo{pages}{965--976}.
\newblock


\bibitem[\protect\citeauthoryear{Juang and Juang}{Juang and Juang}{2012}]%
        {bib:INDIN12:Juang}
\bibfield{author}{\bibinfo{person}{Shih-Yao Juang} {and}
  \bibinfo{person}{Jih-Gau Juang}.} \bibinfo{year}{2012}\natexlab{}.
\newblock \showarticletitle{Real-time indoor surveillance based on smartphone
  and mobile robot}. In \bibinfo{booktitle}{\emph{Proceedings of the
  International Conference on Industrial Informatics}}.
  \bibinfo{publisher}{IEEE}, \bibinfo{address}{Piscataway, NJ, USA},
  \bibinfo{pages}{475--480}.
\newblock


\bibitem[\protect\citeauthoryear{Ke, He, and Zhang}{Ke et~al\mbox{.}}{2018}]%
        {bib:islped2018:ke}
\bibfield{author}{\bibinfo{person}{Liu Ke}, \bibinfo{person}{Xin He}, {and}
  \bibinfo{person}{Xuan Zhang}.} \bibinfo{year}{2018}\natexlab{}.
\newblock \showarticletitle{Nnest: Early-stage design space exploration tool
  for neural network inference accelerators}. In
  \bibinfo{booktitle}{\emph{Proceedings of the International Symposium on Low
  Power Electronics and Design}}. \bibinfo{publisher}{ACM},
  \bibinfo{address}{New York, NY, USA}, \bibinfo{pages}{1--6}.
\newblock


\bibitem[\protect\citeauthoryear{Kim, Jung, Yeo, Ye, and Han}{Kim
  et~al\mbox{.}}{2020}]%
        {bib:sigcomm2020:kim}
\bibfield{author}{\bibinfo{person}{Jaehong Kim}, \bibinfo{person}{Youngmok
  Jung}, \bibinfo{person}{Hyunho Yeo}, \bibinfo{person}{Juncheol Ye}, {and}
  \bibinfo{person}{Dongsu Han}.} \bibinfo{year}{2020}\natexlab{}.
\newblock \showarticletitle{Neural-enhanced live streaming: Improving live
  video ingest via online learning}. In \bibinfo{booktitle}{\emph{Proceedings
  of the SIGCOMM Conference}}. \bibinfo{publisher}{ACM}, \bibinfo{address}{New
  York, NY, USA}, \bibinfo{pages}{107--125}.
\newblock


\bibitem[\protect\citeauthoryear{Lee, Lee, and Kim}{Lee et~al\mbox{.}}{2013}]%
        {bib:TIP13:Lee}
\bibfield{author}{\bibinfo{person}{Chulwoo Lee}, \bibinfo{person}{Chul Lee},
  {and} \bibinfo{person}{Chang-Su Kim}.} \bibinfo{year}{2013}\natexlab{}.
\newblock \showarticletitle{Contrast enhancement based on layered difference
  representation of 2D histograms}.
\newblock \bibinfo{journal}{\emph{IEEE Transactions on Image Processing}}
  \bibinfo{volume}{22}, \bibinfo{number}{12} (\bibinfo{year}{2013}),
  \bibinfo{pages}{5372--5384}.
\newblock


\bibitem[\protect\citeauthoryear{Lee, Kang, Lee, Shin, Han, and Yoo}{Lee
  et~al\mbox{.}}{2020}]%
        {bib:TCS20:Lee}
\bibfield{author}{\bibinfo{person}{Jinsu Lee}, \bibinfo{person}{Sanghoon Kang},
  \bibinfo{person}{Jinmook Lee}, \bibinfo{person}{Dongjoo Shin},
  \bibinfo{person}{Donghyeon Han}, {and} \bibinfo{person}{Hoi-Jun Yoo}.}
  \bibinfo{year}{2020}\natexlab{}.
\newblock \showarticletitle{The hardware and algorithm co-design for
  energy-efficient DNN processor on edge/mobile devices}.
\newblock \bibinfo{journal}{\emph{IEEE Transactions on Circuits and Systems I}}
  \bibinfo{volume}{67}, \bibinfo{number}{10} (\bibinfo{year}{2020}),
  \bibinfo{pages}{3458--3470}.
\newblock


\bibitem[\protect\citeauthoryear{Lee, Venieris, Dudziak, Bhattacharya, and
  Lane}{Lee et~al\mbox{.}}{2019}]%
        {bib:mobicom2019:lee}
\bibfield{author}{\bibinfo{person}{Royson Lee}, \bibinfo{person}{Stylianos~I
  Venieris}, \bibinfo{person}{Lukasz Dudziak}, \bibinfo{person}{Sourav
  Bhattacharya}, {and} \bibinfo{person}{Nicholas~D Lane}.}
  \bibinfo{year}{2019}\natexlab{}.
\newblock \showarticletitle{Mobisr: Efficient on-device super-resolution
  through heterogeneous mobile processors}. In
  \bibinfo{booktitle}{\emph{Proceedings of the International Conference on
  Mobile Computing and Networking}}. \bibinfo{publisher}{ACM},
  \bibinfo{address}{New York, NY, USA}, \bibinfo{pages}{1--16}.
\newblock


\bibitem[\protect\citeauthoryear{Lee and Nirjon}{Lee and Nirjon}{2020}]%
        {bib:rtas2020:lee}
\bibfield{author}{\bibinfo{person}{Seulki Lee} {and} \bibinfo{person}{Shahriar
  Nirjon}.} \bibinfo{year}{2020}\natexlab{}.
\newblock \showarticletitle{SubFlow: A dynamic induced-subgraph strategy toward
  real-time DNN inference and training}. In
  \bibinfo{booktitle}{\emph{Proceedings of the IEEE Real-Time and Embedded
  Technology and Applications Symposium}}. \bibinfo{publisher}{IEEE},
  \bibinfo{address}{Piscataway, NJ, USA}, \bibinfo{pages}{15--29}.
\newblock


\bibitem[\protect\citeauthoryear{Li, Guo, and Chen}{Li et~al\mbox{.}}{2021a}]%
        {bib:TPAMI21:Li2}
\bibfield{author}{\bibinfo{person}{Chongyi Li}, \bibinfo{person}{Chunle Guo},
  {and} \bibinfo{person}{Change~Loy Chen}.} \bibinfo{year}{2021}\natexlab{a}.
\newblock \showarticletitle{Learning to enhance low-light image via
  zero-reference deep curve estimation}.
\newblock \bibinfo{journal}{\emph{IEEE Transactions on Pattern Analysis \&
  Machine Intelligence}} \bibinfo{volume}{0}, \bibinfo{number}{01}
  (\bibinfo{year}{2021}), \bibinfo{pages}{1--1}.
\newblock


\bibitem[\protect\citeauthoryear{Li, Guo, Han, Jiang, Cheng, Gu, and Loy}{Li
  et~al\mbox{.}}{2021b}]%
        {bib:TPAMI21:Li}
\bibfield{author}{\bibinfo{person}{Chongyi Li}, \bibinfo{person}{Chunle Guo},
  \bibinfo{person}{Ling-Hao Han}, \bibinfo{person}{Jun Jiang},
  \bibinfo{person}{Ming-Ming Cheng}, \bibinfo{person}{Jinwei Gu}, {and}
  \bibinfo{person}{Chen~Change Loy}.} \bibinfo{year}{2021}\natexlab{b}.
\newblock \showarticletitle{Low-light image and video enhancement using deep
  learning: a survey}.
\newblock \bibinfo{journal}{\emph{IEEE Transactions on Pattern Analysis \&
  Machine Intelligence}} \bibinfo{volume}{0}, \bibinfo{number}{01}
  (\bibinfo{year}{2021}), \bibinfo{pages}{1--1}.
\newblock


\bibitem[\protect\citeauthoryear{Li, Wang, Wang, Tai, Qian, Yang, Wang, Li, and
  Huang}{Li et~al\mbox{.}}{2019}]%
        {bib:cvpr2019:li}
\bibfield{author}{\bibinfo{person}{Jian Li}, \bibinfo{person}{Yabiao Wang},
  \bibinfo{person}{Changan Wang}, \bibinfo{person}{Ying Tai},
  \bibinfo{person}{Jianjun Qian}, \bibinfo{person}{Jian Yang},
  \bibinfo{person}{Chengjie Wang}, \bibinfo{person}{Jilin Li}, {and}
  \bibinfo{person}{Feiyue Huang}.} \bibinfo{year}{2019}\natexlab{}.
\newblock \showarticletitle{DSFD: dual shot face detector}. In
  \bibinfo{booktitle}{\emph{Proceedings of the Conference on Computer Vision
  and Pattern Recognition}}. \bibinfo{publisher}{IEEE},
  \bibinfo{address}{Piscataway, NJ, USA}, \bibinfo{pages}{5060--5069}.
\newblock


\bibitem[\protect\citeauthoryear{Li, Aaron, Katsavounidis, Moorthy, and
  Manohara}{Li et~al\mbox{.}}{2016}]%
        {bib:li2016toward}
\bibfield{author}{\bibinfo{person}{Zhi Li}, \bibinfo{person}{Anne Aaron},
  \bibinfo{person}{Ioannis Katsavounidis}, \bibinfo{person}{Anush Moorthy},
  {and} \bibinfo{person}{Megha Manohara}.} \bibinfo{year}{2016}\natexlab{}.
\newblock \bibinfo{title}{Toward a practical perceptual video quality metric}.
\newblock
\newblock


\bibitem[\protect\citeauthoryear{Liu and Zhu}{Liu and Zhu}{2018}]%
        {bib:cvpr2018:liu}
\bibfield{author}{\bibinfo{person}{Mason Liu} {and} \bibinfo{person}{Menglong
  Zhu}.} \bibinfo{year}{2018}\natexlab{}.
\newblock \showarticletitle{Mobile video object detection with temporally-aware
  feature maps}. In \bibinfo{booktitle}{\emph{Proceedings of the Conference on
  Computer Vision and Pattern Recognition}}. \bibinfo{publisher}{IEEE},
  \bibinfo{address}{Piscataway, NJ, USA}, \bibinfo{pages}{5686--5695}.
\newblock


\bibitem[\protect\citeauthoryear{Liu, Lin, Zhou, Nan, Liu, and Du}{Liu
  et~al\mbox{.}}{2018}]%
        {bib:Mobisys18:Liu}
\bibfield{author}{\bibinfo{person}{Sicong Liu}, \bibinfo{person}{Yingyan Lin},
  \bibinfo{person}{Zimu Zhou}, \bibinfo{person}{Kaiming Nan},
  \bibinfo{person}{Hui Liu}, {and} \bibinfo{person}{Junzhao Du}.}
  \bibinfo{year}{2018}\natexlab{}.
\newblock \showarticletitle{On-demand deep model compression for mobile
  devices: A usage-driven model selection framework}. In
  \bibinfo{booktitle}{\emph{Proceedings of the Annual International Conference
  on Mobile Systems, Applications, and Services}}. \bibinfo{publisher}{ACM},
  \bibinfo{address}{New York, NY, USA}, \bibinfo{pages}{389--400}.
\newblock


\bibitem[\protect\citeauthoryear{Liu, Li, Fromm, Wang, Jiang, Mariakakis, and
  Patel}{Liu et~al\mbox{.}}{2021}]%
        {bib:IMWUT2020:liu}
\bibfield{author}{\bibinfo{person}{Xin Liu}, \bibinfo{person}{Yuang Li},
  \bibinfo{person}{Josh Fromm}, \bibinfo{person}{Yuntao Wang},
  \bibinfo{person}{Ziheng Jiang}, \bibinfo{person}{Alex Mariakakis}, {and}
  \bibinfo{person}{Shwetak Patel}.} \bibinfo{year}{2021}\natexlab{}.
\newblock \showarticletitle{SplitSR: An end-to-end approach to super-resolution
  on mobile devices}.
\newblock \bibinfo{journal}{\emph{Proceedings of the ACM on Interactive,
  Mobile, Wearable and Ubiquitous Technologies}} \bibinfo{volume}{5},
  \bibinfo{number}{1} (\bibinfo{year}{2021}), \bibinfo{pages}{1--20}.
\newblock


\bibitem[\protect\citeauthoryear{Lore, Akintayo, and Sarkar}{Lore
  et~al\mbox{.}}{2017}]%
        {bib:PR17:Lore}
\bibfield{author}{\bibinfo{person}{Kin~Gwn Lore}, \bibinfo{person}{Adedotun
  Akintayo}, {and} \bibinfo{person}{Soumik Sarkar}.}
  \bibinfo{year}{2017}\natexlab{}.
\newblock \showarticletitle{LLNet: A deep autoencoder approach to natural
  low-light image enhancement}.
\newblock \bibinfo{journal}{\emph{Pattern Recognition}}  \bibinfo{volume}{61}
  (\bibinfo{year}{2017}), \bibinfo{pages}{650--662}.
\newblock


\bibitem[\protect\citeauthoryear{Lv, Lu, Wu, and Lim}{Lv et~al\mbox{.}}{2018}]%
        {bib:BMVC18:Lv}
\bibfield{author}{\bibinfo{person}{Feifan Lv}, \bibinfo{person}{Feng Lu},
  \bibinfo{person}{Jianhua Wu}, {and} \bibinfo{person}{Chongsoon Lim}.}
  \bibinfo{year}{2018}\natexlab{}.
\newblock \showarticletitle{MBLLEN: Low-Light Image/Video Enhancement Using
  CNNs.}. In \bibinfo{booktitle}{\emph{Proceedings of the British Machine
  Vision Conference}}. \bibinfo{publisher}{The BMVA}, \bibinfo{address}{Durham,
  UK}, \bibinfo{pages}{220}.
\newblock


\bibitem[\protect\citeauthoryear{Mittal, Soundararajan, and Bovik}{Mittal
  et~al\mbox{.}}{2012}]%
        {bib:SPL2012:mittal}
\bibfield{author}{\bibinfo{person}{Anish Mittal}, \bibinfo{person}{Rajiv
  Soundararajan}, {and} \bibinfo{person}{Alan~C Bovik}.}
  \bibinfo{year}{2012}\natexlab{}.
\newblock \showarticletitle{Making a “completely blind” image quality
  analyzer}.
\newblock \bibinfo{journal}{\emph{IEEE Signal processing letters}}
  \bibinfo{volume}{20}, \bibinfo{number}{3} (\bibinfo{year}{2012}),
  \bibinfo{pages}{209--212}.
\newblock


\bibitem[\protect\citeauthoryear{Neumann, Karg, Zhang, Scharfenberger, Piegert,
  Mistr, Prokofyeva, Thiel, Vedaldi, Zisserman, and Schiele}{Neumann
  et~al\mbox{.}}{2018}]%
        {data:NightOwls}
\bibfield{author}{\bibinfo{person}{Luk{\'a}{\v{s}} Neumann},
  \bibinfo{person}{Michelle Karg}, \bibinfo{person}{Shanshan Zhang},
  \bibinfo{person}{Christian Scharfenberger}, \bibinfo{person}{Eric Piegert},
  \bibinfo{person}{Sarah Mistr}, \bibinfo{person}{Olga Prokofyeva},
  \bibinfo{person}{Robert Thiel}, \bibinfo{person}{Andrea Vedaldi},
  \bibinfo{person}{Andrew Zisserman}, {and} \bibinfo{person}{Bernt Schiele}.}
  \bibinfo{year}{2018}\natexlab{}.
\newblock \showarticletitle{NightOwls: A pedestrians at night dataset}. In
  \bibinfo{booktitle}{\emph{Asian Conference on Computer Vision}}.
  \bibinfo{publisher}{Springer}, \bibinfo{address}{Berlin, Germany},
  \bibinfo{pages}{691--705}.
\newblock


\bibitem[\protect\citeauthoryear{Parashar, Raina, Shao, Chen, Ying, Mukkara,
  Venkatesan, Khailany, Keckler, and Emer}{Parashar et~al\mbox{.}}{2019}]%
        {bib:ISPASS2019:parashar}
\bibfield{author}{\bibinfo{person}{Angshuman Parashar},
  \bibinfo{person}{Priyanka Raina}, \bibinfo{person}{Yakun~Sophia Shao},
  \bibinfo{person}{Yu-Hsin Chen}, \bibinfo{person}{Victor~A Ying},
  \bibinfo{person}{Anurag Mukkara}, \bibinfo{person}{Rangharajan Venkatesan},
  \bibinfo{person}{Brucek Khailany}, \bibinfo{person}{Stephen~W Keckler}, {and}
  \bibinfo{person}{Joel Emer}.} \bibinfo{year}{2019}\natexlab{}.
\newblock \showarticletitle{Timeloop: A systematic approach to dnn accelerator
  evaluation}. In \bibinfo{booktitle}{\emph{Proceedings of the International
  Symposium on Performance Analysis of Systems and Software}}.
  \bibinfo{publisher}{IEEE}, \bibinfo{address}{Piscataway, NJ, USA},
  \bibinfo{pages}{304--315}.
\newblock


\bibitem[\protect\citeauthoryear{Rahman, Rahman, Abdullah-Al-Wadud, Al-Quaderi,
  and Shoyaib}{Rahman et~al\mbox{.}}{2016}]%
        {bib:JIVP16:Rahman}
\bibfield{author}{\bibinfo{person}{Shanto Rahman},
  \bibinfo{person}{Md~Mostafijur Rahman}, \bibinfo{person}{Mohammad
  Abdullah-Al-Wadud}, \bibinfo{person}{Golam~Dastegir Al-Quaderi}, {and}
  \bibinfo{person}{Mohammad Shoyaib}.} \bibinfo{year}{2016}\natexlab{}.
\newblock \showarticletitle{An adaptive gamma correction for image
  enhancement}.
\newblock \bibinfo{journal}{\emph{EURASIP Journal on Image and Video
  Processing}} \bibinfo{volume}{2016}, \bibinfo{number}{1}
  (\bibinfo{year}{2016}), \bibinfo{pages}{1--13}.
\newblock


\bibitem[\protect\citeauthoryear{Ren, Liu, Ma, Xu, Xu, Cao, Du, and Yang}{Ren
  et~al\mbox{.}}{2019}]%
        {bib:TIP19:Ren}
\bibfield{author}{\bibinfo{person}{Wenqi Ren}, \bibinfo{person}{Sifei Liu},
  \bibinfo{person}{Lin Ma}, \bibinfo{person}{Qianqian Xu},
  \bibinfo{person}{Xiangyu Xu}, \bibinfo{person}{Xiaochun Cao},
  \bibinfo{person}{Junping Du}, {and} \bibinfo{person}{Ming-Hsuan Yang}.}
  \bibinfo{year}{2019}\natexlab{}.
\newblock \showarticletitle{Low-light image enhancement via a deep hybrid
  network}.
\newblock \bibinfo{journal}{\emph{IEEE Transactions on Image Processing}}
  \bibinfo{volume}{28}, \bibinfo{number}{9} (\bibinfo{year}{2019}),
  \bibinfo{pages}{4364--4375}.
\newblock


\bibitem[\protect\citeauthoryear{Ronneberger, Fischer, and Brox}{Ronneberger
  et~al\mbox{.}}{2015}]%
        {bib:MICCAI15:Ronneberger}
\bibfield{author}{\bibinfo{person}{Olaf Ronneberger}, \bibinfo{person}{Philipp
  Fischer}, {and} \bibinfo{person}{Thomas Brox}.}
  \bibinfo{year}{2015}\natexlab{}.
\newblock \showarticletitle{U-net: Convolutional networks for biomedical image
  segmentation}. In \bibinfo{booktitle}{\emph{Proceedings of the International
  Conference on Medical Image Computing and Computer-Assisted Intervention}}.
  \bibinfo{publisher}{Springer}, \bibinfo{address}{Berlin, Germany},
  \bibinfo{pages}{234--241}.
\newblock


\bibitem[\protect\citeauthoryear{Song, Kim, and Jeon}{Song
  et~al\mbox{.}}{2014}]%
        {bib:icce2014:song}
\bibfield{author}{\bibinfo{person}{Inchul Song}, \bibinfo{person}{Hyun-Jun
  Kim}, {and} \bibinfo{person}{Paul~Barom Jeon}.}
  \bibinfo{year}{2014}\natexlab{}.
\newblock \showarticletitle{Deep learning for real-time robust facial
  expression recognition on a smartphone}. In
  \bibinfo{booktitle}{\emph{Proceedings of the International Conference on
  Consumer Electronics}}. \bibinfo{publisher}{IEEE},
  \bibinfo{address}{Piscataway, NJ, USA}, \bibinfo{pages}{564--567}.
\newblock


\bibitem[\protect\citeauthoryear{Sun, Yin, Jiang, Sekar, Lin, Wang, Liu, and
  Sinopoli}{Sun et~al\mbox{.}}{2016}]%
        {bib:SIGCOMM2016:sun}
\bibfield{author}{\bibinfo{person}{Yi Sun}, \bibinfo{person}{Xiaoqi Yin},
  \bibinfo{person}{Junchen Jiang}, \bibinfo{person}{Vyas Sekar},
  \bibinfo{person}{Fuyuan Lin}, \bibinfo{person}{Nanshu Wang},
  \bibinfo{person}{Tao Liu}, {and} \bibinfo{person}{Bruno Sinopoli}.}
  \bibinfo{year}{2016}\natexlab{}.
\newblock \showarticletitle{CS2P: Improving video bitrate selection and
  adaptation with data-driven throughput prediction}. In
  \bibinfo{booktitle}{\emph{Proceedings of the SIGCOMM Conference}}.
  \bibinfo{publisher}{ACM}, \bibinfo{address}{New York, NY, USA},
  \bibinfo{pages}{272--285}.
\newblock


\bibitem[\protect\citeauthoryear{T-Mobile}{T-Mobile}{2022}]%
        {t-mobile-5G-report}
\bibfield{author}{\bibinfo{person}{T-Mobile}.} \bibinfo{year}{2022}\natexlab{}.
\newblock \bibinfo{title}{{Internet Services | T-Mobile’s Broadband Internet
  Access Services}}.
\newblock
\newblock
\urldef\tempurl%
\url{https://www.t-mobile.com/responsibility/consumer-info/policies/internet-service}
\showURL{%
\tempurl}


\bibitem[\protect\citeauthoryear{Teli, Zvanovec, and Ghassemlooy}{Teli
  et~al\mbox{.}}{2019}]%
        {bib:ConTEL19:Teli}
\bibfield{author}{\bibinfo{person}{Shivani~Rajendra Teli},
  \bibinfo{person}{Stanislav Zvanovec}, {and} \bibinfo{person}{Zabih
  Ghassemlooy}.} \bibinfo{year}{2019}\natexlab{}.
\newblock \showarticletitle{The first tests of smartphone camera exposure
  effect on optical camera communication links}. In
  \bibinfo{booktitle}{\emph{Proceedings of the International Conference on
  Telecommunications}}. \bibinfo{publisher}{IEEE},
  \bibinfo{address}{Piscataway, NJ, USA}, \bibinfo{pages}{1--6}.
\newblock


\bibitem[\protect\citeauthoryear{Vonikakis}{Vonikakis}{2021}]%
        {data:vv}
\bibfield{author}{\bibinfo{person}{Vasileios Vonikakis}.}
  \bibinfo{year}{2021}\natexlab{}.
\newblock \bibinfo{title}{California-ND: An annotated dataset for
  near-duplicates in personal photo-collections}.
\newblock
  \bibinfo{howpublished}{\url{https://sites.google.com/site/vonikakis/datasets}}.
\newblock


\bibitem[\protect\citeauthoryear{Wang, Wang, Zhang, Jiang, and Chen}{Wang
  et~al\mbox{.}}{2019}]%
        {bib:hotcloud2019:wang}
\bibfield{author}{\bibinfo{person}{Yiding Wang}, \bibinfo{person}{Weiyan Wang},
  \bibinfo{person}{Junxue Zhang}, \bibinfo{person}{Junchen Jiang}, {and}
  \bibinfo{person}{Kai Chen}.} \bibinfo{year}{2019}\natexlab{}.
\newblock \showarticletitle{Bridging the edge-cloud barrier for real-time
  advanced vision analytics}. In \bibinfo{booktitle}{\emph{Proceedings of the
  Workshop on Hot Topics in Cloud Computing}}. \bibinfo{publisher}{USENIX},
  \bibinfo{address}{Berkeley, CA, USA}, \bibinfo{pages}{1--7}.
\newblock


\bibitem[\protect\citeauthoryear{Wang, Bovik, Sheikh, and Simoncelli}{Wang
  et~al\mbox{.}}{2004}]%
        {bib:TIP2004:wang}
\bibfield{author}{\bibinfo{person}{Zhou Wang}, \bibinfo{person}{Alan~C Bovik},
  \bibinfo{person}{Hamid~R Sheikh}, {and} \bibinfo{person}{Eero~P Simoncelli}.}
  \bibinfo{year}{2004}\natexlab{}.
\newblock \showarticletitle{Image quality assessment: from error visibility to
  structural similarity}.
\newblock \bibinfo{journal}{\emph{IEEE transactions on image processing}}
  \bibinfo{volume}{13}, \bibinfo{number}{4} (\bibinfo{year}{2004}),
  \bibinfo{pages}{600--612}.
\newblock


\bibitem[\protect\citeauthoryear{Wei, Wang, Yang, and Liu}{Wei
  et~al\mbox{.}}{2018}]%
        {bib:arXiv2018:wei}
\bibfield{author}{\bibinfo{person}{Chen Wei}, \bibinfo{person}{Wenjing Wang},
  \bibinfo{person}{Wenhan Yang}, {and} \bibinfo{person}{Jiaying Liu}.}
  \bibinfo{year}{2018}\natexlab{}.
\newblock \bibinfo{title}{Deep retinex decomposition for low-light
  enhancement}.
\newblock
\newblock


\bibitem[\protect\citeauthoryear{Wiegand, Sullivan, Bjontegaard, and
  Luthra}{Wiegand et~al\mbox{.}}{2003}]%
        {bib:H2642003overview}
\bibfield{author}{\bibinfo{person}{Thomas Wiegand}, \bibinfo{person}{Gary~J
  Sullivan}, \bibinfo{person}{Gisle Bjontegaard}, {and} \bibinfo{person}{Ajay
  Luthra}.} \bibinfo{year}{2003}\natexlab{}.
\newblock \showarticletitle{Overview of the H. 264/AVC video coding standard}.
\newblock \bibinfo{journal}{\emph{IEEE Transactions on Circuits and Systems for
  Video Technology}} \bibinfo{volume}{13}, \bibinfo{number}{7}
  (\bibinfo{year}{2003}), \bibinfo{pages}{560--576}.
\newblock


\bibitem[\protect\citeauthoryear{Wijnands, Thompson, Nice, Aschwanden, and
  Stevenson}{Wijnands et~al\mbox{.}}{2020}]%
        {bib:nca2020:wijnands}
\bibfield{author}{\bibinfo{person}{Jasper~S Wijnands}, \bibinfo{person}{Jason
  Thompson}, \bibinfo{person}{Kerry~A Nice}, \bibinfo{person}{Gideon~DPA
  Aschwanden}, {and} \bibinfo{person}{Mark Stevenson}.}
  \bibinfo{year}{2020}\natexlab{}.
\newblock \showarticletitle{Real-time monitoring of driver drowsiness on mobile
  platforms using 3D neural networks}.
\newblock \bibinfo{journal}{\emph{Neural Computing and Applications}}
  \bibinfo{volume}{32}, \bibinfo{number}{13} (\bibinfo{year}{2020}),
  \bibinfo{pages}{9731--9743}.
\newblock


\bibitem[\protect\citeauthoryear{Wu, Emer, and Sze}{Wu et~al\mbox{.}}{2019}]%
        {bib:ICCAD2019:wu}
\bibfield{author}{\bibinfo{person}{Yannan~Nellie Wu}, \bibinfo{person}{Joel~S
  Emer}, {and} \bibinfo{person}{Vivienne Sze}.}
  \bibinfo{year}{2019}\natexlab{}.
\newblock \showarticletitle{Accelergy: An architecture-level energy estimation
  methodology for accelerator designs}. In
  \bibinfo{booktitle}{\emph{Procceedings of the International Conference on
  Computer-Aided Design}}. \bibinfo{publisher}{IEEE},
  \bibinfo{address}{Piscataway, NJ, USA}, \bibinfo{pages}{1--8}.
\newblock


\bibitem[\protect\citeauthoryear{Yang, Chen, Emer, and Sze}{Yang
  et~al\mbox{.}}{2017b}]%
        {bib:ssc2017:yang}
\bibfield{author}{\bibinfo{person}{Tien-Ju Yang}, \bibinfo{person}{Yu-Hsin
  Chen}, \bibinfo{person}{Joel Emer}, {and} \bibinfo{person}{Vivienne Sze}.}
  \bibinfo{year}{2017}\natexlab{b}.
\newblock \showarticletitle{A method to estimate the energy consumption of deep
  neural networks}. In \bibinfo{booktitle}{\emph{Proceedings of the asilomar
  conference on signals, systems, and computers}}. \bibinfo{publisher}{IEEE},
  \bibinfo{address}{Piscataway, NJ, USA}, \bibinfo{pages}{1916--1920}.
\newblock


\bibitem[\protect\citeauthoryear{Yang, Chen, and Sze}{Yang
  et~al\mbox{.}}{2017a}]%
        {bib:CVPR17:Yang}
\bibfield{author}{\bibinfo{person}{Tien-Ju Yang}, \bibinfo{person}{Yu-Hsin
  Chen}, {and} \bibinfo{person}{Vivienne Sze}.}
  \bibinfo{year}{2017}\natexlab{a}.
\newblock \showarticletitle{Designing energy-efficient convolutional neural
  networks using energy-aware pruning}. In
  \bibinfo{booktitle}{\emph{Proceedings of the Conference on Computer Vision
  and Pattern Recognition}}. \bibinfo{publisher}{IEEE},
  \bibinfo{address}{Piscataway, NJ, USA}, \bibinfo{pages}{5687--5695}.
\newblock


\bibitem[\protect\citeauthoryear{Yang, Yuan, Ren, Liu, Scheirer, Wang, Zhang,
  Zhong, Xie, Pu, et~al\mbox{.}}{Yang et~al\mbox{.}}{2020}]%
        {bib:top20:yang}
\bibfield{author}{\bibinfo{person}{Wenhan Yang}, \bibinfo{person}{Ye Yuan},
  \bibinfo{person}{Wenqi Ren}, \bibinfo{person}{Jiaying Liu},
  \bibinfo{person}{Walter~J Scheirer}, \bibinfo{person}{Zhangyang Wang},
  \bibinfo{person}{Taiheng Zhang}, \bibinfo{person}{Qiaoyong Zhong},
  \bibinfo{person}{Di Xie}, \bibinfo{person}{Shiliang Pu}, {et~al\mbox{.}}}
  \bibinfo{year}{2020}\natexlab{}.
\newblock \showarticletitle{Advancing image understanding in poor visibility
  environments: A collective benchmark study}.
\newblock \bibinfo{journal}{\emph{IEEE Transactions on Image Processing}}
  \bibinfo{volume}{29} (\bibinfo{year}{2020}), \bibinfo{pages}{5737--5752}.
\newblock


\bibitem[\protect\citeauthoryear{Yeo, Do, and Han}{Yeo et~al\mbox{.}}{2017}]%
        {bib:hotnet2017:yeo}
\bibfield{author}{\bibinfo{person}{Hyunho Yeo}, \bibinfo{person}{Sunghyun Do},
  {and} \bibinfo{person}{Dongsu Han}.} \bibinfo{year}{2017}\natexlab{}.
\newblock \showarticletitle{How will deep learning change internet video
  delivery?}. In \bibinfo{booktitle}{\emph{Proceedings of the Workshop on Hot
  Topics in Networks}}. \bibinfo{publisher}{ACM}, \bibinfo{address}{New York,
  NY, USA}, \bibinfo{pages}{57--64}.
\newblock


\bibitem[\protect\citeauthoryear{Yi, Kim, Kim, and Choi}{Yi
  et~al\mbox{.}}{2020}]%
        {bib:tmc2020:yi}
\bibfield{author}{\bibinfo{person}{Juheon Yi}, \bibinfo{person}{Seongwon Kim},
  \bibinfo{person}{Joongheon Kim}, {and} \bibinfo{person}{Sunghyun Choi}.}
  \bibinfo{year}{2020}\natexlab{}.
\newblock \showarticletitle{Supremo: Cloud-Assisted Low-Latency
  Super-Resolution in Mobile Devices}.
\newblock \bibinfo{journal}{\emph{IEEE Transactions on Mobile Computing}}
  \bibinfo{volume}{0} (\bibinfo{year}{2020}), \bibinfo{pages}{1--1}.
\newblock


\bibitem[\protect\citeauthoryear{Zhang, Li, You, and Fu}{Zhang
  et~al\mbox{.}}{2021b}]%
        {bib:CVPR21:Zhang}
\bibfield{author}{\bibinfo{person}{Fan Zhang}, \bibinfo{person}{Yu Li},
  \bibinfo{person}{Shaodi You}, {and} \bibinfo{person}{Ying Fu}.}
  \bibinfo{year}{2021}\natexlab{b}.
\newblock \showarticletitle{Learning Temporal Consistency for Low Light Video
  Enhancement From Single Images}. In \bibinfo{booktitle}{\emph{Proceedings of
  the Conference on Computer Vision and Pattern Recognition}}.
  \bibinfo{publisher}{IEEE}, \bibinfo{address}{Piscataway, NJ, USA},
  \bibinfo{pages}{4967--4976}.
\newblock


\bibitem[\protect\citeauthoryear{Zhang, Han, Wei, Zheng, Cao, Yang, and
  Liu}{Zhang et~al\mbox{.}}{2021a}]%
        {bib:mobisys21:zhang}
\bibfield{author}{\bibinfo{person}{Li~Lyna Zhang}, \bibinfo{person}{Shihao
  Han}, \bibinfo{person}{Jianyu Wei}, \bibinfo{person}{Ningxin Zheng},
  \bibinfo{person}{Ting Cao}, \bibinfo{person}{Yuqing Yang}, {and}
  \bibinfo{person}{Yunxin Liu}.} \bibinfo{year}{2021}\natexlab{a}.
\newblock \showarticletitle{nn-Meter: towards accurate latency prediction of
  deep-learning model inference on diverse edge devices}. In
  \bibinfo{booktitle}{\emph{Proceedings of the Annual International Conference
  on Mobile Systems, Applications, and Services}}. \bibinfo{publisher}{ACM},
  \bibinfo{address}{New York, NY, USA}, \bibinfo{pages}{81--93}.
\newblock


\bibitem[\protect\citeauthoryear{Zhu, Du, Wen, Bian, Ling, Hu, Peng, Zheng,
  Wang, Zhang, et~al\mbox{.}}{Zhu et~al\mbox{.}}{2019}]%
        {bib:ICCV2019:zhu}
\bibfield{author}{\bibinfo{person}{Pengfei Zhu}, \bibinfo{person}{Dawei Du},
  \bibinfo{person}{Longyin Wen}, \bibinfo{person}{Xiao Bian},
  \bibinfo{person}{Haibin Ling}, \bibinfo{person}{Qinghua Hu},
  \bibinfo{person}{Tao Peng}, \bibinfo{person}{Jiayu Zheng},
  \bibinfo{person}{Xinyao Wang}, \bibinfo{person}{Yue Zhang}, {et~al\mbox{.}}}
  \bibinfo{year}{2019}\natexlab{}.
\newblock \showarticletitle{Visdrone-vid2019: The vision meets drone object
  detection in video challenge results}. In
  \bibinfo{booktitle}{\emph{Proceedings of the International Conference on
  Computer Vision Workshops}}. \bibinfo{publisher}{IEEE},
  \bibinfo{address}{Piscataway, NJ, USA}, \bibinfo{pages}{0--0}.
\newblock


\end{thebibliography}
\bibliographystyle{ACM-Reference-Format}

\end{document}